\documentclass[sigconf]{acmart}
\AtBeginDocument{%
  }

\copyrightyear{2025}
\acmYear{2025}
\setcopyright{CC}
\acmConference[WSDM '25]{Proceedings of the Eighteenth ACM International Conference on Web Search and Data Mining}{March 10--14, 2025}{Hannover, Germany}
\acmBooktitle{Proceedings of the Eighteenth ACM International Conference on Web Search and Data Mining (WSDM '25), March 10--14, 2025, Hannover, Germany}\acmDOI{10.1145/3701551.3703562}
\acmISBN{979-8-4007-1329-3/25/03}

\usepackage{booktabs} 
\usepackage{amsmath}

\usepackage{amssymb}
\usepackage{mathtools}
\usepackage{amsthm}
\usepackage{amsfonts}
\usepackage{tabularray}
\usepackage{xcolor}
\usepackage{multirow}
\usepackage{adjustbox}
\usepackage{color,soul}
\usepackage{rotating} 
\usepackage{geometry}
\usepackage{graphicx}
\usepackage{subcaption}
\usepackage{subfloat}

\usepackage{array}
\input{Definitions.sty}

\newcommand{\change}[1]{{\color{black}#1}}

\usepackage{soul}
\newcommand{\mathcolorbox}[2]{\colorbox{#1}{$\displaystyle #2$}}

\theoremstyle{plain}

\theoremstyle{definition}

\theoremstyle{remark}

\newtheorem{example}{Example}

\newcommand{\subject}{\ensuremath{x}}
\newcommand{\relation}{\ensuremath{r}}
\newcommand{\object}{\ensuremath{y}}
\newcommand{\setOfObjects}{\ensuremath{\mathcal{Y}}}
\newcommand{\setOfFacts}{\ensuremath{\mathcal{F}}}
\newcommand{\prefix}{\ensuremath{\sigma}}

\newcommand{\llm}{\ensuremath{\theta}}
\newcommand{\fact}{\ensuremath{f}}

\newcommand{\lke}{\ensuremath{\phi}}

\newcommand{\dataset}{\ensuremath{\mathcal{D}}}
\DeclareMathOperator{\pred}{pred}
\DeclareMathOperator{\acc}{acc}

\newcommand{\prompt}{\ensuremath{\sigma}}

\newcommand{\trex}{T-REx}
\newcommand{\trexmc}{T-REx-MC}
\newcommand{\iclke}{ZP-LKE}
\newcommand{\eiclke}{EIC-LKE}
\newcommand{\asubr}{\eta}

\def\lkeone{PB-LKE}

\def\lketwo{IC-LKE}

\def\lkethree{EIC-LKE}

\begin{document}

\title{Towards Reliable Latent Knowledge Estimation in LLMs:\\Zero-Prompt Many-Shot Based Factual Knowledge Extraction}

\author{Qinyuan Wu}
\email{qwu@mpi-sws.org}
\orcid{}
\affiliation{%
  \institution{MPI-SWS}
  \city{Saarbruecken}
  \state{ }
  \country{Germany}
}

\author{Mohammad Aflah Khan}
\affiliation{%
  \institution{MPI-SWS}
  \city{Saarbruecken}
  \state{ }
  \country{Germany}
}

\author{Soumi Das}
\affiliation{%
  \institution{MPI-SWS}
  \city{Saarbruecken}
  \state{ }
  \country{Germany}
}

\author{Vedant Nanda}
\affiliation{%
  \institution{MPI-SWS}
  \city{Saarbruecken}
  \state{ }
  \country{Germany}
}

\author{Bishwamittra Ghosh}
\affiliation{%
  \institution{MPI-SWS}
  \city{Saarbruecken}
  \state{ }
  \country{Germany}
}
\author{Camila Kolling}
\affiliation{%
  \institution{MPI-SWS}
  \city{Saarbruecken}
  \state{ }
  \country{Germany}
}

\author{Till Speicher}
\affiliation{%
  \institution{MPI-SWS}
  \city{Saarbruecken}
  \state{ }
  \country{Germany}
}

\author{Laurent Bindschaedler}
\affiliation{%
  \institution{MPI-SWS}
  \city{Saarbruecken}
  \state{ }
  \country{Germany}
}

\author{Krishna Gummadi}
\affiliation{%
  \institution{MPI-SWS}
  \city{Saarbruecken}
  \state{ }
  \country{Germany}
}

\author{Evimaria Terzi}
\affiliation{
  \institution{Boston University}
  \city{Boston}
  \state{Massachusetts}
  \country{United States}
}

\renewcommand{\shortauthors}{Wu et al.}

\pdfminorversion=5
\pdfobjcompresslevel=0
\pdfcompresslevel=0

\begin{abstract}
In this paper, we focus on the challenging task of {\it reliably} estimating factual knowledge that is embedded inside large language models (LLMs). To avoid reliability concerns with prior approaches, we propose to eliminate prompt engineering when probing LLMs for factual knowledge. Our approach, called {\it Zero-Prompt Latent Knowledge Estimator (\iclke)}, leverages the in-context learning ability of LLMs to communicate both the factual knowledge question as well as the expected answer format.
Our knowledge estimator is both conceptually simpler (i.e., doesn't depend on meta-linguistic judgments of LLMs) and easier to apply (i.e., is not LLM-specific), and we demonstrate that it can surface more of the latent knowledge embedded in LLMs.
We also investigate how different design choices affect the performance of \iclke.
Using the proposed estimator, we perform a large-scale evaluation of the factual knowledge of a variety of open-source LLMs, like OPT, Pythia, Llama(2), Mistral, Gemma, etc. over a large set of relations and facts from the Wikidata knowledge base.
We observe differences in the factual knowledge between different model families and models of different sizes, that some relations are consistently better known than others but that models differ in the precise facts they know, and differences in the knowledge of base models and their finetuned counterparts.
\footnote{Code available at: \url{https://github.com/QinyuanWu0710/ZeroPrompt_LKE}}
\end{abstract}




\maketitle

\section{Introduction}
\label{sec:intro}
\begin{figure*}[htbp]
  \includegraphics[width=0.95\textwidth]{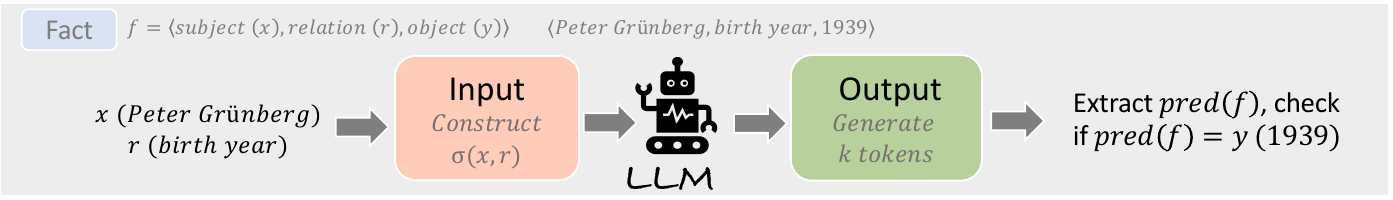}
  \caption{Overview of how Latent Knowledge Estimators (LKEs) work
  }
  \Description{}
  \label{fig:overview}
\end{figure*}

\begin{figure*}[htbp]
  \includegraphics[width=0.95\textwidth]{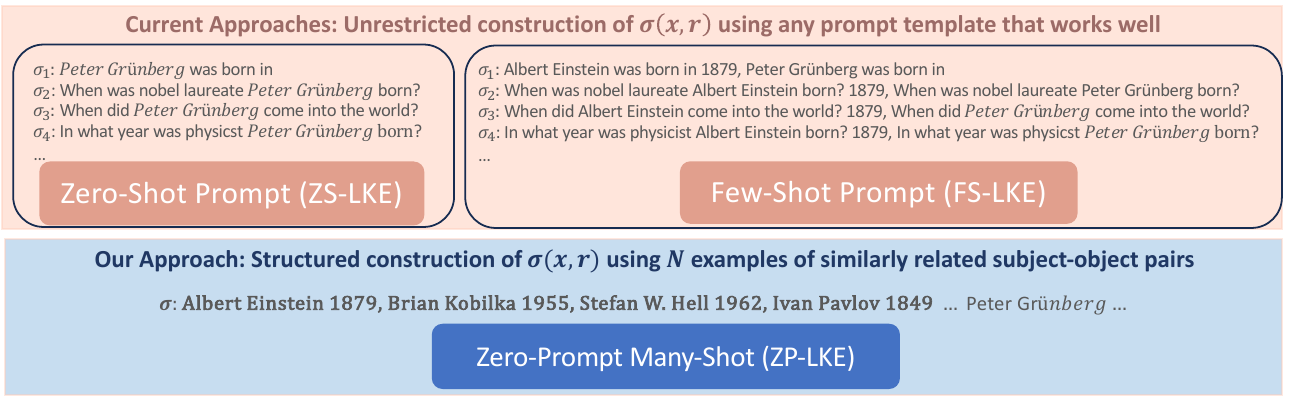}
  \caption{Current prompt-based (zero-shot and few-shot) LKE approaches vs. Our zero-prompt (many-shot) LKE approach
  }
  \Description{}
  \label{fig:input}
\end{figure*}

Conversational chatbots (e.g., OpenAI's ChatGPT) built around large language models (e.g., OpenAI's GPT) are increasingly being used for a variety of information retrieval tasks such as searching for information or seeking recommendations related to real-world entities like people or places~\citep{wu2023survey, zhu2023large}. A worrisome concern in such scenarios is the factual correctness of information generated by the LLMs~\citep{peng2023check, hu-etal-2023-refchecker,snyder2023early,yao2023llm,ji2023survey,zhang2023siren,wang2023survey, kalo2022kamel, cao2021knowledgeable, youssef2023give}. 

{\bf The latent knowledge estimation problem:} To avoid making false assertions about a real-world entity, an LLM first needs to have factual (true) knowledge about the entity. Given a prompt like {\it ``Einstein was born in the year''}, LLMs may generate both the correct answer ({\it ``1879''}) and wrong answers (e.g., {\it ``1878'' or ``1880''}) with some probabilities. If an LLM {\it knows} the fact, one can hope that the probability with which it would generate the correct answer would be much higher than the wrong answers~\cite{jiang2021can}. 
As LLMs are typically pretrained over a Web corpus (including Wikipedia data) with millions of facts about real-world entities, they have the opportunity to learn factual knowledge about our world and latently embed the knowledge in their parameters. But, {\it how can we estimate the extent of LLMs' knowledge of real-world facts?}

{\bf Reliability of latent knowledge estimates:}
Following~\cite{petroni2019language}, many prior works
\cite{jiang2020can,bouraoui2020inducing,shin2020autoprompt} 
represent factual knowledge 
in the form of triplets $\langle \subject, \relation, \object \rangle$, where the subject $\subject$ has a relation of type $\relation$ with the object $\object$ (e.g., $\langle\text{Einstein}, \text{birth-year}, 1879\rangle $). The central challenge of latent knowledge estimation is to infer $\object$ given $\subject$ and $\relation$ by {\it only} using information extracted from the LLM. Typically, the inference relies on probing the LLM with prompt templates, $\prompt(\subject, \relation)$, constructed to communicate the information of $\subject$ and $\relation$ and analyzing the generated responses for presence of $\object$ (see Figure~\ref{fig:overview}). 
Current approaches ~\cite{hu2023large, kalo2022kamel, jiang2021can, jiang2020can, shin2020autoprompt, sun_head--tail_2023}
allow unrestricted choice of prompt templates with few well-defined rules (see Figure~\ref{fig:input}). As a result, they are vulnerable to prompt engineering and prompt hacking, 
which raises serious concerns about the reliability of their estimates \cite{cao2021knowledgeable}. 
Against this background, in this paper, we make {\bf four} primary contributions:

{\bf 1. A structured reliable latent knowledge estimator (LKE) based on zero prompting (ZP):} Our latent knowledge estimator, called \iclke, is based on the following simple, yet novel and powerful insight. Rather than engineer input prompts $\prompt(\subject, \relation)$ that best communicate $\relation$ to an LLM, we let the LLM infer $\relation$ by simply providing multiple examples of $\langle \subject, \object \rangle$ pairs that share the same relation $\relation$ (see Figure~\ref{fig:input}). The key distinguishing feature of \iclke~ is its adherence to {\it zero prompting}, i.e., the input $\prompt(\subject, \relation)$ is highly structured and contains no prompt tokens outside other similarly related $\langle \subject, \object \rangle$ pairs. Thus, \iclke~ avoids the reliability risks associated with prompt engineering such as side-channels and over-fitting. (We discuss these reliability concerns in Section~\ref{sec:reliability_concerns}.)

\begin{table*}[h!]
\centering
 \small{
    \centering
     \scalebox{0.95}{
\begin{tabular}{p{5.4cm}|p{3.2cm}|p{5.4cm}|p{2.8cm}}
\toprule
\multicolumn{2}{c|}{\textbf{Prompt based LKEs (ZS-LKE \& FS-LKE)}}&\multicolumn{2}{|c}{\textbf{Zero-Prompt based LKE (\iclke)}} \\
\toprule
\textbf{Input} & \textbf{Output: Next $10$ tokens} & \textbf{Input} & \textbf{Output: Next $10$ tokens} \\
\hline
\textbf{Peter Grünberg} was born in & Munich in \textcolor{teal}{1939} and obtained & Albert Einstein {\color{teal}1879}, \textbf{Peter Grünberg} & \textcolor{red}{2007} Mohammed Hanif \\
\hline
Albert Einstein was born in {\color{teal}1879}, Brian Kobilka was born in {\color{teal}1955}, \textbf{Peter Grünberg} was born in & \textcolor{teal}{1939} and has a birthday & Albert Einstein {\color{teal}1879}, Brian Kobilka {\color{teal}1955}, Stefan W. Hell {\color{teal}1962}, Ivan Pavlov {\color{teal}1849}, \textbf{Peter Grünberg} & \textcolor{teal}{1939}, Albert Szent- \\
\hline
Albert Einstein was born in {\color{red}1872}, Brian Kobilka was born in {\color{red}1965}, \textbf{Peter Grünberg} was born in & \textcolor{teal}{1939} and has a birthday & Albert Einstein {\color{red}1872}, Brian Kobilka {\color{red}1927}, Stefan W. Hell {\color{teal}1962}, Ivan Pavlov {\color{red}1878}, \textbf{Peter Grünberg} & \textcolor{red}{2007}, Albert Szent- \\
\hline
\textcolor{brown}{Author-1} was born in {\color{brown}1872}, {\color{brown} Author-2} was born in {\color{brown}1965}, \textbf{Peter Grünberg} was born in & \textcolor{teal}{1939} and has a life expect & \textcolor{brown}{Author-2 1879, Author-3 1955, Author-4 1962, Author-1 1849}, \textbf{Peter Grünberg} & \textcolor{red}{2022}, Rossitza \\
\hline
\end{tabular}
}
}
\caption{Comparison of latent knowledge estimators for the test fact $\langle \text{Peter Grünberg}, \text{Birth Year}, \text{1939} \rangle$ using Llama2-7B. Correct years are in {\color{teal}teal}, incorrect years in {\color{red}red}, and unknown examples in {\color{brown}brown}. Author-\# represents subjects unknown to the LLM.}
\label{tab:example}
\end{table*}

\textbf{2. \iclke~ requires many-shots and is fundamentally different from few-shot prompting:}
A recent work \citep{kalo2022kamel} shows that few-shot prompting (FS-LKE in  Figure~\ref{fig:input}) can yield improved knowledge estimates compared to zero-shot prompting (ZS-LKE in  Figure~\ref{fig:input}) by providing the LLM with examples to {\it format generated answers}. In contrast, \iclke~ also uses examples to effectively {\it communicate the question at hand}. The different modes in which examples are being used in FS-LKE and \iclke~ is illustrated further in Table~\ref{tab:example}. Adding a few examples to the zero-shot prompt {\it "Peter Grünberg was born in"} results in the model generating the correct answer "1939" right-away. However, it does not appear to matter whether the examples provided are reflecting correct information or information about subjects known to the LLM (consistent with our hypothesis that the examples are being used to infer answer format). In contrast, Table~\ref{tab:example} suggests that for \iclke~ not only is the number of examples needed larger, but it also matters whether they are correct and known to the LLM (consistent with our hypothesis that examples are being used to infer the question). The two distinct ways in which examples are being used by FS-LKE and \iclke~map well to the dual modes of in-context learning namely, {\it task recognition} and {\it task learning}, respectively, that have been identified in a recent work\cite{pan2023context}.  

We systematically investigate how factors such as how many examples are provided in an \iclke, whether some of those examples are unknown to the model or simply incorrect, as well as how the examples are ordered affect knowledge estimation. We find that \iclke~requires {\it many-shots}, which make it relatively {\it robust to unknown examples}, but \iclke~remains vulnerable to {\it incorrect examples}. Our findings represent a nuanced exploration of in-context learning\cite{brown2020language}, where the dominant learning mode is task learning rather than task recognition.

\textbf{3. \iclke~significantly outperforms previous prompt-based approaches across different open-source models and different types of factual relations:}
We empirically compared the performance of \iclke~against prior approaches that relied on a variety of human-generated prompts (HGPs) as well as machine-mined prompts (MMPs)\cite{jiang2020can}. 
Across a large set of facts spanning different types of relations from the widely-used T-Rex dataset~\cite{elsahar2018t}, we find that \iclke~improves the fraction of facts accurately extracted from four open-source models by an average of 35\% for HGPs (from 0.45 to 0.61) and 90\% for MMPs (from 0.32 to 0.61).
These performance gains of \iclke~arise from a better comprehension of the question as well as the expected answer format. 
To quantify the performance gains from only better question comprehension, we propose in Section~\ref{sec:design_lkes} a multiple-choice testing that accounts for answer formats.
We find that \iclke~still outperforms existing approaches by an average of 9.41\% for HGPs (from 0.71 to 0.78) and 57\% for MMPs (from 0.50 to 0.78), with improvements for specific relations like "position played on team/specialty" varying from 152\% for HGPs (from 0.17 to 0.43) and 310\% for MMPs (from 0.10 to 0.43).
Thus, \iclke~represents a better way to retrieve knowledge stored internally within an LLM, surpassing the model's ability to follow instructions in prompt templates.

\textbf{4. Being model-agnostic, \iclke~enables a systematic comparison of latent knowledge of open source LLMs at scale:}
In contrast to prompt-based LKEs \citep{jiang2020can,shin2020autoprompt}, which are tailored to specific relations and models, \iclke~creates a single input to test for facts pertaining to a relation that can be used flexibly for any model. This simplicity and versatility allows for cross-LLM comparisions of factual knowledge.
Using \iclke, we evaluated the knowledge of 49 open-source LLMs from various families like Llama(2), Gemma, Mistral, OPT, and Pythia. These models vary in size and were tested with and without instruction-finetuning on 50 different relations and 20,000 facts from Wikidata. We found that models from families such as Llama2, Mistral, and Gemma, as well as larger models, know more facts. Models within the same family differ in the specific facts they know, even if trained on the same data. 
Additionally, instruction fine-tuning reduces the amount of factual knowledge that can be extracted from these models.
Our findings will likely be of interest to developers that wish to train models with lots of embedded factual knowledge.

\textbf{Related Work:}
Researchers have proposed several approaches \citep{youssef2023give} to estimate latent knowledge from LLMs, which can be categorized into two main methods:
(i) Model-internals based approaches: These methods use various internal aspects of the LLM, such as attention maps \cite{wang2020language}, activation functions \cite{burns_discovering_2022}, or model parameters \cite{kazemnejad2023measuring}, to determine whether factual information can be extracted from the model.
(ii) Model-responses-based approaches, which are generally applicable to a wide range of LLM models and there are two key parts of the model-responses-based approach: constructing the input and evaluating the output from the LLM.

{\it Input construction:} There are different prompting techniques to verify if a target fact is stored in the model. These prompt-based methods differ in their choice of prompts, which can be divided into human-generated prompts (HGPs) \cite{cao2021knowledgeable, chern_factool_2023,sun_head--tail_2023,wang2020language,petroni2019language,jiang2021can,newman2022padapters,jiang2020can,kalo2022kamel} and machine-mined prompts (MMPs) \cite{shin2020autoprompt,jiang2020can}. All the prompting-based methods try to find the best template for the question that the model can comprehend best; however, the optimization of the prompting template can be unreliable \cite{zamfirescu2023johnny, DBLP:conf/iclr/AroraNCOGBCR23,sclar2023quantifying, cao2021knowledgeable}. Instead of finding the best comprehensive template, our approach proposes using structured triplet examples to communicate and prob the tested fact, uncover deeper relations and knowledge in the LLM. This method communicates the question through in-context examples, a strategy that, to our knowledge, has not been explored before. A similar approach is to use few-shot prompting. For example, \cite{kalo2022kamel} first found the best prompt template and then applied few-shot prompting to guide and limit the model's response format. However, this approach still holds the limitation of template searching and relies on the model's comprehension of the template, which is fundamentally different from our approach. 

{\it Output evaluation:} 
The evaluation methods of early works are LLM specific,
limiting the evaluated objects to single-token outputs \cite{bouraoui2020inducing, jiang2020can, shin2020autoprompt, petroni2019language}. 
More recent works evaluate the generation by checking the next $k$ generated tokens to see whether the potentially multi-token ground truth appears in the $k$ generated tokens \cite{cao2021knowledgeable, chern_factool_2023,sun_head--tail_2023,wang2020language,jiang2021can,newman2022padapters, kalo2022kamel}. However, the final performance is significantly influenced by the choice of $k$, and the generation quality also relies heavily on various sampling parameters, which introduces uncertainty \cite{lin2023generating}. In order to respond more fundamentally to the model's level of knowledge, without focusing on metrics such as the fluency of generation, we constructed a multiple-choice dataset with 100 unique possible choices for each evaluated fact and judged whether or not the model knew the fact by comparing the relative probabilities of these 100 objects. 

{\it Factual knowledge datasets:} Different from existing knowledge evaluation benchmarks like TruthfulQA \cite{lin2021truthfulqa} and MMLU \cite{hendrycks2020measuring}, which already provide templates of questions, our approach considers facts from existing knowledge graphs for performing knowledge estimation of LLMs. As a test bed \cite{elsahar2018t,hu2023large,sun_head--tail_2023,petroni2019language,zhu2023physics,kryscinski_evaluating_2019}, we utilize knowledge graphs, allowing our method to be applied to any knowledge graph database without additional effort.

\section{Designing Reliable LKEs}

Today, there exist many general-purpose as well as domain-specific factual knowledge bases that contain a very large number (millions to billions) of facts. The facts can be encapsulated as triplets, represented as  $\langle$\textit{subject ($\subject$), relation ($\relation$), object ($\object$)}$\rangle$. These triplets offer a general way to represent factual knowledge about real-world entities in knowledge graphs or other structured knowledge bases. The goal of latent knowledge estimation is to infer what fraction of the facts are {\it known} to an LLM.
We call methods that estimate the amount of latent knowledge inside an LLM \textit{latent knowledge estimators} (LKEs).

\subsection{Reliability concerns with existing LKEs}
\label{sec:reliability_concerns}

Existing approaches to estimating latent knowledge in LLMs use various factual knowledge tests. We identify several reliability concerns (RCs) with current designs that motivate our new LKE design. While some related works address some of these concerns, none have comprehensively solved all the issues~\cite{kalo2022kamel,cao2021knowledgeable}.

{\bf RC 1.} {\it Reliance on unrestricted prompt engineering:} 
Many past works have attempted to use test prompts without any restrictions, including both human-generated or machine-mined prompts
~\citep{jiang2020can,zamfirescu2023johnny,DBLP:conf/iclr/AroraNCOGBCR23,sclar2023quantifying}.
They typically intersperse the subject $\subject$ and object $\object$ between additional relationship context-communicating tokens. 
Some analyze the performance of a variety of prompts and then pick the best-performing or use an ensemble of the best-performing prompts~\citep{jiang2020can,newman2022padapters, fernando2023promptbreeder}.
However, unrestricted prompt engineering risks introducing side-channels and over-fitting. First, the generated prompts, particularly those that are machine-mined, may include tokens that can implicitly or explicitly introduce additional (side-channel) information that makes it easier to answer the question. As a specific example, in a prior work~\cite{jiang2020can}, for the relation {\it ``position held"}, the prompt {\it ``$\subject$ has the position of $\object$"} performed worse than {\it ``$\subject$ is elected $\object$"}. But, note that the second prompt potentially introduces a side-channel: it implicitly rules out answer choices for unelected positions like Professor and favors elected positions like President. Second, 
selecting from an unbounded number of potential prompt choices raises concerns about the complexity of LKEs (the size of the set of all considered prompts) and the associated risks of over-fitting, which in turn affect the reliability of estimates. 

{\bf RC 2.} {\it Reliance on LLMs' meta-linguistic judgments:}
Prior works used prompt templates with instructions~\cite{chern_factool_2023,sun_head--tail_2023,wang2020language,petroni2019language,jiang2021can,newman2022padapters,jiang2020can, cao2021knowledgeable, kalo2022kamel, youssef2023give, zhao2024matters} for communicating the question as well as the expected format of answers. But, the scores (estimates) resulting from such prompt-based testing conflate an LLM's latent knowledge of the facts with the LLM's meta-linguistic judgments, i.e., the LLM's ability to comprehend the prompt, understand the question embedded within the prompt and output the answer in some expected format~\citep{hu2023prompting}.
The impact on meta-linguistic judgments can be seen from the fact that multiple semantically-equivalent prompts result in different responses from an LLM and thereby, different estimates of latent knowledge~\citep{hu2023prompting}. 

{\bf RC 3.} {\it Reliance on LLM-specific prompts:} 
Many prior works~\cite{petroni2019language,jiang2020can,shin2020autoprompt} limit the choice of facts that can be used in tests to those where the surface form of the objects ($\object$) is represented by a single token by the LLM's tokenizer. 
Even though some works are able to evaluate multiple-token objects, prompt-based approaches need careful prompt engineering for different LLMs to get the best prompt template~\cite{cao2021knowledgeable, chern_factool_2023,sun_head--tail_2023,wang2020language,jiang2021can,newman2022padapters, kalo2022kamel}, which makes it hard to estimate and compare factual knowledge across a large number of LLMs and the prompt optimisation would be very expensive and inefficient for large models.

Motivated by the above, we derive the following three design principles (DPs) for LKEs. A reliable LKE design should:
\begin{itemize}
\item \textbf{DP1:} {\it limit prompt hacking to avoid over-fitting \& side-channels}.
\item \textbf{DP2:} {\it minimize reliance on meta-linguistic prompts}. 
\item \textbf{DP3:} {\it avoid LLM-specific prompts}.
\end{itemize}

\subsection{A new \textcolor{red}{Z}ero- \textcolor{red}{P}rompt 
based \textcolor{red}{LKE} (\textcolor{red}{\iclke})}
\label{sec:design_lkes}

Our goal is to estimate whether an LLM knows a fact $\fact = \langle \subject, \relation, \object \rangle$.
The challenge is to probe the LLM and evaluate its responses in a way compatible with the design principles defined in Section~\ref{sec:reliability_concerns}.

The key idea here is to {\bf eliminate prompts meant to capture the relation $\relation$ (\textit{zero-prompt})} and instead {\bf rely on examples of similarly related $\langle \subject,\object \rangle$ pairs} to probe the internal knowledge.
LLMs have been shown to exhibit In-Context Learning (ICL) abilities~\citep{brown2020language} that allow them to infer and extrapolate patterns in their inputs.
We leverage this ability to communicate information about relation $\relation$ without additional instructions to the LLMs (DP1 and DP2) by providing it with a list of facts based on $\relation$.

\begin{example}
\label{example:iclke}
Assume that we want to probe for whether an LLM knows the fact \textit{$\langle$ Einstein, birth-year, 1879 $\rangle$}.
We can use other facts for the birth-year relation such as \textit{$\langle$ Feynman, birth-year, 1918 $\rangle, \langle$ Heisenberg, birth-year, 1901 $ \rangle$} to construct an input \textit{``Feynman 1918 Heisenberg 1901 Einstein''}.
By providing such zero-prompt in-context examples to the model, we expect to communicate the underlying relation between subjects and objects.
To correctly extrapolate the pattern, the model needs to retrieve Einstein's birth-year as the completion of the sequence.
\end{example}

More formally, given a training dataset of facts $\setOfFacts_{\relation} = \{\langle \subject_i, \relation, \object_i \rangle\}_{i=1}^n$ for relation $\relation$, as well as a test fact $\fact = \langle \subject, \relation, \object \rangle$, we leverage ICL to construct prompts that elicit information about $\fact$ as
\begin{equation}\label{eq:prompt}
    \prompt(\subject, \relation) = \subject_1 \, \object_1 \, \dots \, \subject_n \, \object_n \, \subject
\end{equation}
We use $\relation$ to pick facts from $\setOfFacts_{\relation}$ and concatenate the tokens corresponding to the subjects and objects, but do not include any other information about $\relation$.
We use space `` '' as the separator token and discuss this choice in detail in Section~\ref{section:compare with baseline}.
We discuss other design choices for the construction of ZP-LKE in Section~\ref{sec:icl_for_lke}. When further details are not needed, we simply refer to \textit{some} input as $\prompt$.

\textbf{\iclke~design satisfies all our design principles (DPs):}
\begin{itemize}
    \item \textbf{DP1:} by construction, zero-prompting eliminates prompt hacking and thus, risks of over-fitting and side-channels. 
    \item \textbf{DP2:} it only relies on the in-context learning abilities and not meta-linguistic judgments of an LLM.
    \item \textbf{DP3:} by construction, input $\prompt(\subject, \relation)$ is LLM-agnostic and hence, enables cross-LLM latent knowledge comparisons.
\end{itemize}

\subsection{Evaluating model outputs} 
We evaluate the output of model $\llm$ for input $\prompt(\subject, \relation)$ 
in two ways: (1) Open-ended generation that lets the model generate till $k$ tokens  ~\cite{kalo2022kamel,yu2024kola} after which the presence of the ground truth is checked within the response, (2) Multiple-choice test that forces the model to predict from a list of options \cite{jiang2020can}.


\textbf{(1) Response testing in open-ended generation.} Given a fact \( f = \langle \subject, \relation, \object^* \rangle \) and a model \( \llm \), we provide the input \( \prompt(\subject, \relation) \) to the model and let it generate for $k$ tokens $t_1, t_2,..t_k$. We consider the answer to be correct if $y^* \subseteq \{t_1,t_2,...,t_k\}$ leading to the prediction $pred_{\theta}(f) = y^*$.





\textbf{(2) Multiple-choice testing.}
In the multiple-choice testing, we extract the answer based on the probabilities $\llm$ assigned to the tokens of the corresponding object $\object$.
To allow for objects $\object $ consisting of multiple tokens and to be independent of the specific tokenization scheme or LLM (DP3), we compute the \textit{object probability} over multiple tokens as follows:
\begin{equation}
    P_{\llm}(\object \mid \prefix) = \prod_{i=2}^{|\object|} P_{\llm}(\object^{(i)} \mid \object^{[i-1:1]} \, \prefix) \cdot P_{\llm}(\object^{(1)} \mid \prefix)
\end{equation}

where $|\object|$ denotes the number of tokens in $\object$ and $P_{\llm}(\object^{(i)} \mid \object^{[i-1:1]} \, \prefix)$ is the conditional probability of predicting the $i$-th token $\object^{(i)}$ of $\object$ given the preceding tokens $\object^{(i-1)}, \dots, \object^{(1)}$, and $\prefix$.
To determine whether model $\llm$ knows a fact $f = \langle \subject, \relation, \object^* \rangle$, we test whether given an input $\prompt(\subject, \relation)$, $\llm$ can choose the correct object $\object^*$ from among a set of $M$ unique alternatives.
Specifically, given fact $f$, we redefine it as $f = \langle \subject, \relation, \object^*, \setOfObjects \rangle$, where $\setOfObjects$ is a set of $M$ plausible but incorrect alternatives.
We discuss the choice of $\setOfObjects$ in Section~\ref{sec:results}.
\begin{equation}
\label{eq:correct_prediction}
\pred_{\llm}(f) \triangleq \underset{\object \, \in \, \{ \object^* \} \, \cup \, \setOfObjects}{\mathrm{argmax}} \, P_{\llm}(\object \mid \prompt(\subject, \relation))
\end{equation}
denotes the prediction of $\llm$ for the fact $f = \langle \subject, \relation, \object^*, \setOfObjects \rangle$.
The predicted object has the maximal object probability within $\{\object^*\} \cup \setOfObjects$.

We evaluate the factual knowledge of model $\llm$ over a dataset of test facts $\dataset = \{f_i\}_{i=1}^m$ using accuracy as a metric for both the response test and multiple-choice test:
\begin{equation}\label{eq:accuracy_main}
\begin{split}
    \acc(\llm, \dataset) \triangleq
    \frac{\sum_{f \in \dataset} \delta \left(\object^* = \pred_{\llm}(f) \right)}{|\dataset|}
\end{split}
\end{equation}

where \(\delta(\cdot)\) is the indicator function.

\section{Exploring the design space of \iclke}
\label{sec:icl_for_lke}

\begin{figure}
    \centering
    \includegraphics[width=0.4\textwidth]{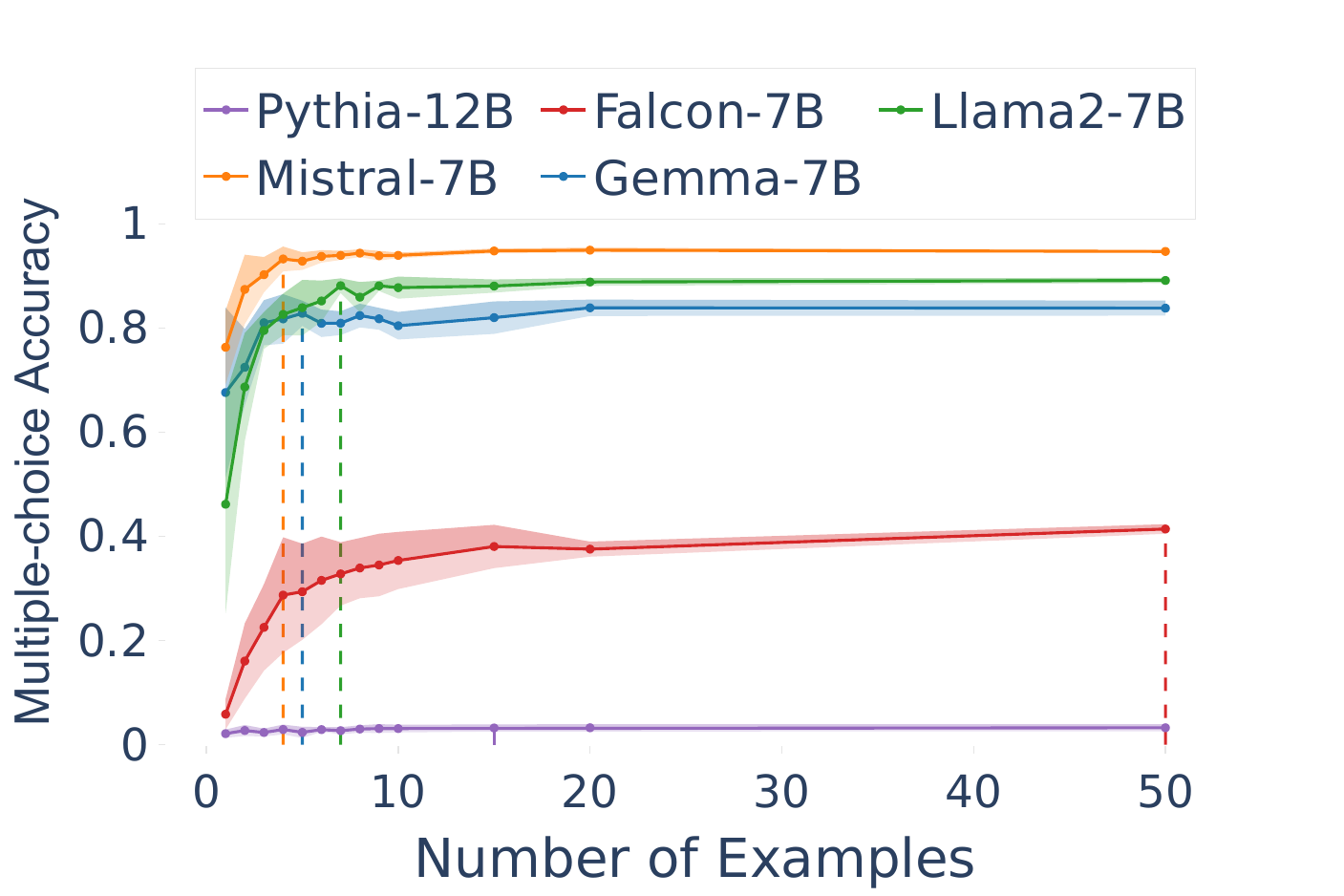}
    \caption{\change{Impact of in-context example count on multiple-choice accuracy across LLMs. The dashed line marks the number needed for 95\% stable accuracy with 50 examples. 
    }}
    \label{fig:num-incontext}
\end{figure}

\begin{figure}
\centering
\subfloat[(Subject, object) examples in a prompt]{%
  \includegraphics[width=0.3\textwidth]{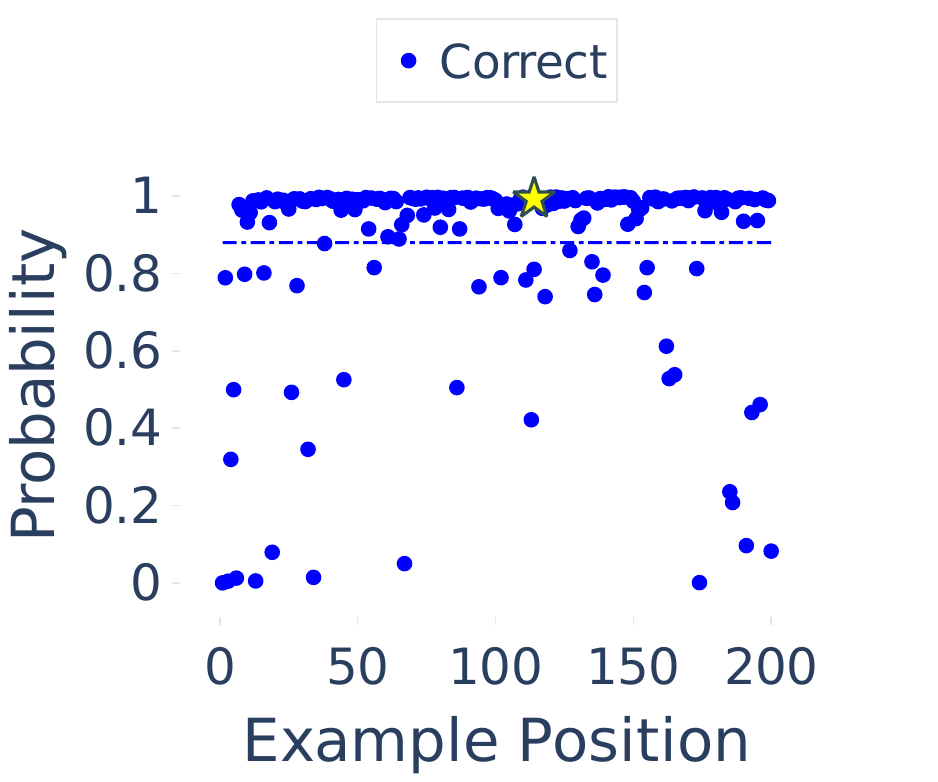}%
  \label{fig:correct_sequence}%
}
\\
\subfloat[Distributed unknown examples]{%
  \includegraphics[width=0.25\textwidth]{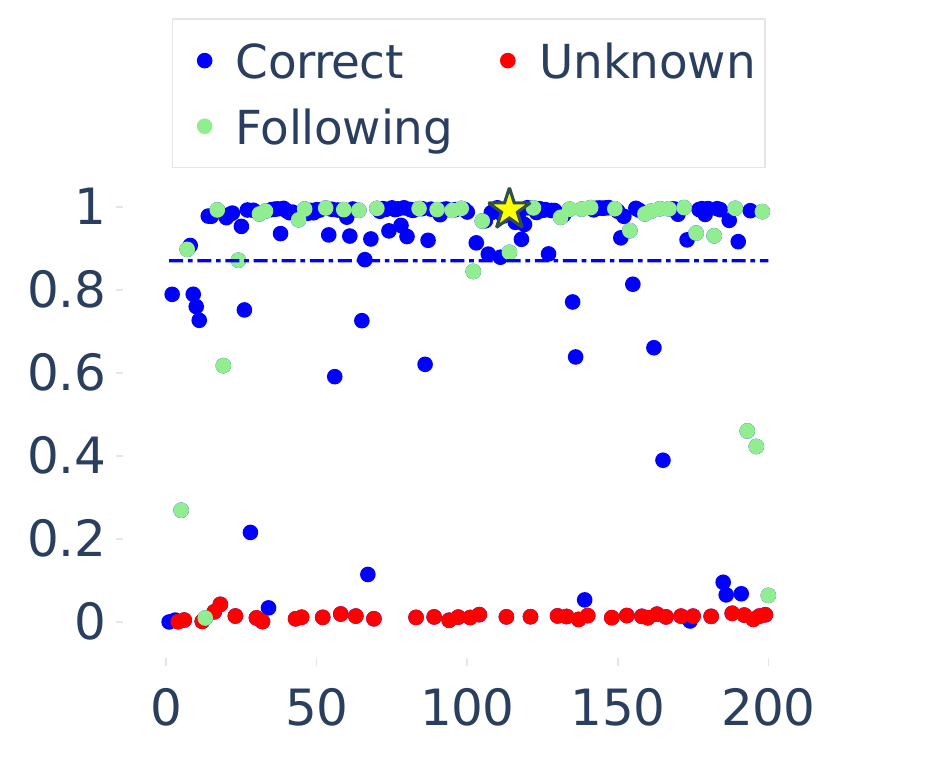}%
  \label{fig:unknown_distributed}%
}
\subfloat[Continuous unknown examples]{%
  \includegraphics[width=0.25\textwidth]{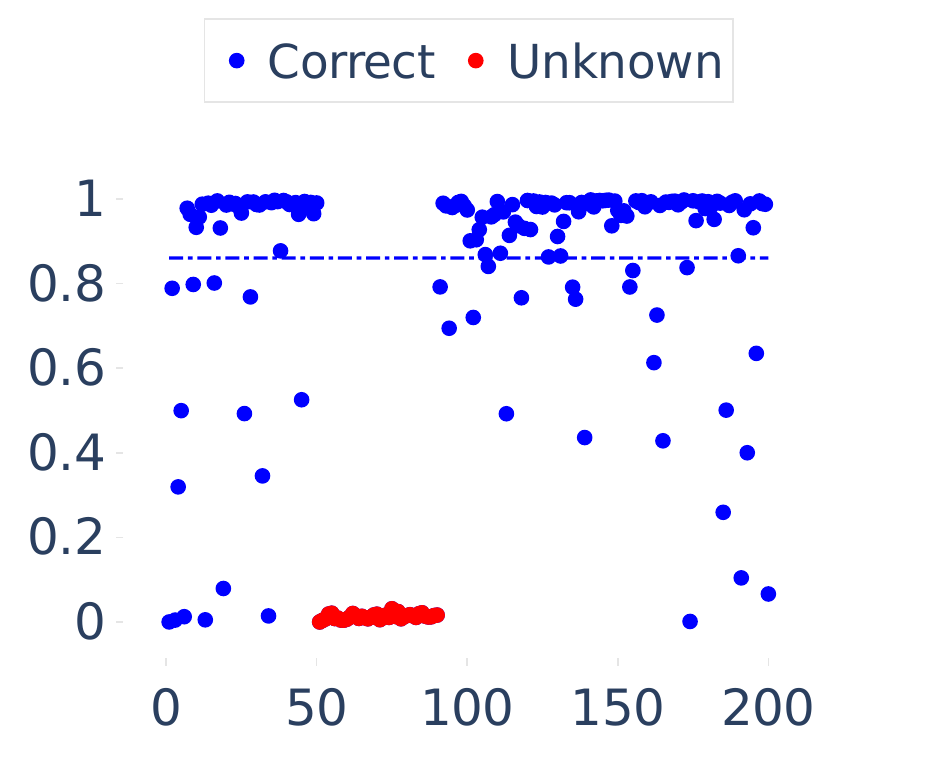}%
  \label{fig:unknown_simultaneous}%
}
\\

\subfloat[Distributed incorrect examples]{%
  \includegraphics[width=0.25\textwidth]{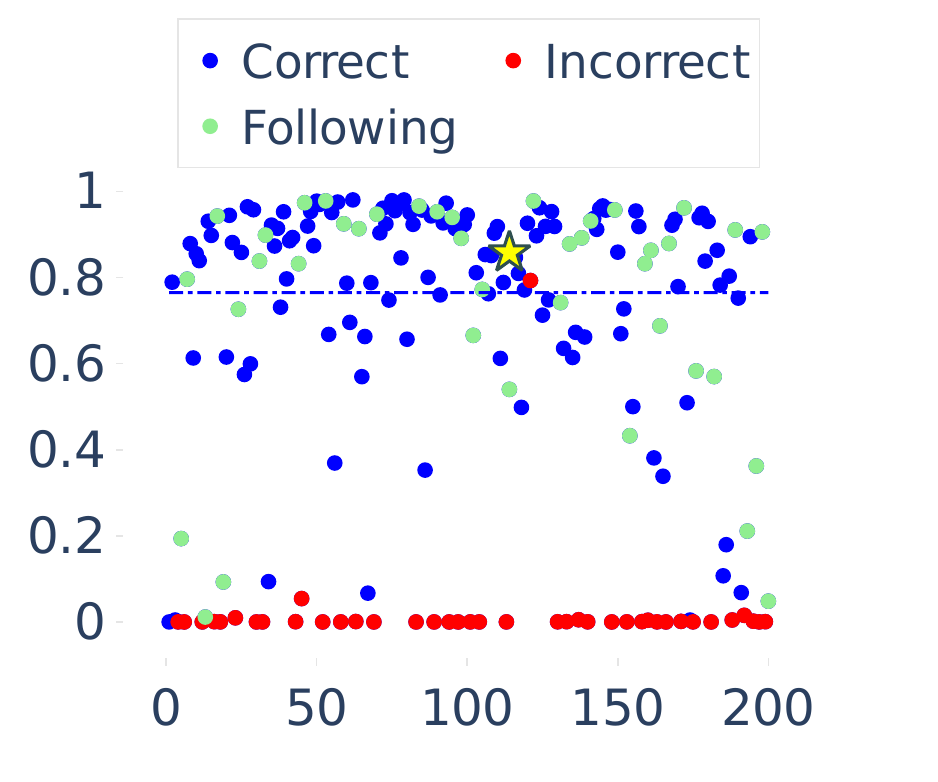}%
  \label{fig:incorrect_distributed}%
}
\subfloat[Continuous incorrect examples]{%
  \includegraphics[width=0.25\textwidth]{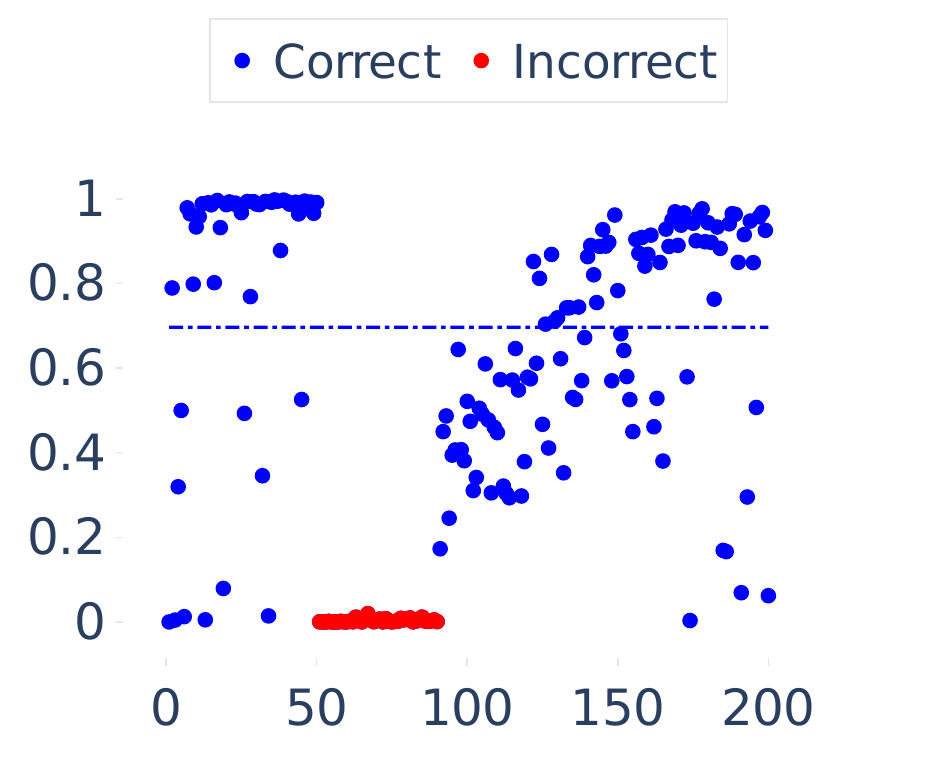}%
  \label{fig:incorrect_simultaneous}%
}

\caption{\change{Variation in Nobel laureate data probabilities using Mistral-7B. 
Figure~\ref{fig:correct_sequence} illustrates object probabilities at various positions in the prompt. Figures~\ref{fig:unknown_distributed} and~\ref{fig:unknown_simultaneous} show impacts of unknown objects at random and continuous positions, while Figures~\ref{fig:incorrect_distributed} and~\ref{fig:incorrect_simultaneous} show effects of incorrect examples. The dashed line indicates average correct probabilities (blue dots).}}
\label{fig:mistral-7b-probabilities}
\end{figure}



Our \iclke~design avoids many of the reliability concerns of prior works \citep{jiang2020can, kalo2022kamel, yu2024kola, hogan2021knowledge,cao2021knowledgeable,petroni2019language}. 
However, \iclke~also introduce a few design choices for the input i.e., $\prompt(\subject, \relation)$ in Equation~\eqref{eq:prompt}. One must decide the right $n$, the number of in-context examples included in $\prompt(\subject, \relation)$. Furthermore, it is unclear how \iclke~would be affected if some chosen examples are unknown to the model or are incorrect or appeared in a different order. 

We study this by varying 
 $n$ and introducing unknown or incorrect examples within these $n$ examples.
 While many prior works investigated the number of in-context examples needed for various tasks~\cite{brown2020language, agarwal2024many, chen2023many, lin2024dual, pan2023context}, it is worth re-examining them for \iclke~for three reasons: (i) prior works report differing results, with some reporting that increasing the number of examples improves performance~\cite{agarwal2024many}, while others argue the opposite~\cite{lin2024dual}, (ii) most don't carefully distinguish between the two learning modes of in-context learning (as noted in Section~\ref{sec:intro}, \iclke~relies on one of the modes), and (iii) only a few~\cite{min2022rethinking,chen2023many} studied the influence of incorrect examples and none studied the impact of unknown examples.  
 
\change{Our experiments help us understand the number of in-context examples needed, as well as how the in-context example's generation probability changes with different types of noise.}
We perform an in-depth empirical analysis on a \textit{Nobel Laureate} dataset for the relation \textit{`birth year'} (details in Appendix~\ref{appendix:nobel_dataset}). The dataset consists of facts formatted as $\langle \text{Person} (\subject), \text{birth-year} (\relation), \text{YYYY} (\object) \rangle$.

\textbf{The number of required in-context samples for communicating both the question and answer format varies across LLMs.}
In Figure~\ref{fig:num-incontext}, we report multiple choice accuracy (Eq.~\eqref{eq:accuracy_main}) for different LLMs evaluated on 900 test samples, with varying numbers of in-context examples ($n$) that are randomly sampled from a separate training set using 5 random seeds.
As the number of in-context examples increases, the mean accuracy rises while the standard deviation decreases across different LLMs, indicating that the models gradually converge to stable performance. 

The dashed vertical lines show the minimum number of examples required by different LLMs to achieve 95\% of the accuracy reached with 50 in-context examples.
Interestingly, LLMs with higher estimation accuracy require fewer in-context examples than those with lower accuracy to effectively interpret the underlying question. This maybe attributed to the amount of internal knowledge contained in the LLMs.
To enable \iclke~across all the LLMs, we set $n=50$ for the following experiments.

We delve deeper to further investigate which individual facts may be known or unknown to a model. We examine the generation probability of in-context objects in 200 correct subject ($\subject$)-object ($\object$) pairs using the Mistral-7B model. 
We can see in Figure~\ref{fig:correct_sequence}that Mistral-7B model shows a gradual increase in the probability of generating correct objects from left to right on the x-axis (where points on the right have more context to leverage)
stabilizing at a mean probability of approximately 0.85. Some objects at later positions, however, have a lower generation probability,
suggesting that the LLM may be less confident about its knowledge of the facts corresponding to those pairs. Thus, we can leverage the generation probability as a signal of the LLM's confidence when evaluating LKEs (see Appendix~\ref{appendix:diff_metrics}). Similar results for additional models are presented in Appendix~\ref{appendix:icl}.

\textbf{Models are robust to unknown examples.} 
Next, we investigate the robustness of estimates to the occurrence of unknown examples. We insert unknown examples in two distinct ways: randomly distributed throughout $\prompt(\subject, \relation)$ and in a more extreme scenario, where a continuous block of examples is replaced with unknown ones. We selected 40 out of the 200 examples and replaced them with unknown examples created using fictitious names and birth years~\footnote{generated via \url{https://en.namefake.com/api}}.
Our findings are shown in Figures~\ref{fig:unknown_distributed} and \ref{fig:unknown_simultaneous} for distributed and continuous replacement respectively. Unknown examples are marked by \textcolor{red}{red} dots, examples immediately following unknown ones in \textcolor{cyan}{cyan} dots and the rest in \textcolor{blue}{blue} dots. The unknown examples show generation probabilities close to zero, confirming the LLM's tendency to assign low probabilities to unknown data. However, interestingly, \textit{unknown examples minimally impact the generation probability of the surrounding data in both settings}.

\textbf{Models are vulnerable to incorrect examples.} 
Similar to the setup for unknown examples, we also insert 40 (out of 200) incorrect examples randomly (Figure~\ref{fig:incorrect_distributed}) and simultaneously (Figure~\ref{fig:incorrect_simultaneous}). In our experiments, these incorrect examples are created by altering the birth years of known Nobel laureates and are marked by \textcolor{red}{red} dots in the plots. 
\textit{In contrast to inserting unknown examples, the LLM significantly struggles with the injection of incorrect examples.} It detrimentally affects the LLM's performance in both settings thus revealing the vulnerability of the models towards incorrect examples. We highlight one randomly marked \textcolor{yellow}{yellow} star example in Figure~\ref{fig:correct_sequence}, Figure~\ref{fig:unknown_distributed}, and Figure~\ref{fig:incorrect_distributed} to show how the presence of incorrect samples significantly brings down the probability of the neighboring points.

\textbf{Summary:} 
The key takeaways while exploring the design space of \iclke~ are - (a) Different LLMs take varying numbers of in-context samples to comprehend both the question and format of the answer alongside, with $50$ being an optimal number for our setup. (b) The models are robust to unknown examples but vulnerable to incorrect examples. 
To the best of our knowledge, we are the first to \textit{distinguish between examples unknown to the model and incorrect examples known to the model} and study their impact on in-context learning. As \iclke~relies on many examples, this distinction and understanding is important in practice.
Also, while ~\cite{agarwal2024many} found that the order of examples has varying effects in different domains, we identify the distribution of unknown and incorrect examples as a crucial underlying factor.

\section{Experiments and Results}
\label{sec:results}
As \iclke~ inputs are model-agnostic and easy to adapt for a large variety of relations, it can be used to very effectively to conduct cross-LLM latent knowledge comparisons.
We 
leverage {\iclke} to estimate latent knowledge across 49 open-source (pre-trained and fine-tuned) LLMs, spanning different LLM families (Llama (2), Mistral, Mixtral, Gemma, Falcon, Pythia, Bloom, and OPT) and sizes (from 70M to 8$\times$22B). To the best of our knowledge, we are the first to evaluate a knowledge estimation framework across a large number of models. We list the models and their simplified names used in this paper in Appendix~\ref{appendix:all_eval_results}, Table~\ref{table:all_models_name}, and provide a leaderboard of models based on \iclke~in Appendix~\ref{appendix:all_eval_results}, Table~\ref{table:model_order}. We hope our results and the framework can help future LLM developers reliably and efficiently estimate the latent knowledge of their models.


\textbf{Dataset:}
We evaluate the knowledge of models on a large set of facts 
from the \trex~dataset~\cite{elsahar2018t}. 
We selected relations from \trex~with at least 500 samples that are linked to a minimum of 100 unique objects. We create a list of multiple choices for each sample and ensure that instances with multiple correct objects do not have any of their correct answers in their multiple-choice list.
This filtering leads to 50 distinct relations spanning categories like birth dates, directorial roles, parental relationships, and educational lineage. 
The resulting \trex~Multiple Choice (\trexmc) dataset comprises 5,000 training and 20,000 test facts. Appendix~\ref{appendix:dataset_construct} contains detailed information on the dataset and relations.

\textbf{Choosing the set $\setOfObjects$ \& its impact on test difficulty:}
For each fact $\langle$\textit{subject ($\subject$), relation ($\relation$), object ($\object^*$)}$\rangle$, we generate alternative objects $\setOfObjects$ to create multiple choices.
Note that the alternative objects in $\setOfObjects$ are viable choices and cannot be easily eliminated.
Therefore, for each fact $\langle \subject, \relation, \object^* \rangle$ we select $y \in \setOfObjects$ from other facts in the dataset that share the same relationship $\relation$.
For computational feasibility, we sample $|\setOfObjects| = 99$ alternative objects per fact, so that a random guess between $\{ \object^* \} \cup \setOfObjects$ has a $0.01$ probability of being correct.

\subsection{\iclke~ vs. prompt-based approaches}
\label{section:compare with baseline}

\begin{figure}
\centering
\subfloat[Response Accuracy]{%
  \includegraphics[width=0.33\textwidth]{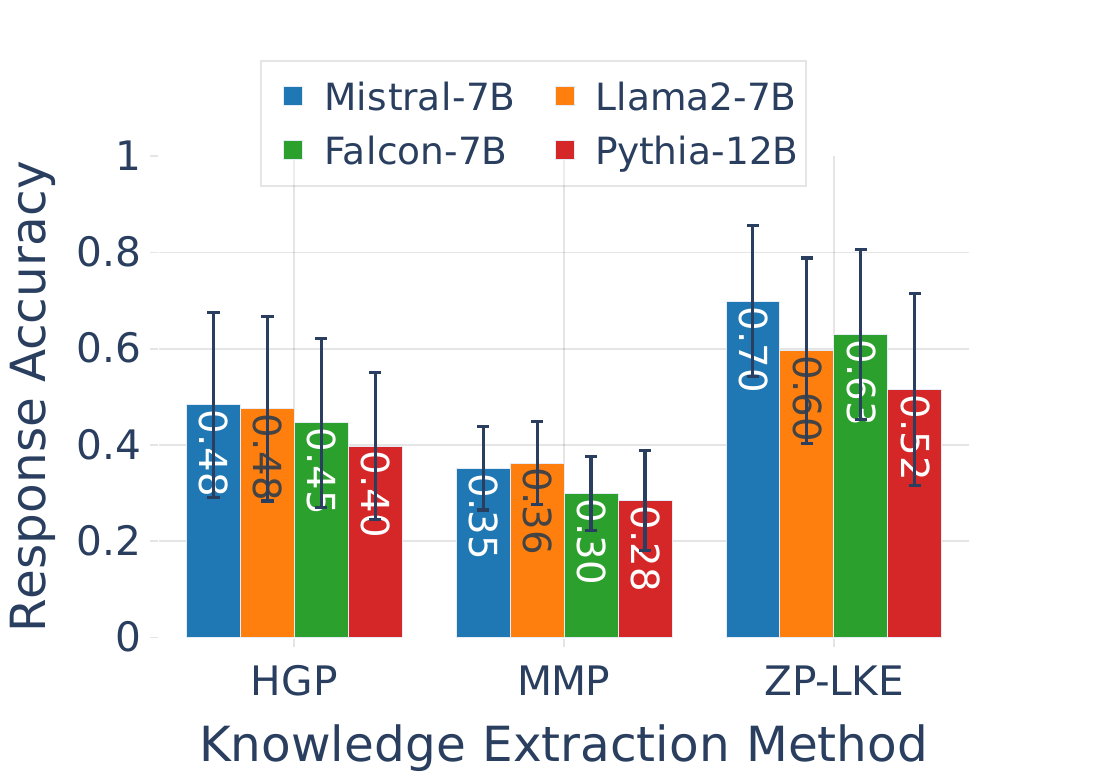}%
  \label{fig:gen_acc}%
}
\\
\subfloat[Multiple-choice Accuracy]{%
\includegraphics[width=0.33\textwidth]{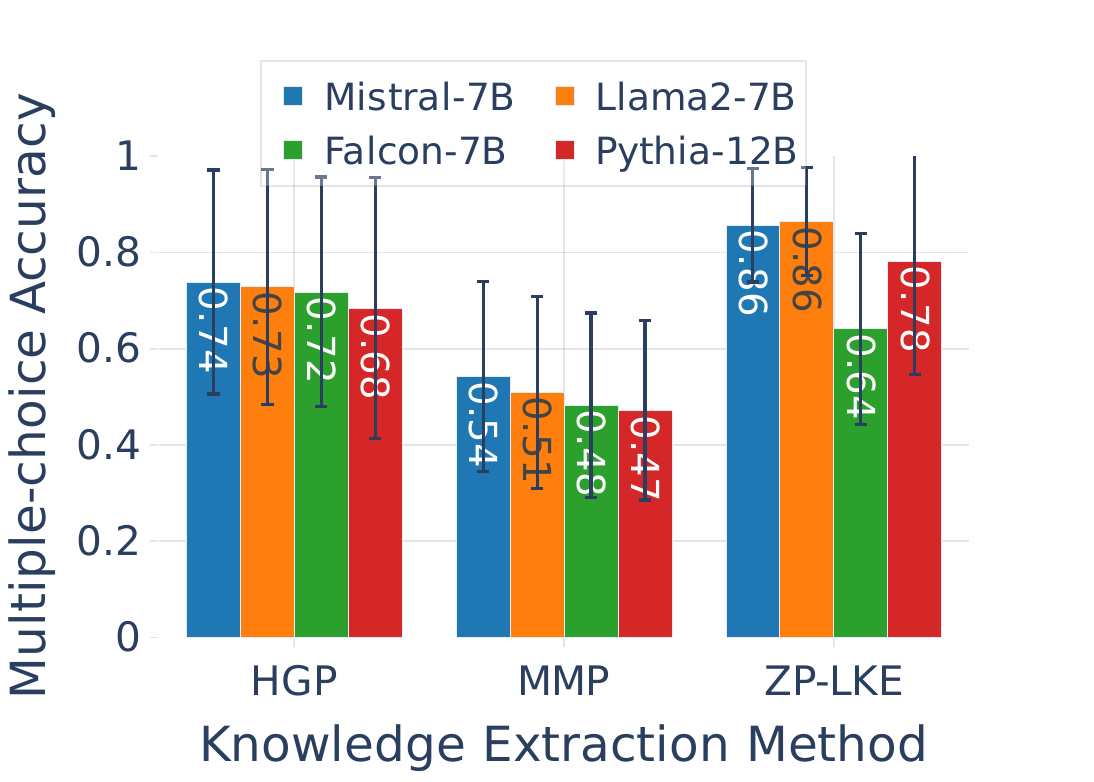}%
  \label{fig:mul_acc}%
}
    \caption{\change{Comparison of LKEs using response and multiple-choice accuracy across 12 relations from \trexmc. \iclke~is evaluated against the baseline method \citep{jiang2020can}.}}
    \label{fig:comparision_with_baselines_combined}
\end{figure}
We compare the performance of \iclke~ with the existing prompt-based approaches ~\citep{jiang2020can} using both the response accuracy and the multiple-choice accuracy defined in Section \ref{sec:design_lkes}. 

\textbf{\iclke~outperforms prompt-based approaches.}
We randomly sample three human-generated prompts (HGP) and machine-mined prompts (MMP) from \citep{jiang2020can} for 12 common relations between {\trexmc} and \citep{jiang2020can}.
We show that \iclke~outperforms HGP and MMP in terms of the accuracy measures by a large margin, across different models and 12 relations in Figure~\ref{fig:comparision_with_baselines_combined}; the detailed accuracy for each relation can be found in the Appendix~\ref{appendix:baseline_comp}, Figure~\ref{fig:respons_acc_all} and ~\ref{fig:mul_acc_all}.

Figure \ref{fig:gen_acc} shows that \iclke~improves the fraction of facts accurately extracted from four open-source models by an average of 35\% for HGPs (from 0.45 to 0.61) and 90\% for MMPs (from 0.32 to 0.61). In the case of multiple-choice accuracy, having controlled the influence of the answer format, we observe that all the knowledge estimation methods improve in their performance. Alongside, \iclke~still outperforms existing approaches by an average of 9.41\% for HGPs (from 0.71 to 0.78) and 57\% for MMPs (from 0.50 to 0.78). The multiple-choice accuracy metric disentangles the answer format from the question leading to better factual knowledge estimation across the board. Hence, we primarily report the multiple-choice accuracy metric for the experiments in the rest of the paper.


\begin{figure}
    \centering
    \includegraphics[width=0.4\textwidth]{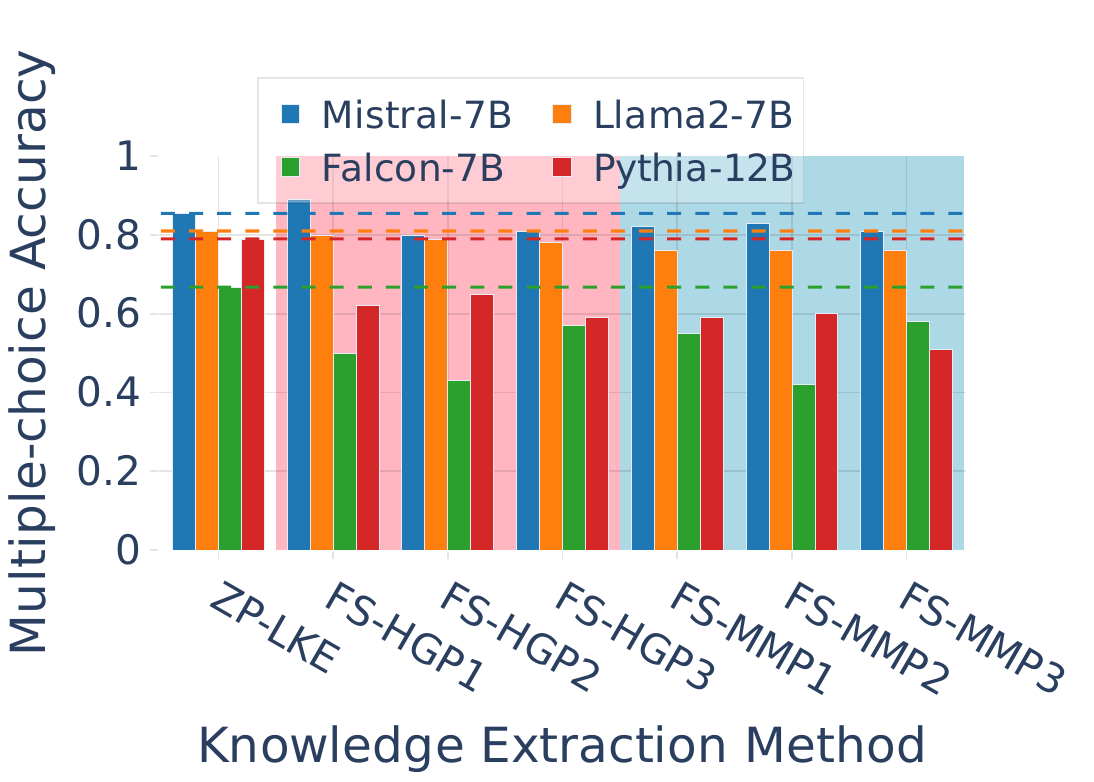}
    \caption{Impact of separators on the relation 'original broadcaster.' Subject-object pairs are separated by human-generated prompts (HGP, \textcolor{red}{red} background) or machine-mined prompts (MMP, \textcolor{blue}{blue} background).
    }
\label{fig:diff_in_between}
\end{figure}


\textbf{\iclke~ performs better than FS-LKE with the same number of examples.} 
We adapt {\iclke} by replacing the separator token `` '' between subjects and objects with three prompts each from HGPs and MMPs for the relation `original broadcaster' and report the multiple choice accuracy in Figure \ref{fig:diff_in_between}. We intend to understand whether the additional prompt tokens can help in communicating the question better.  The \iclke~ with `` '' token performs equally or even better for some models compared to the semantically meaningful prompts from HGP and MMP, which now correspond to the FS-LKE. Thus, \textit{relation-specific separators (or prompts) have limited impact on factual knowledge estimation if subject-object pairs are correctly presented}. Additionally, finding relation-specific prompts requires hand-crafted efforts or additional computation~\cite {shin2020autoprompt}, unlike our zero-prompt many-shot approach using (subject, object) pairs. \textit{Therefore, {\iclke} can potentially extend to any fact from knowledge graphs over any LLM, while HGPs and MMPs require additional supervision and relation-specific validation.}
\subsection{Evaluating Diverse Models and Relations}
\label{subsec:evaluation }
We investigate the performance of 35 pre-trained LLMs and 14 fine-tuned LLMs across 50 relations using the \iclke~framework.
Our analysis aims to uncover nuanced insights into the knowledge levels  within these models. We will examine the results through two primary lenses: (1) the variations in knowledge across different model families, and (2) the influence of model size and fine-tuning within the same model family on their knowledge attributes.

\subsubsection{Comparing different LLM families}

\begin{figure}
    \centering
    \includegraphics[width=0.44\textwidth]{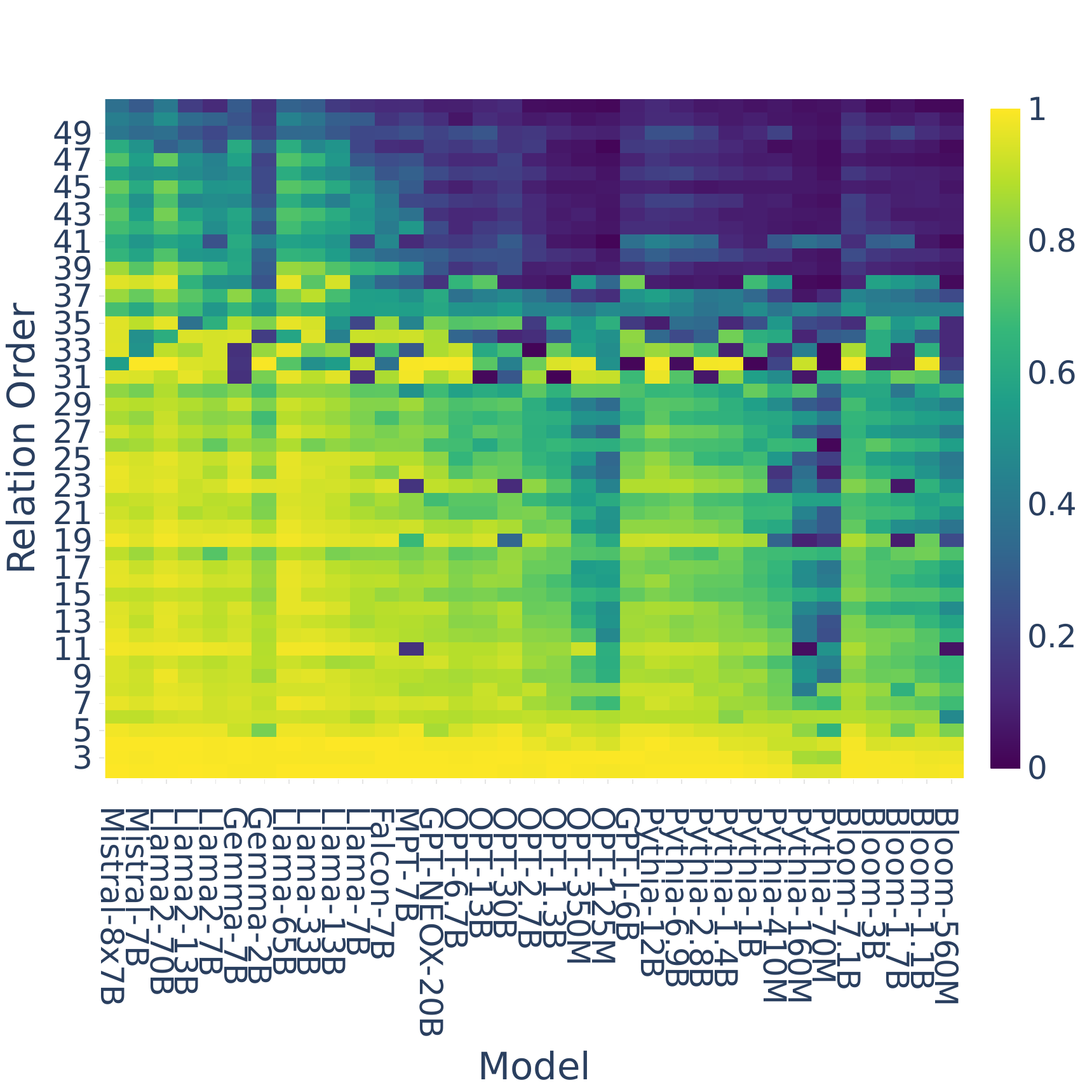}
    \caption{\textbf{Multiple-choice accuracy for 35 pre-trained LLMs on 50 relations from \trexmc.} 
 Models are grouped by family, ordered by their average accuracy, and arranged from left to right based on proximity to 7 billion parameters. Within each family, models are ordered by their average accuracy.
    }
    
    
    \label{fig:accuracy_across_all}
\end{figure}

\textbf{Some model families are consistently more knowledgeable than the rest.}
We sort the model families based on the performance of the model closest to 7B parameters~\footnote{7B parameters is a good reference point since all model families except GPT-NEO-X have models within a gap of $\leq$ 1B parameters: Mistral-7B, Gemma-7B, Llama-7B, Falcon-7B, MPT-7B, OPT-6.7B, GPT-J-6B, Pythia-6.9B, and Bloom-7.1B.}, and the models within each family based on average accuracy across 50 relations.
Figure \ref{fig:accuracy_across_all} shows that Mistral, Llama2, Gemma, and Llama families have higher performance on most of the relations than Pythia, Bloom, and OPT, indicating the latter's lower factual knowledge.



\begin{figure}[!t]
    \centering
    \includegraphics[width=0.33\textwidth]{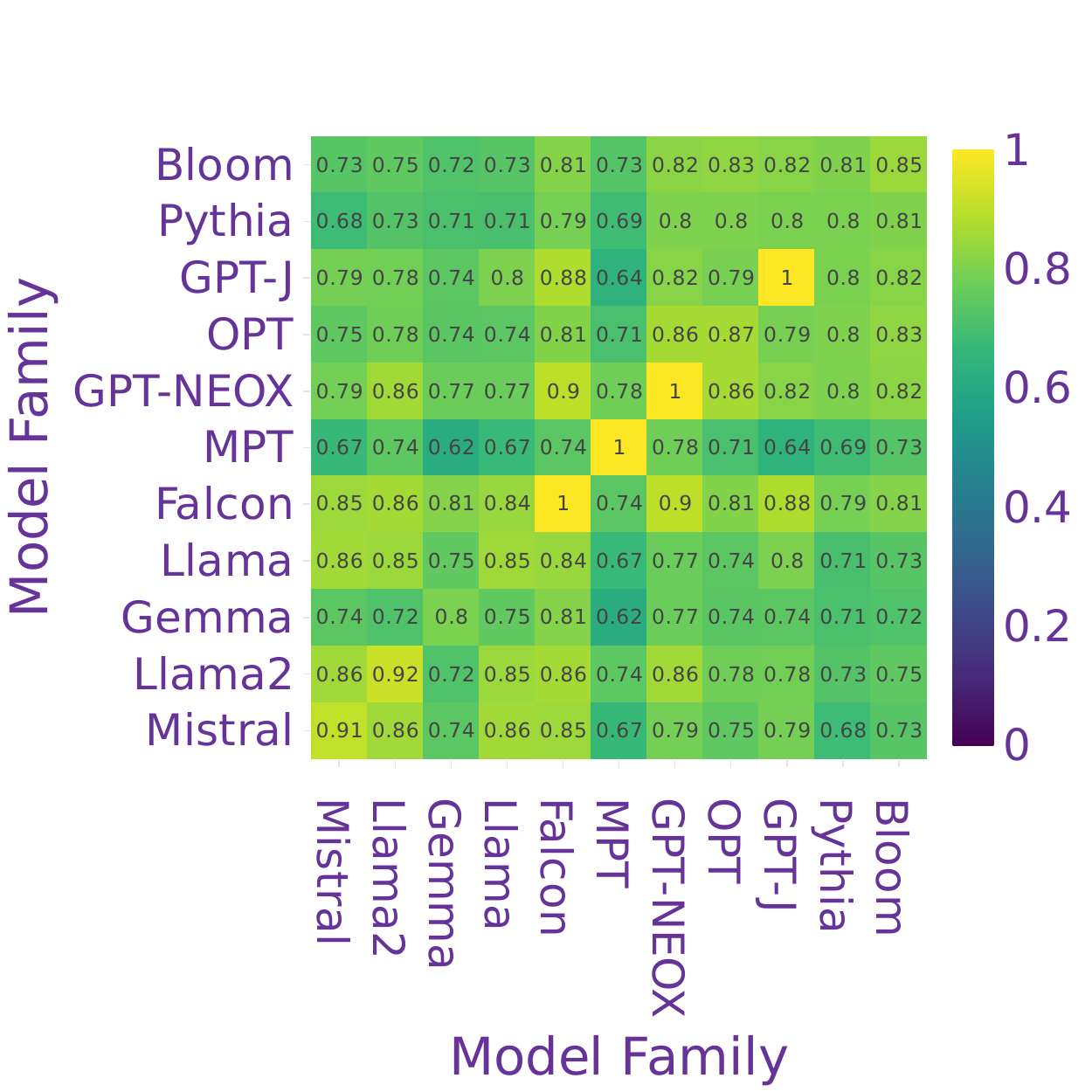}
    \caption{\change{Pearson correlation coefficients between model families. We compute pairwise Pearson correlations between models and calculate the average score within each family.}
    } 
    \label{fig:family_correlation}
\end{figure}

\textbf{Different model families align in their relative factual knowledge.}
Although different model families have different knowledge levels, they have similar knowledge structures. We investigate the correlations between each model pair's performance over 50 relations to assess the agreement in their knowledge levels. We compute the average correlations within each model family (e.g. Llama2 7B, 13B, 70B) in Figure \ref{fig:family_correlation}. Despite differences in architecture and training datasets among model families, there is a significant consensus (correlation > 0.6) regarding the hierarchy of knowledge across various relations. 
We also compile the three best and worst-performing relations for each model in Table~\ref{table:top3_relation_order}, illustrating the consensus among all models. The consistent underperformance across specific relations also suggests that \textit{certain types of knowledge are universally less well-represented across different models}, regardless of their architecture or size. This consistency in less-known knowledge across models highlights a potential vulnerability that could be exploited if these weaknesses are not addressed. Figure~\ref{fig:all_models_correlation} shows the correlations between all the models within each family.



\subsubsection{Comparing within the same LLM family}



\textbf{Larger models embed more knowledge with certain exceptions.}
 Figure~\ref{fig:accuracy_across_all} shows that within each model family, larger models (e.g., Llama-65B) generally outperform smaller ones (e.g., Llama-13B). Models within the same family are typically pre-trained on the same datasets~\citep{biderman2023pythia, zhang2022opt, touvron2023llama2}. The results suggest that, when trained on identical datasets, larger models in general capture a broader set of facts. An exception lies in the OPT group of models \citep{zhang2022opt} that have also been trained on the Pile dataset \citep{gao2020pile} like some of the other models \citep{biderman2023pythia}. This may call for investigating the ways of knowledge injection and validating if that can be attributed to such a performance deviation.

\textbf{Despite being trained on the same data, models might remember different facts.}
From the above results, it is not clear if the larger models are subsuming smaller models in their factual knowledge, i.e., do the larger models correctly identify the facts that the smaller models are correct on?
To assess this, we compute the \emph{subsumption rate} $\eta$:

\small
\[
    \eta(\llm_1 | \llm_2, \setOfFacts) = \frac{|\lke(\llm_1, \setOfFacts) \cap \lke(\llm_2, \setOfFacts)|}{|\lke(\llm_1, \setOfFacts)|}
\]
\normalsize
that measures how much of the fraction of facts from $\setOfFacts$ known by smaller model $\llm_1$ are also recognised by the larger model $\llm_2$.
A subsumption rate of $\sim$ 1 indicates that all of the 
smaller model's knowledge is also contained in the larger model.

\begin{table}[t]
     \caption{
    Average subsumption rate ($\asubr$) for different model families over the relations in \trexmc. Accuracy corresponds to the multiple-choice accuracy.
    }
    \label{tab:avg_subsumption}
    \centering
     \scalebox{0.9}{
    \begin{tabular}{ccccc|c}
  
         & \multicolumn{2}{c}{Smallest Model} & \multicolumn{2}{c}{Largest Model} \\
         \cmidrule{2-5}
        Family  & \#Parameters & Accuracy & \#Parameters & Accuracy  & $\asubr$ \\
        \midrule
        Llama & 7B & 0.699 & 65B & 0.836 & 0.769 \\
        Llama-2 & 7B & 0.741 & 70B & 0.846 & 0.801 \\
        Gemma & 2B & 0.666 & 7B & 0.750 & 0.710 \\
        OPT & 125m & 0.430 & 30B & 0.588 & 0.481 \\
        Pythia & 70m & 0.334 & 12B & 0.648 & 0.403 \\
        Bloom & 560m & 0.410 & 7.1B & 0.548 & 0.498 \\

        \bottomrule
    \end{tabular}    }
\end{table}

Table~\ref{tab:avg_subsumption} shows the average subsumption rate ($\asubr$) between the largest and smallest models in a family, as well as the average accuracy, over all relations for different model families.
Interestingly, $\asubr$ is relatively low (< 0.5) for OPT, Pythia and Bloom (i.e., the larger models know less than 50\% of what the smaller models know) and only reaching up to 0.8 for Gemma, Llama and Llama-2. 
Therefore, \textit{even though models within each family are trained on the same datasets and generally agree on the relative knowledge of different relations (Figure \ref{fig:family_correlation}), there are differences in the knowledge of specific facts they retain from their training data}. These discrepancies suggest that simply increasing the model size may not be sufficient to enhance factual knowledge, thus requiring the need for proper factual knowledge injection into the models. 

\begin{figure}[t]
    \centering
    \includegraphics[width=0.4\textwidth, trim=0 0 0 2cm]{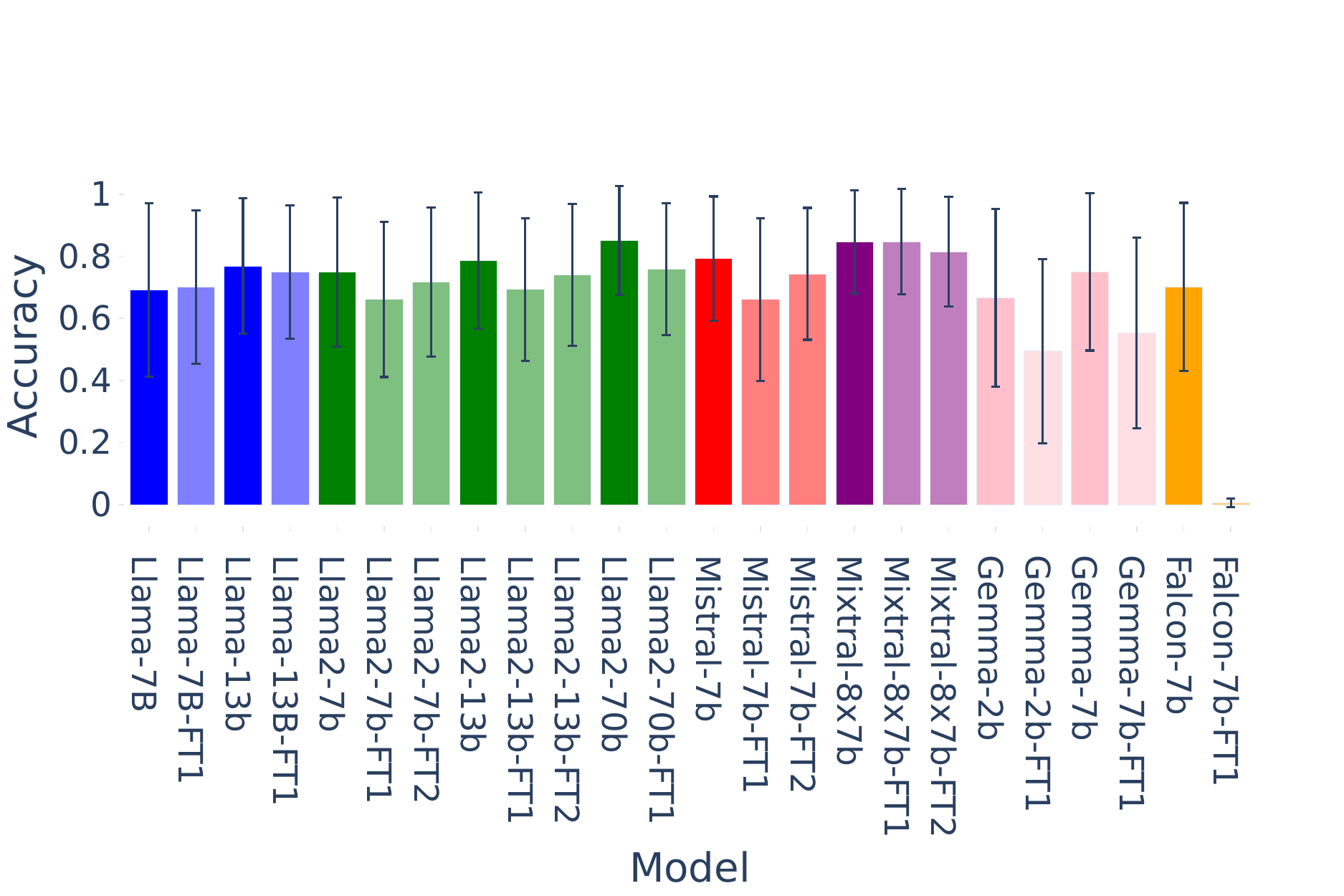}
    \caption{
    Multiple-choice accuracy of base vs. chat-finetuned models. Finetuned models (lighter shades) show lower accuracy across \trexmc~relations compared to pre-trained models (darker shades).
    }
    \label{fig:ft-bar}
\end{figure}

\textbf{Instruction fine-tuning reduces latent knowledge.}
\label{subsec:results_finetuning}
Finally, we investigate the effects of chat-based instruction fine-tuning on the factual knowledge of models. Base language models are often fine-tuned (using a mix of supervised and reinforcement learning~\cite{ouyang2022training}) to improve their ability to follow instructions. While previous studies have shown that fine-tuning enhances performance on various benchmarks, its impact on latent knowledge is unclear.
Figure~\ref{fig:ft-bar} illustrates the comparative accuracy of pre-trained models and their fine-tuned counterparts.
In almost all cases, \textit{the fine-tuned models obtain lower accuracy than their base versions, suggesting that fine-tuning reduces the amount of latent knowledge estimation.} A similar observation was made by~\cite{yu2024kola}.
 To further assess if fine-tuned models acquire new knowledge, we compute the subsumption rate between pre-trained and fine-tuned versions (Table~\ref{tab:avg_subsumption_ft}). 
 We find that most latent knowledge in fine-tuned models is already present in base models (high $\asubr$). This outcome highlights the need for caution when fine-tuning models, as these adjustments might inadvertently compromise with the existing internal knowledge.
\vspace{-1mm}
\section{Concluding Discussion}
In this work, we investigate a new way to estimate latent factual knowledge from an LLM.
Unlike prior approaches, our method does not engineer prompts (\textbf{zero-prompting}). Rather it relies on LLMs' in-context learning ability to infer the factual knowledge question and the expected answer format. Our method not only addresses many reliability concerns with prompting, but it also recollects significantly more factual knowledge than prompting.
In contrast to prompting, which requires relationship-specific and LLM-specific prompt engineering, 
Our method can be applied with minimal effort to test factual knowledge of relations across a variety of structured knowledge bases and LLMs.
This ability enables us to compare the latent knowledge captured by many different families of open-source LLMs; we expect our results to be of interest to the designers of these LLMs.
Finally, to design our zero-prompt many-shot LKE, we explore the impact of the number and order of correct, incorrect, and unknown examples used as inputs; our findings may be of independent interest to developing a better understanding of different learning modes of in-context learning. 

A fundamental question posed by our and prior work on estimating latent knowledge in LLMs: {\it What does it mean for an LLM to know a fact?} Suppose we tried to infer if an LLM knows the capital of Germany using the input "France Paris; Spain Madrid; Germany " and suppose the answer was \textit {Berlin}. What we have learned is that the LLM knows that the relationship $\relation$ between Germany and Berlin is similar to that between France and Paris or Spain and Madrid. What we have not learned is whether the LLM knows that the relation $\relation$ is called "capital" in English or "hauptstadt" in German. The latter is revealed by prompts such as "The capital of Germany is ".
But, such prompts don't reveal whether the LLM knows that what Berlin means to Germany is similar to what Paris means to France.  

\textit {Is one type of knowing facts better than another?} It is difficult to answer in general. Neither type of knowing guarantees that the knowledge can be put to use in different contexts and tasks, such as when we ask the LLM where the parliament of Germany is located. 
However, they lead to different strategies for getting LLMs to generate correct outputs. With the first type of knowing, we can use a list of facts as input such as "The parliament of France is in Paris; The parliament of Spain is in Madrid; The parliament of Germany is in ". With the second type of knowing, we can hope to use a chain of thought prompts such as "The parliament of a country is in its capital. The parliament of Germany is in ".
Nevertheless, one clear takeaway from our study is related to {\it how factual knowledge is latently embedded in an LLM}.
We show that more factual knowledge can be recollected using in-context learning, i.e., the representations of subjects and objects that share the same relationship, than by prompting with the name of their relationship.
\newpage

\section{Ethical Considerations}
Our research utilizes public datasets and open-source LLMs, which mitigates immediate privacy concerns. However, our findings on the factual knowledge capabilities of various LLMs could influence their deployment in real-world applications, potentially leading to over-reliance on models for tasks requiring factual accuracy. We encourage users of our methodology to consider these implications and to use the knowledge estimation techniques responsibly, with appropriate safeguards against potential misuse. Furthermore, as our work may reveal biases or gaps in the factual knowledge of LLMs, we urge developers to address these issues to ensure fair and equitable AI systems.

\bibliographystyle{ACM-Reference-Format}
\bibliography{custom}
\balance

\appendix
\onecolumn
\section{Dataset}
\label{appendix:dataset_construct}

\subsection{Creation of Nobel laureates dataset from Wikidata}
\label{appendix:nobel_dataset}

The Nobel Dataset is a collection of biographical information about all Nobel laureates up until the year 2022, totaling 954 individuals. This dataset was curated using data obtained from Wikidata's querying service\footnote{https://query.wikidata.org/}. The following attributes are included for each laureate:
\begin{itemize}
    \item \textbf{Name:} The full name of the Nobel laureate.
    \item \textbf{Birth Year:} The year in which the laureate was born.
    \item \textbf{Award Year:} The year(s) in which the laureate was awarded the Nobel Prize.
    \item \textbf{Nature of Award:} A brief description of the reason for the award, including the field of the Nobel Prize (e.g., Physics, Peace).
    \item \textbf{Gender:} The gender of the laureate.
\end{itemize}

Here are some examples from the Nobel Dataset:

\begin{table}[h!]
\centering
\caption{Excerpt from the Nobel Dataset}
\begin{tabular}{|l|c|c|l|c|}
\hline
\textbf{Name} & \textbf{Birth Year} & \textbf{Award Year} & \textbf{Nature of Award} & \textbf{Gender} \\
\hline
Albert Einstein & 1879 & 1921 & Physics & male \\
Louis de Broglie & 1892 & 1929 & Physics & male \\
Carl D. Anderson & 1905 & 1936 & Physics & male \\
Polykarp Kusch & 1911 & 1955 & Physics & male \\
Melvin Schwartz & 1932 & 1988 & Physics & male \\
Jerome I. Friedman & 1930 & 1990 & Physics & male \\
\hline
\end{tabular}
\label{table:nobel}
\end{table}

\subsection{Creation of multiple choices from T-REx: TREx-MC}
\label{appendix:trex_mc}

T-REx~\cite{elsahar2018t} is a large-scale alignment dataset that aligns between Wikipedia abstracts and Wikipedia triples. We have utilized the processed version of T-REx available on HuggingFace~\footnote{\url{https://huggingface.co/datasets/relbert/t_rex}} for our experiments. We filtered out the relations that have more than 500 facts and 100 unique object entities. The unique objects ensure having 100 feasible multiple choices for each fact in each relation. 
We also manually filtered out relations with multiple correct objects (e.g. ``America", ``USA", ``American") to avoid ambiguity. Additionally for relations that have objects in the form of partial matches (e.g. ``French", ``French language"), the respective objects have been standardized to uniform values (e.g. ``French").
We curated 50 relations for our dataset TREx-MC that essentially consists of $<$ \textit{subject, relation, multiple choices} $>$. The multiple choices comprise the correct answer along with 99 other potential choices. We list the 50 relations in  Table~\ref{tab:trex-mc} below.

The following attributes are included in TREx-MC dataset for each relation:

\begin{itemize}
    \item \textbf{Subject} : The subject entity for each fact.
    \item \textbf{Object}: The object entity or the correct answer for each fact.
    \item \textbf{Multiple choices}: The list of other potential choices for each fact.
    \item \textbf{Title} : The Wikipedia title for each fact.
    \item \textbf{Text}: The Wikipedia abstract corresponding to each fact.
\end{itemize}

\begin{table}[]
\centering
\caption{List of 50 relations from T-REx-MC}
\label{tab:trex-mc}
\scriptsize
\begin{tabular}{|l|l|l|l|l|l|l|l|l|l|}
\hline
\begin{tabular}[c]{@{}l@{}}date of \\ birth\end{tabular}   & \begin{tabular}[c]{@{}l@{}}date of \\ death\end{tabular}                              & director                                                       & father                                                                                         & spouse                                                   & child                                                            & sibling                                                     & composer                                                                           & \begin{tabular}[c]{@{}l@{}}is a \\ tributary of\end{tabular}                         & student of                                                    \\ \hline
instance of                                                & \begin{tabular}[c]{@{}l@{}}cast \\ member\end{tabular}                                & genre                                                          & \begin{tabular}[c]{@{}l@{}}contains the \\ administrative\\ territorial \\ entity\end{tabular} & \begin{tabular}[c]{@{}l@{}}educated \\ at\end{tabular}   & \begin{tabular}[c]{@{}l@{}}parent \\ taxon\end{tabular}          & \begin{tabular}[c]{@{}l@{}}screen\\ writer\end{tabular}     & performer                                                                          & capital                                                                              & producer                                                      \\ \hline
is made by                                                 & \begin{tabular}[c]{@{}l@{}}named \\ after\end{tabular}                                & developer                                                      & publisher                                                                                      & \begin{tabular}[c]{@{}l@{}}founded \\ by\end{tabular}    & \begin{tabular}[c]{@{}l@{}}drafted \\ by\end{tabular}            & \begin{tabular}[c]{@{}l@{}}has \\ played \\ at\end{tabular} & \begin{tabular}[c]{@{}l@{}}part of \\ the series\end{tabular}                      & manufacturer                                                                         & \begin{tabular}[c]{@{}l@{}}production \\ company\end{tabular} \\ \hline
mother                                                     & \begin{tabular}[c]{@{}l@{}}cause of \\ death\end{tabular}                             & \begin{tabular}[c]{@{}l@{}}has\\ subsidiary\end{tabular}       & creates                                                                                        & \begin{tabular}[c]{@{}l@{}}point in \\ time\end{tabular} & inception                                                        & \begin{tabular}[c]{@{}l@{}}publication\\ date\end{tabular}  & \begin{tabular}[c]{@{}l@{}}languages\\ spoken, \\ written\\ or signed\end{tabular} & \begin{tabular}[c]{@{}l@{}}original \\ language\\ of film or \\ TV show\end{tabular} & \begin{tabular}[c]{@{}l@{}}official \\ language\end{tabular}  \\ \hline
\begin{tabular}[c]{@{}l@{}}native \\ language\end{tabular} & \begin{tabular}[c]{@{}l@{}}position \\ played \\ on team / \\ speciality\end{tabular} & \begin{tabular}[c]{@{}l@{}}original\\ broadcaster\end{tabular} & \begin{tabular}[c]{@{}l@{}}record\\ label\end{tabular}                                         & author                                                   & \begin{tabular}[c]{@{}l@{}}discoverer\\ or inventor\end{tabular} & characters                                                  & lyrics by                                                                          & distributed by                                                                       & home venue                                                    \\ \hline
\end{tabular}
\end{table}

Some examples from the T-REx-MC dataset for 2 relations are listed in Table \ref{tab:excerpt-trexmc}

\begin{table}[]
\centering
\caption{Excerpts from T-REx-MC Dataset}
\label{tab:excerpt-trexmc}
\scriptsize
\begin{tabular}{lllll}
\hline
\multicolumn{1}{c}{\textbf{Subject}} & \multicolumn{1}{c}{\textbf{Object}} & \multicolumn{1}{c}{\textbf{Multiple choices}}                                                                       & \multicolumn{1}{c}{\textbf{Title}} & \multicolumn{1}{c}{\textbf{Text}}                                                                                                                                  \\ \hline
\multicolumn{5}{c}{\textbf{Date of birth}}                                                                                                                                                                                                                                                                                                                                                                 \\ \hline
Giovanni Bia                         & 24 October 1968                     & \begin{tabular}[c]{@{}l@{}}{[}'26 September 1981', \\ '20 February 1981', \\ ..,'20 September 1960'{]}\end{tabular} & Giovanni Bia                       & \begin{tabular}[c]{@{}l@{}}Giovanni Bia \\ (born 24 October 1968) \\ is a former \\ Italian footballer...\end{tabular}                                             \\ \hline
Brian May                            & 19 July 1947                        & \begin{tabular}[c]{@{}l@{}}{[}'24 December 1931', \\ '1 December 1976',\\ ... '23 August 1964{]}\end{tabular}       & Brian May                          & \begin{tabular}[c]{@{}l@{}}Brian Harold May, CBE \\ (born 19 July 1947) \\ is an English musician...\end{tabular}                                                  \\ \hline
\multicolumn{5}{c}{\textbf{Composer}}                                                                                                                                                                                                                                                                                                                                                                      \\ \hline
Mexico Trilogy                       & Robert Rodriguez                    & \begin{tabular}[c]{@{}l@{}}{[}'Fred Schneider', 'Brandy',\\ .., 'Tommaso Traetta'{]}\end{tabular}                   & Mexico Trilogy                     & \begin{tabular}[c]{@{}l@{}}The Mexico Trilogy or \\ Mariachi Trilogy \\ (also Desperado Trilogy \\ on some DVD releases)\\  is a series of American..\end{tabular} \\ \hline
Chelsea Walls                        & Jeff Tweedy                         & \begin{tabular}[c]{@{}l@{}}{[}'Carmine Coppola', \\ 'Jimmy Chi', \\ ...'Maurice Ravel'{]}\end{tabular}              & Chelsea Walls                      & \begin{tabular}[c]{@{}l@{}}Chelsea Walls is a 2001\\ independent  film\\ directed by Ethan Hawke \\ and released by Lions Gate\\ Entertainment.\end{tabular}       \\ \hline
\end{tabular}
\end{table}


\section{Inference Setup}
\label{appendix: inf_setup}

We experiment with and use three different inference setups:

\begin{enumerate}
    \item Transformers Based Setup: This setup utilizes the utilities present in the transformers library \cite{wolf-etal-2020-transformers} to obtain the log probabilities for generating the different options.
    
    \item vLLM Based Setup: vLLM (\citep{vLLM}) is a fast inference library for large language models (LLMs). It efficiently manages attention key and value memory using PagedAttention. We observed considerable speed boosts for all 3 LKEs compared to the standard Transformers API.
    
    \item SGLang Based Setup: SGLang \cite{SGLang} is a structured generation language designed for large language models (LLMs). It speeds up LLM interactions and provides enhanced control through tight integration of its frontend language and backend runtime system. SGLang also leverages Radix Attention to cache common components across queries in the KV cache, enabling substantial speedups. We observed sizable speed boosts for \iclke~ over vLLM. However, we are constrained by SGLang's limited model family support at the moment, and only utilize it for the Llama, Mistral, and Mixtral families.
\end{enumerate}

\clearpage
\section{Implementation Details}

\iclke~ leverages 50 randomly chosen samples from the training data as in-context examples but does not use the relation name. The base prompt is now composed of 50 different examples followed by the name of the entity being tested. A sample would be ``Albert Einstein 14 March 1879 Ernest Rutherford 30 August 1871 ... J.J. Thomson 18 December 1856 Max Planck."

A single forward pass is conducted for each sequence, generating log probabilities for the entire sequence. The common part, represented by the tokens for the base prompt is then removed from the tokens of the concatenated base prompt and option resulting in the log probabilities for the option. If the option is tokenized into multiple tokens, a single probability value is obtained by multiplying the individual token probabilities. The resulting values are normalized across multiple choices, and the option with the highest probability is selected as the correct answer. We use the vLLM Based \& SGLang Based Setup for this LKE.

\section{Different Metrics}
\label{appendix:diff_metrics}
\begin{figure}[h]
    \centering
    \includegraphics[width=0.7\textwidth]{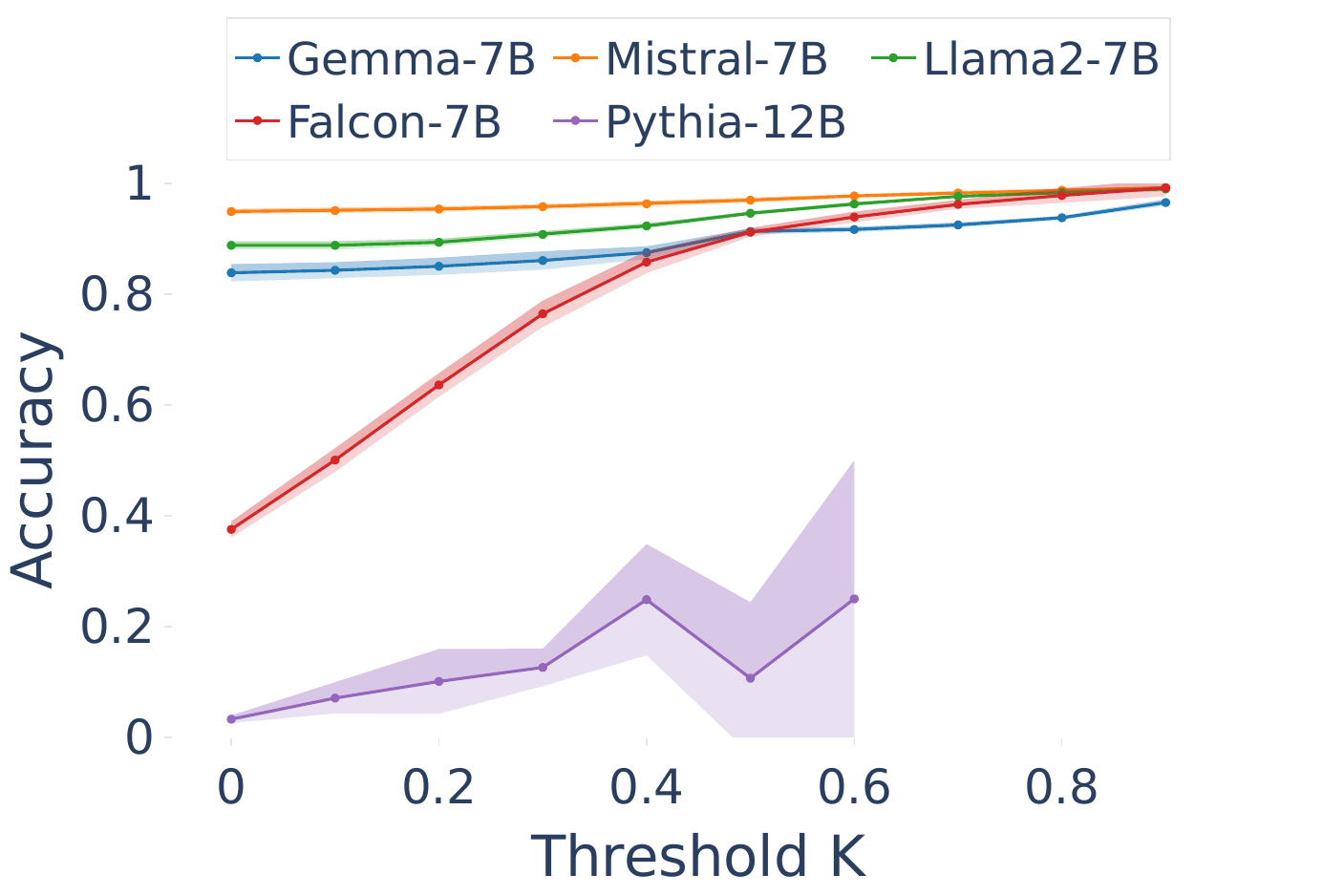}
    \caption{\textbf{Multiple-choice Accuracy@K for different models}. We evaluated five models on the Nobel dataset, which consists of 50 examples. Each model's performance was measured using the Accuracy@K metric at various thresholds.}
    \label{fig:accuracy_k}
\end{figure}

The evaluation metric can readily be adapted to existing classification metrics. 
For example, we introduced the metric Accuracy@K, a calibrated measure that assesses a model's confidence in its predictions. This metric quantifies how accurately the model identifies knowledge at specified confidence levels for a given relation.
We filter the instances that have their confidence levels > threshold $K$ and form the set
$\dataset_K=\{c_i|\pred_{\llm}(c_i) \geq K\ \forall c\in \dataset\}$ . Following this, we use our accuracy measure to compute Accuracy@K for varying values of $K$, the results of which are shown in Figure \ref{fig:accuracy_k}.

\begin{equation}\label{eq:accuracy}
\begin{split}
    \acc_K(\llm, \dataset_K) \triangleq
    \frac{\underset{\langle\subject, \relation, \object^*, \setOfObjects\rangle \in \dataset_K}{\sum} \delta \left(\object^* = \pred_{\llm}(\subject, \relation, \object^*, \setOfObjects) \right)}{|\dataset_K|}
\end{split}
\end{equation}.

\clearpage
\section{Probabilities of objects in sequence}
\label{appendix:icl}

\begin{figure}[h!]
\centering
\hfill
\subfloat[Correct Subject-Object Pairs]{%
  \includegraphics[width=0.33\textwidth]{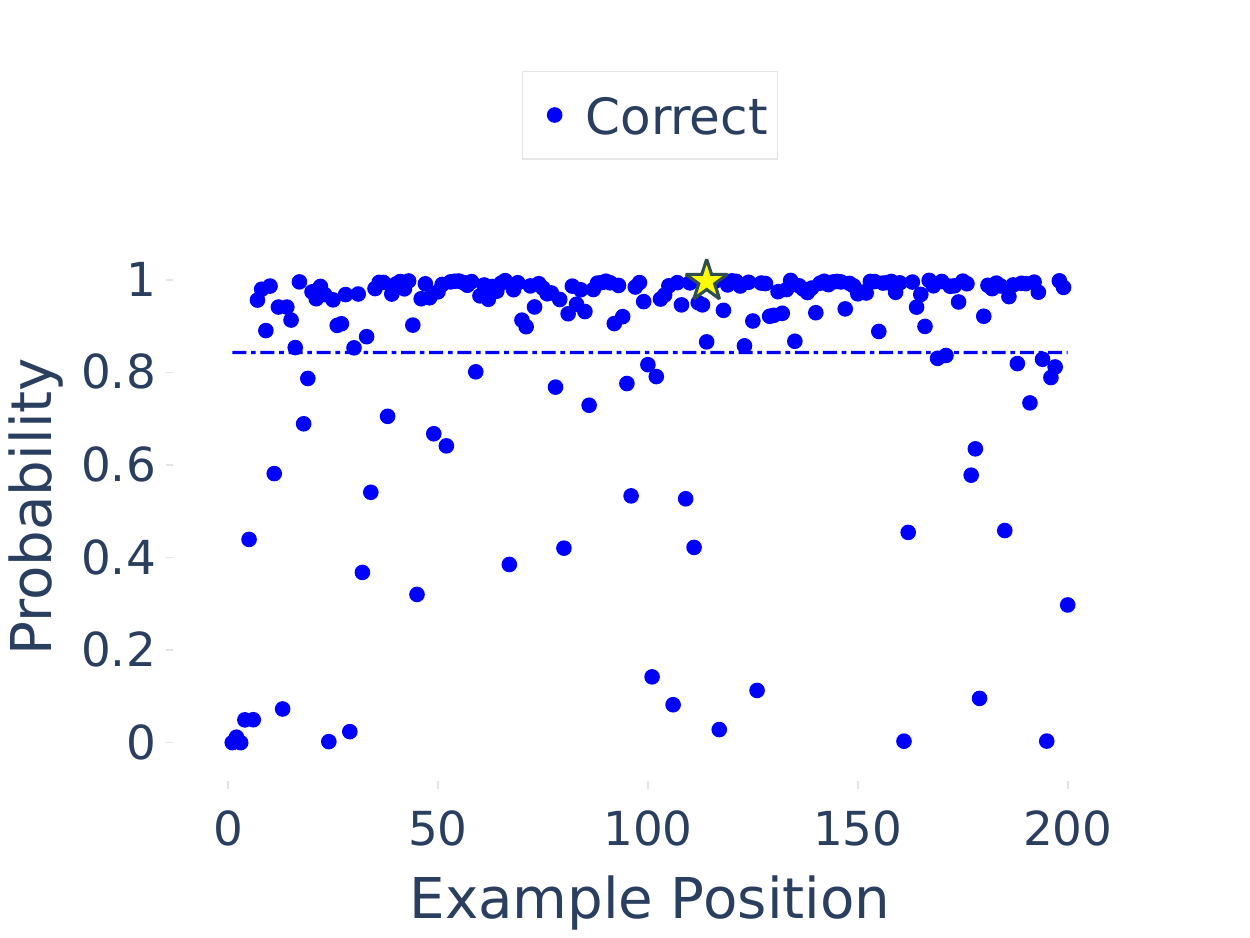}%
  \label{fig:llama_correct_sequence}%
}
\hfill
\subfloat[Distributed incorrect examples]{%
  \includegraphics[width=0.33\textwidth]{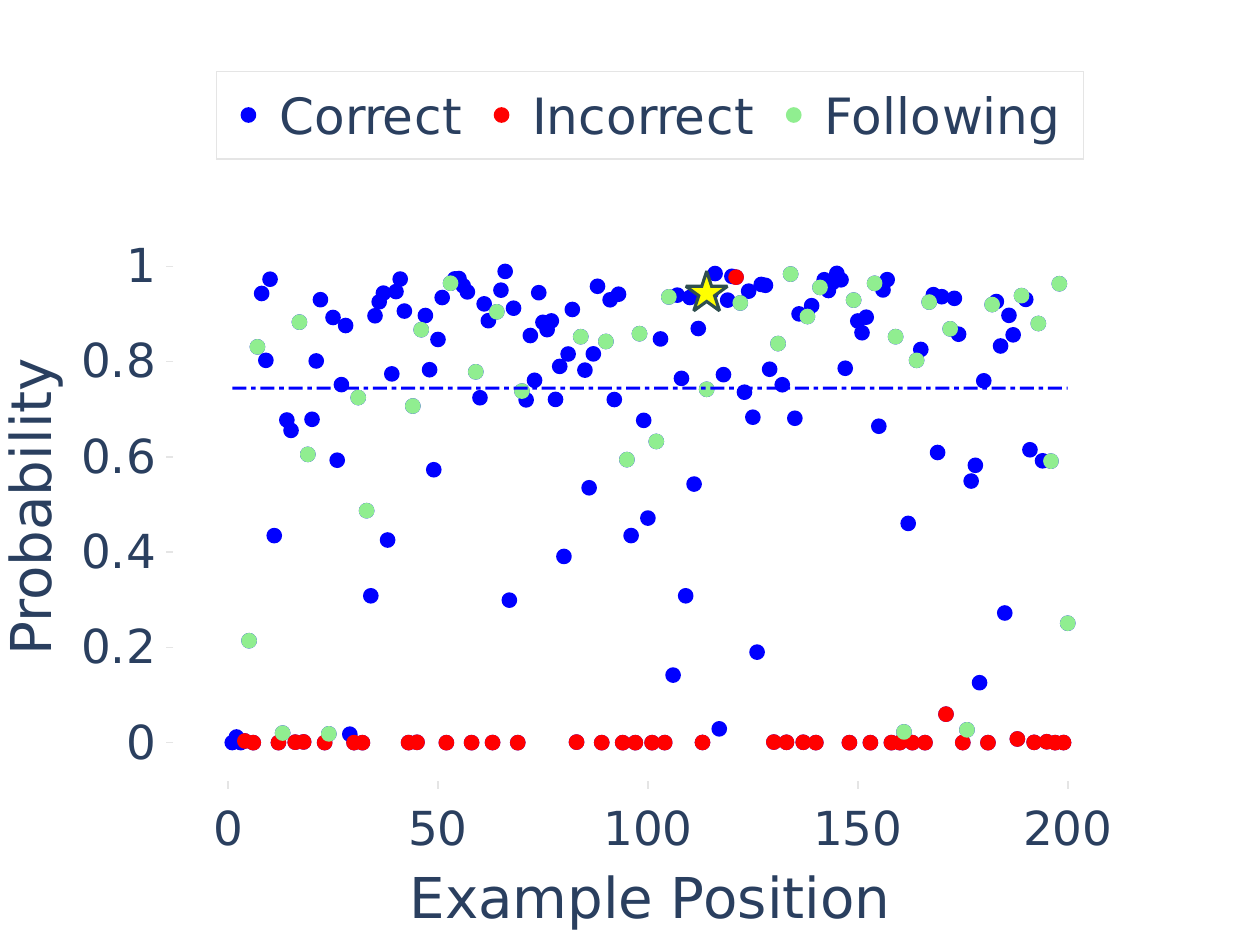}%
  \label{fig:llama_incorrect_distributed}%
}
\subfloat[Simultaneous incorrect examples]{%
  \includegraphics[width=0.33\textwidth]{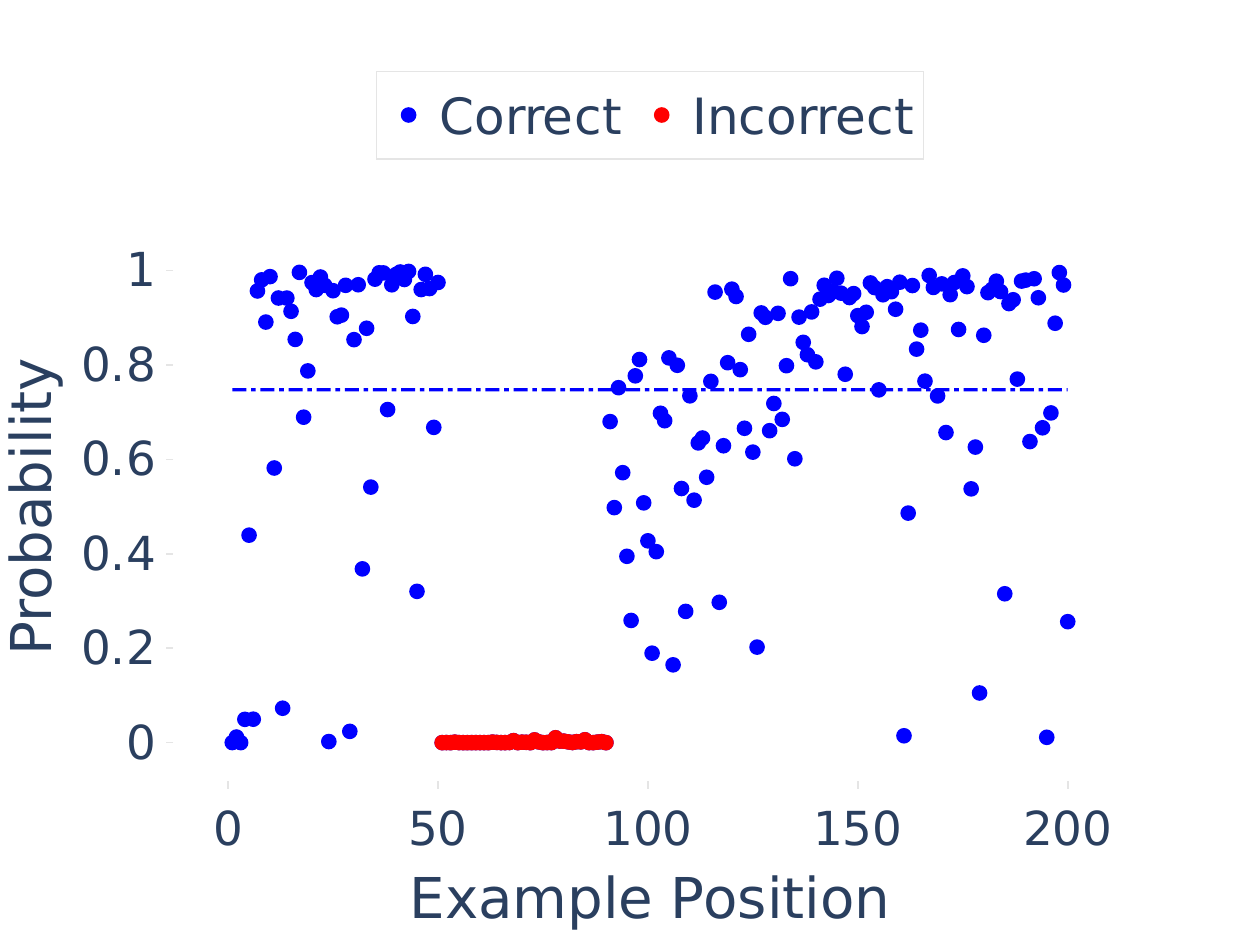}%
  \label{fig:llama_incorrect_simultaneous}%
}
\hfill
\subfloat[Distributed unknown examples]{%
  \includegraphics[width=0.33\textwidth]{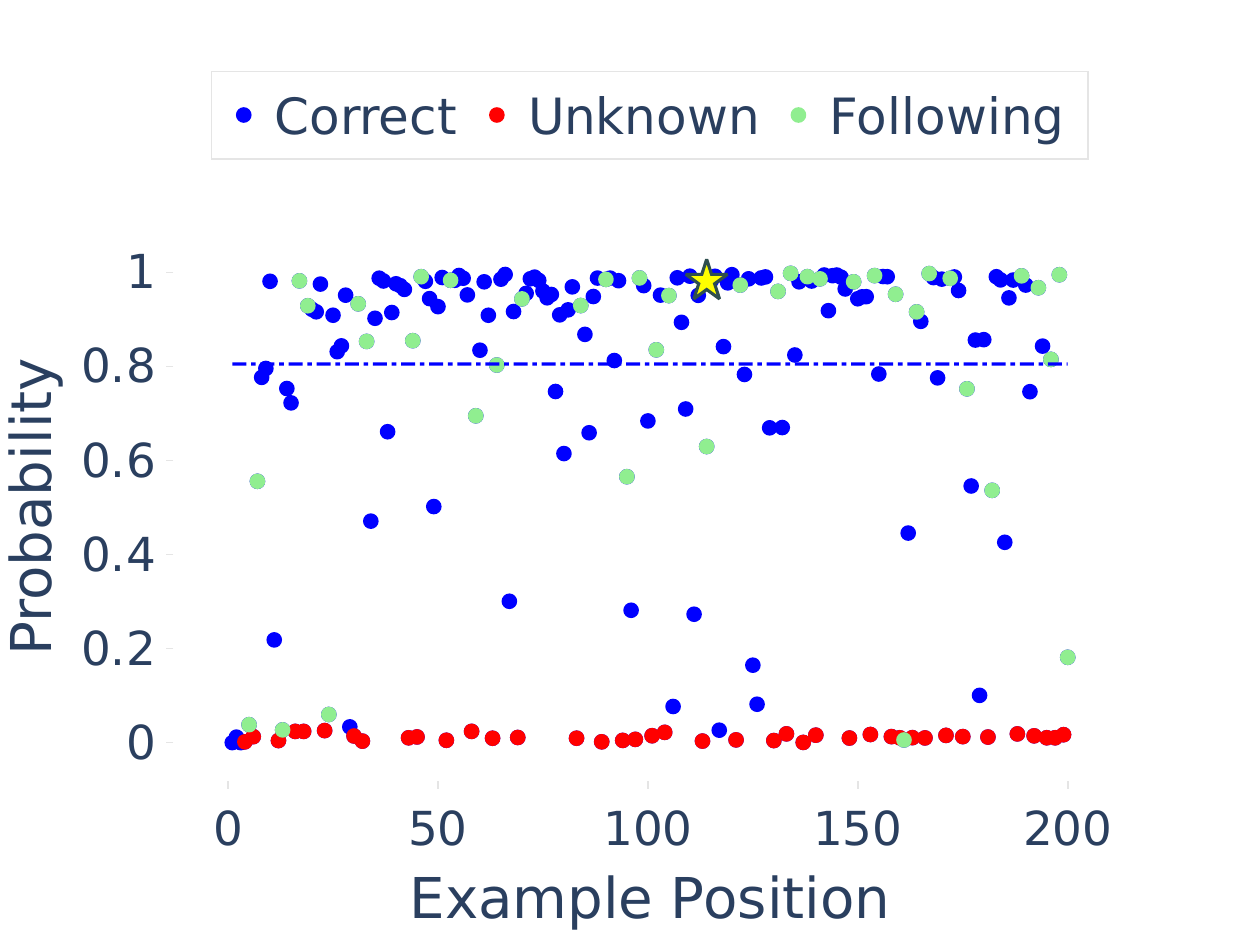}%
  \label{fig:llama_unknown_distributed}%
}
\subfloat[Simultaneous unknown examples]{%
  \includegraphics[width=0.33 \textwidth]{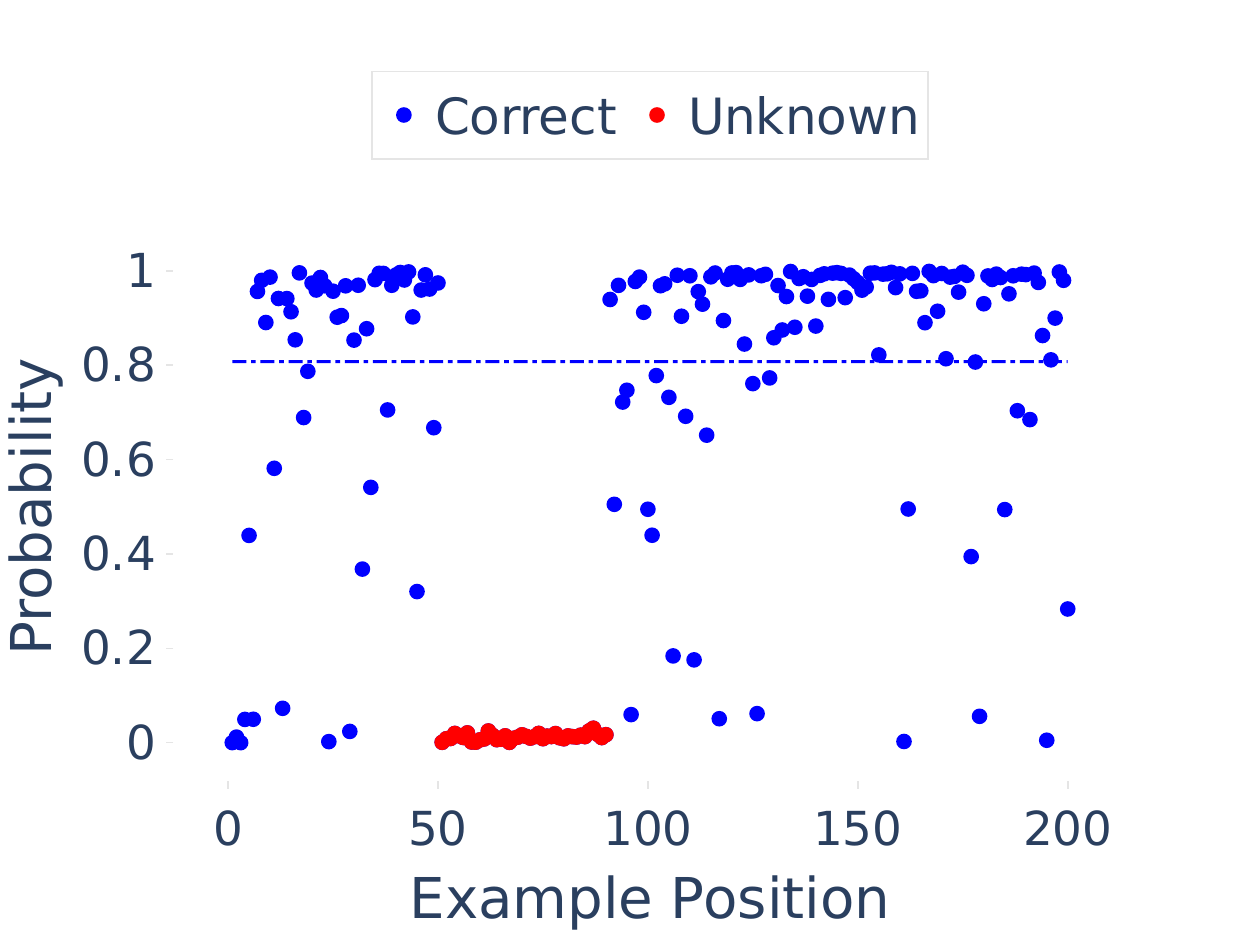}%
  \label{fig:llama_unknown_simultaneous}%
}
\caption{\textbf{Analysis of object probability in one sequence of Nobel laureate data using Llama2-7b} 
}
\label{fig:llama2-7b-probabilities}
\end{figure}

\begin{figure}[h!]
\centering
\hfill
\subfloat[Correct Subject-Object Pairs]{%
  \includegraphics[width=0.33\textwidth]{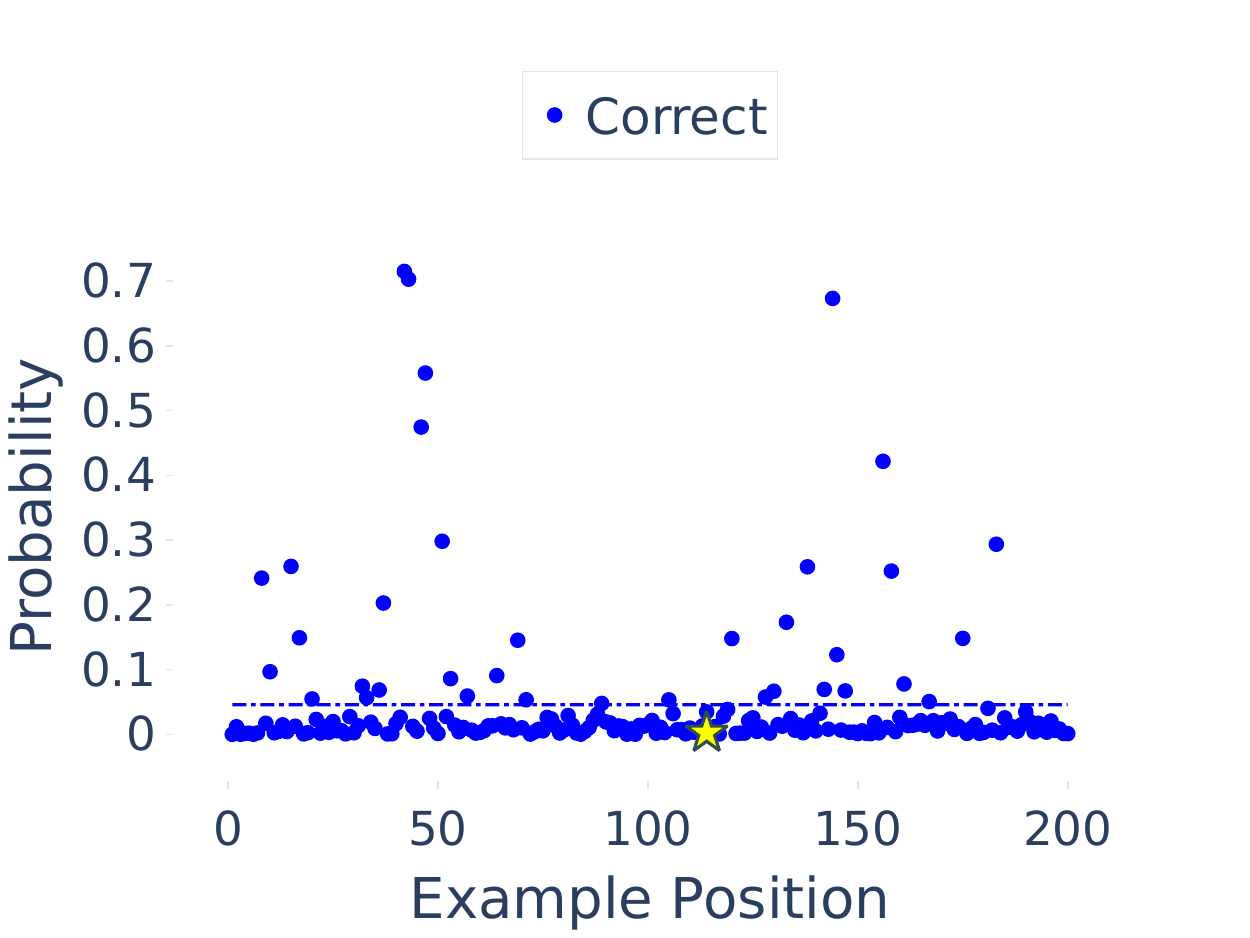}%
  \label{fig:pyt_correct_sequence}%
}
\hfill
\subfloat[Distributed incorrect examples]{%
  \includegraphics[width=0.33\textwidth]{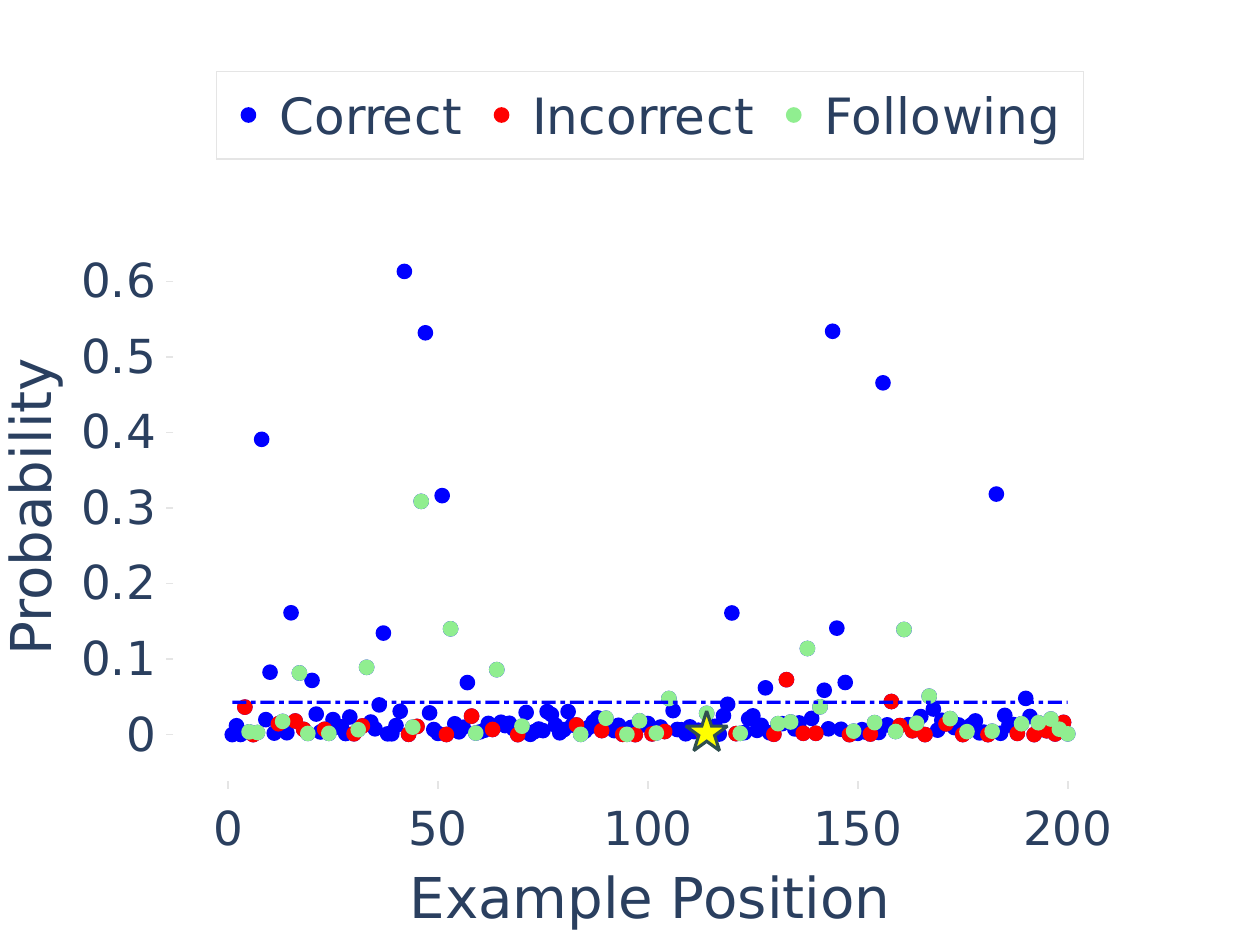}%
  \label{fig:pyt_incorrect_distributed}%
}
\subfloat[Simultaneous incorrect examples]{%
  \includegraphics[width=0.33\textwidth]{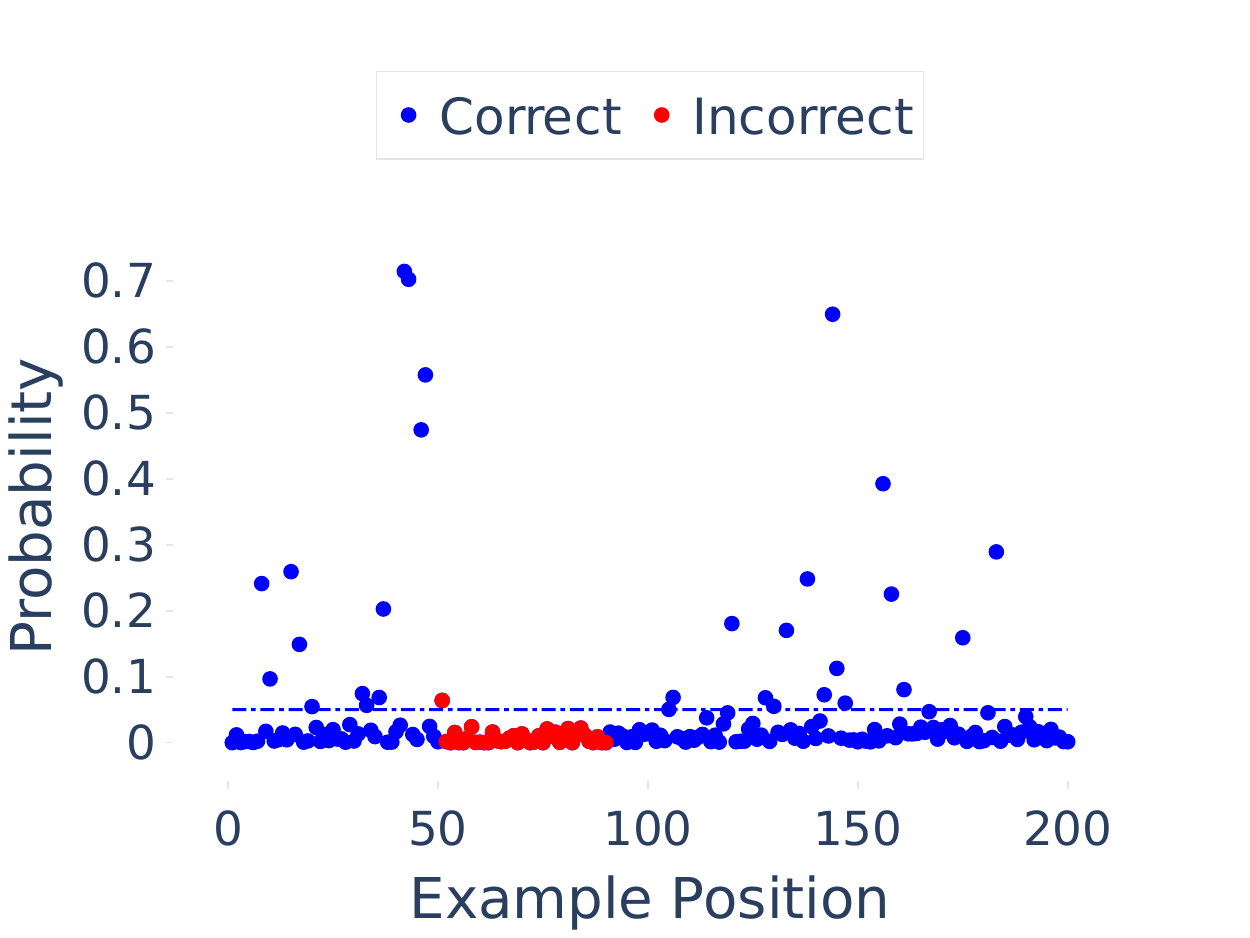}%
  \label{fig:pyt_incorrect_simultaneous}%
}
\hfill
\subfloat[Distributed incorrect examples]{%
  \includegraphics[width=0.33\textwidth]{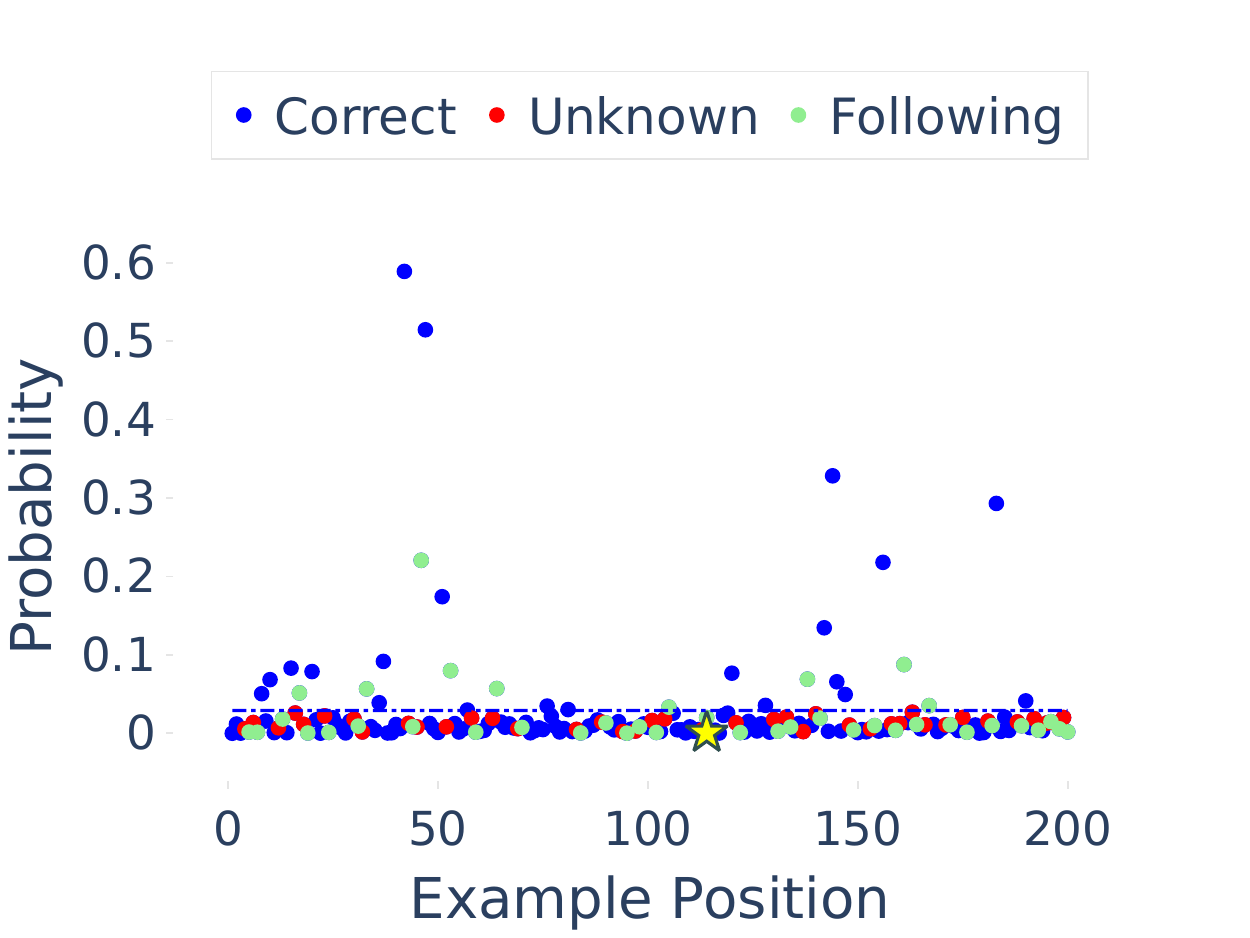}%
  \label{fig:pyt_unknown_distributed}%
}
\subfloat[Simultaneous incorrect examples]{%
  \includegraphics[width=0.33 \textwidth]{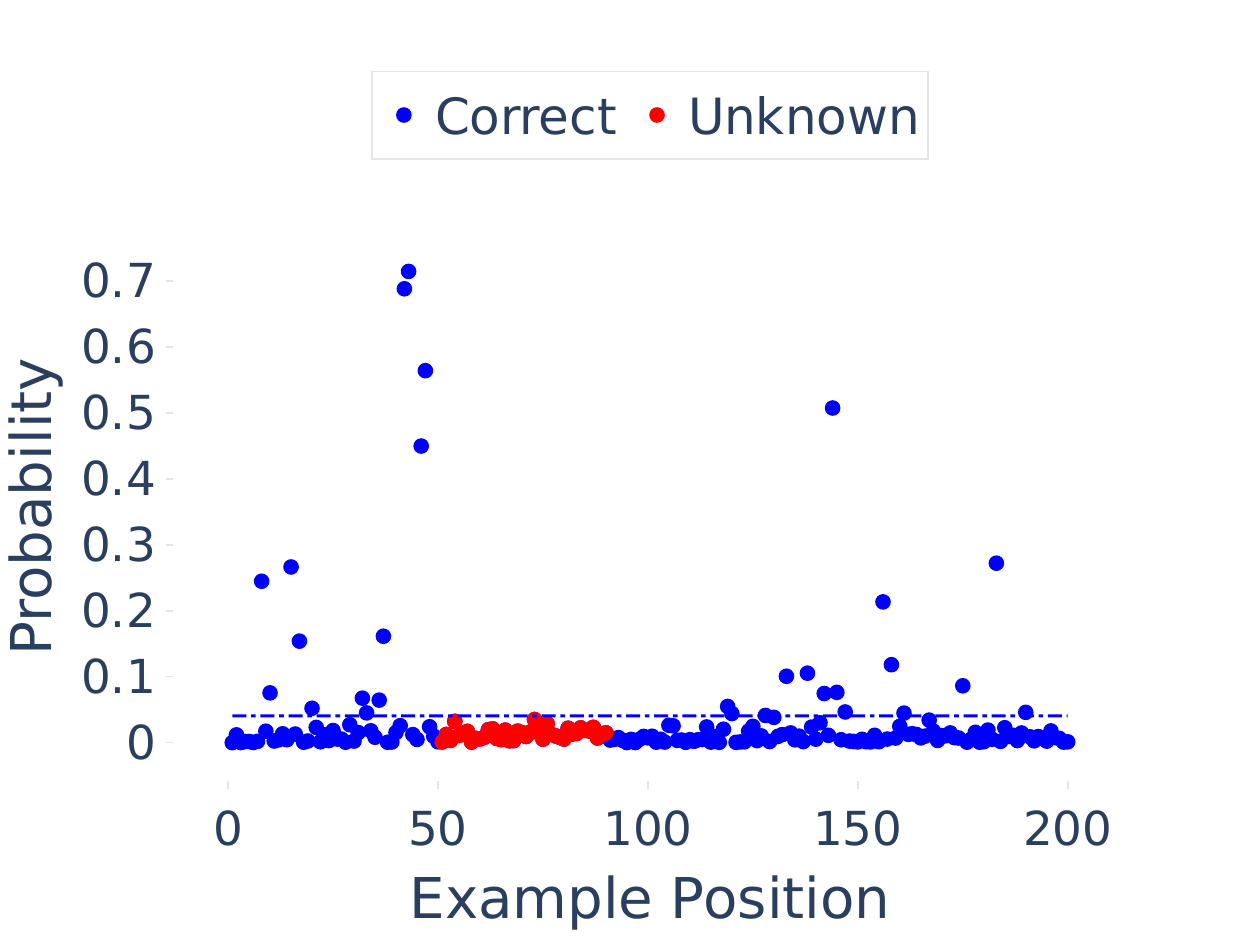}%
  \label{fig:pyt_unknown_simultaneous}%
}
\caption{\textbf{Analysis of object probability in one sequence of Nobel laureate data using Pythia-12B} 
}
\label{fig:pyt-12b-probabilities}
\end{figure}

\begin{figure}[h!]
\centering
\hfill
\subfloat[Correct Subject-Object Pairs]{%
  \includegraphics[width=0.33\textwidth]{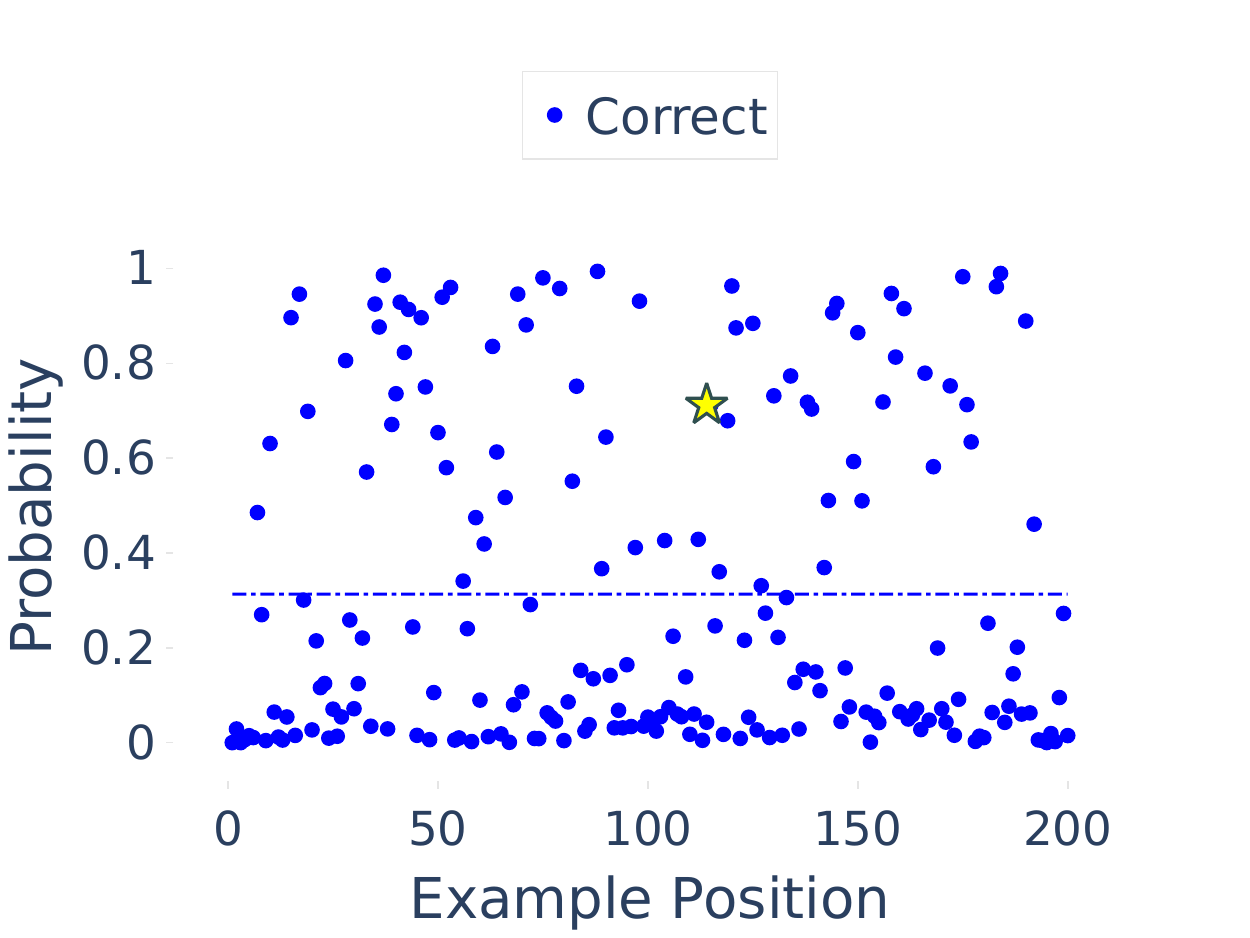}%
  \label{fig:falcon_correct_sequence}%
}
\hfill
\subfloat[Distributed incorrect examples]{%
  \includegraphics[width=0.33\textwidth]{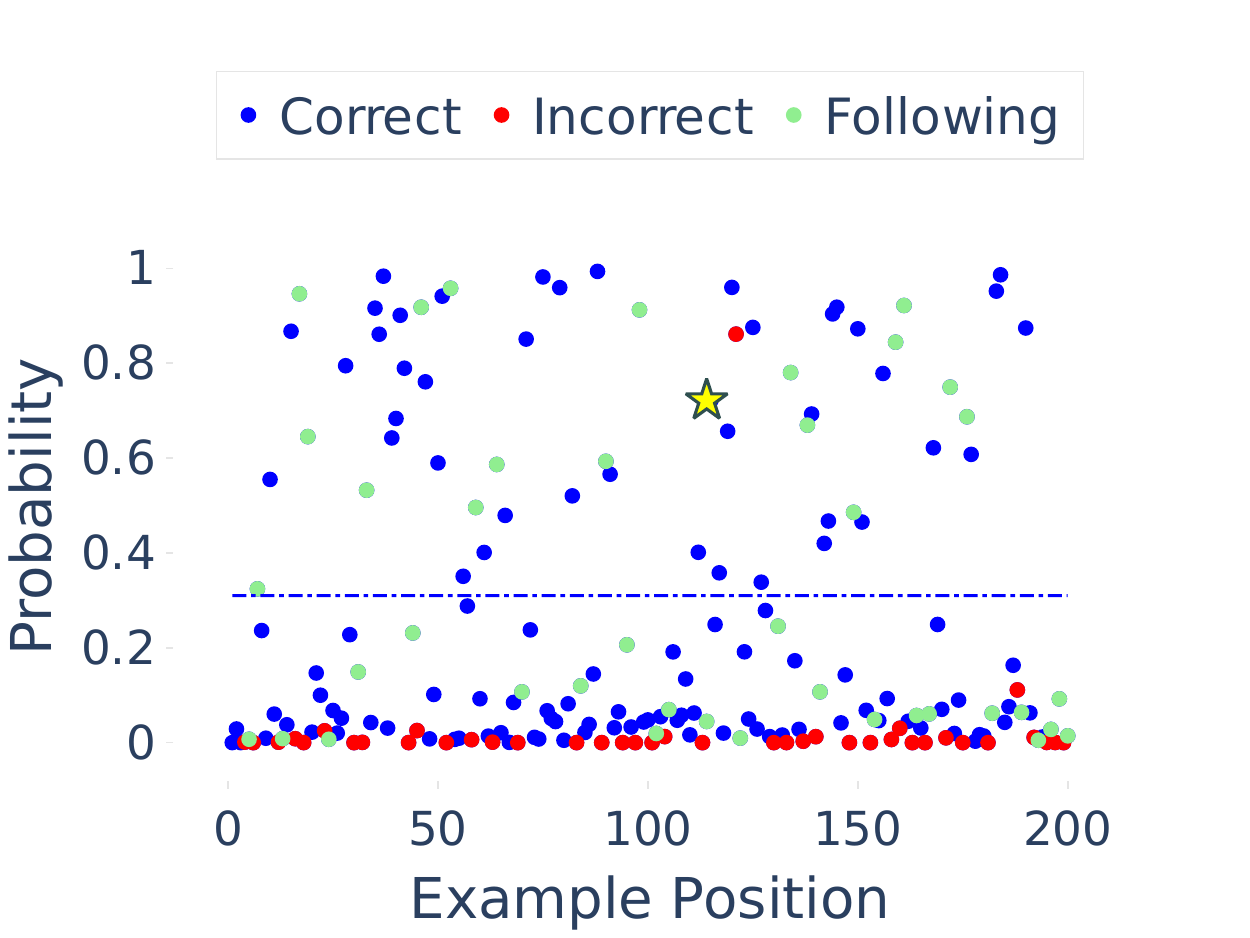}%
  \label{fig:falcon_incorrect_distributed}%
}
\subfloat[Simultaneous incorrect examples]{%
  \includegraphics[width=0.33\textwidth]{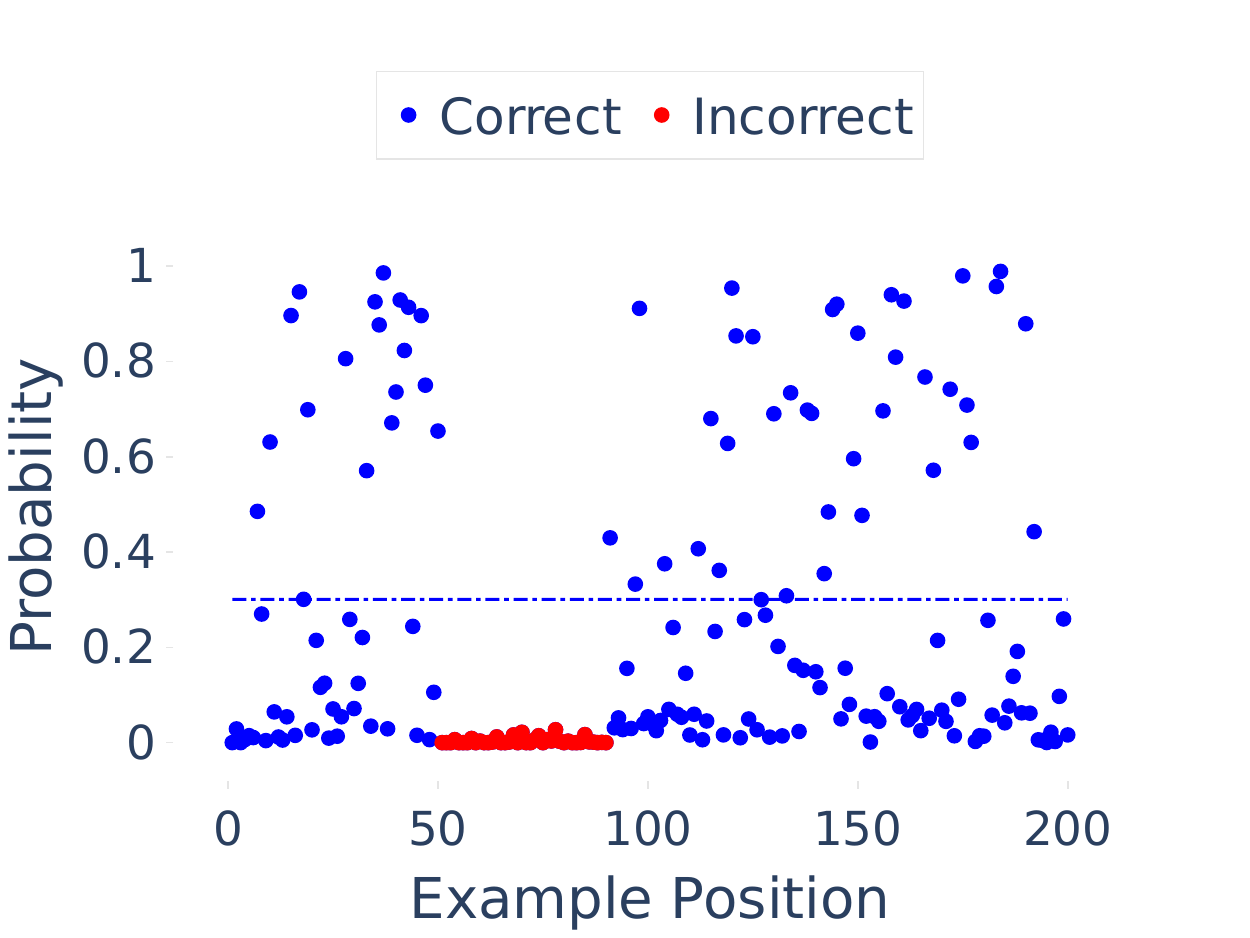}%
  \label{fig:falcon_incorrect_simultaneous}%
}
\hfill
\subfloat[Distributed incorrect examples]{%
  \includegraphics[width=0.33\textwidth]{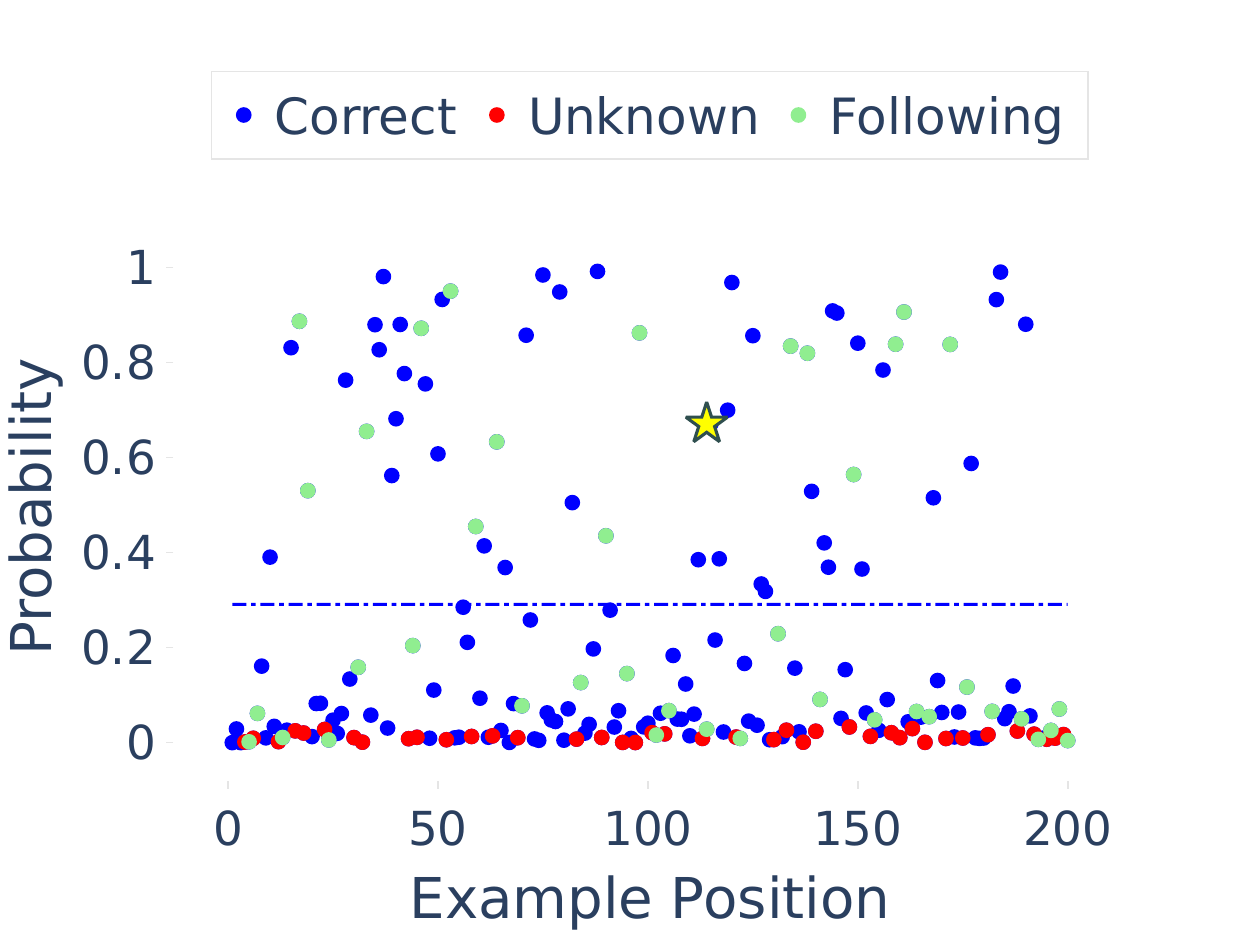}%
  \label{fig:falcon_unknown_distributed}%
}
\subfloat[Simultaneous incorrect examples]{%
  \includegraphics[width=0.33 \textwidth]{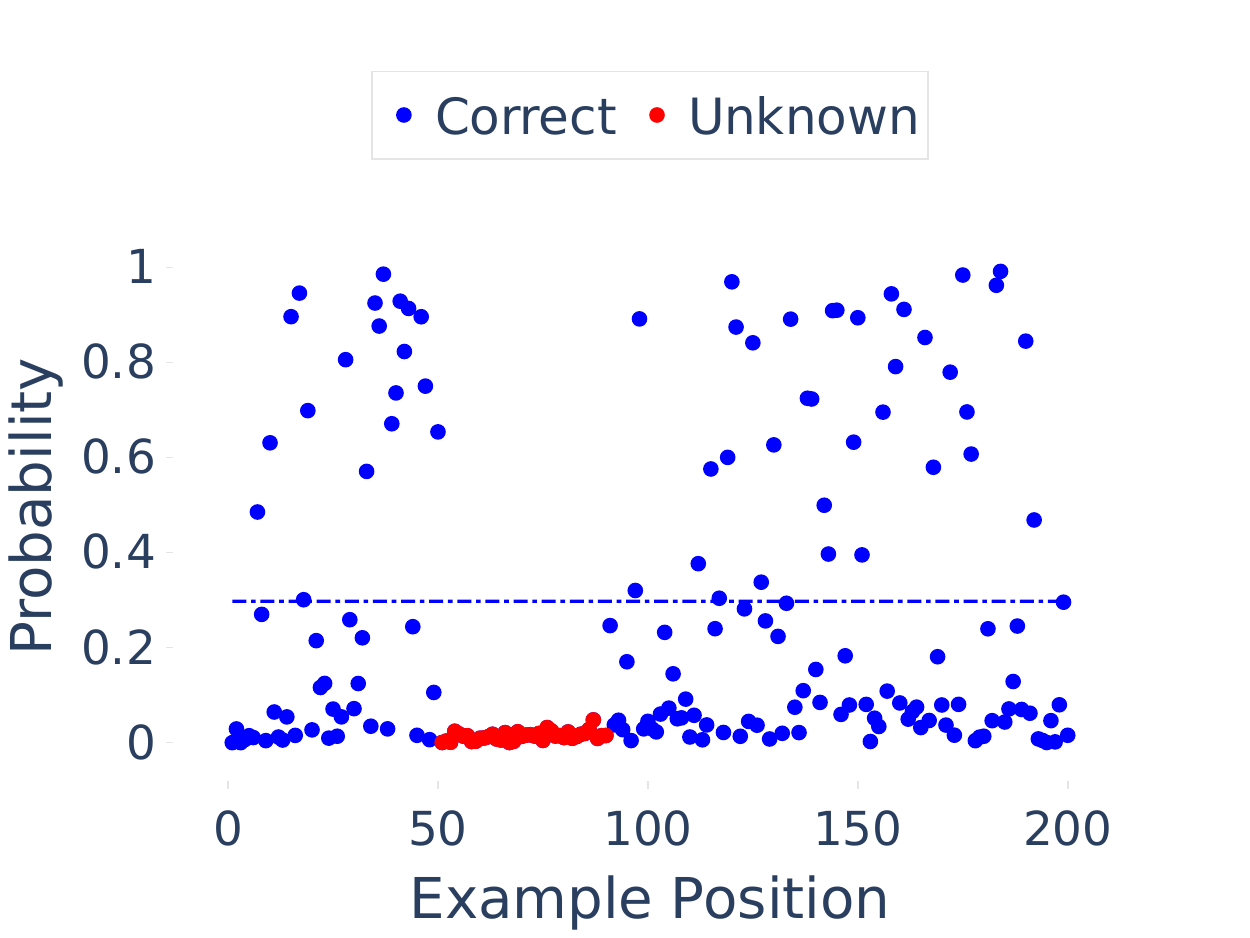}%
  \label{fig:falcon_unknown_simultaneous}%
}
\caption{\textbf{Analysis of object probability in one sequence of Nobel laureate data using Falcon-7B} 
}
\label{fig:falcon-7b-probabilities}
\end{figure}

We first consider $200$ correct examples (subject-object pairs) and report the \textit{absolute} generation probability of objects in corresponding examples. We showed the results for Llama2-7B, Falcon-7B, and Pythia-12B in Figure~\ref{fig:llama2-7b-probabilities}, Figure~\ref{fig:pyt-12b-probabilities} 
 and Figure~\ref{fig:falcon-7b-probabilities}. Figure~\ref{fig:llama_correct_sequence},  Figure~\ref{fig:pyt_correct_sequence} and Figure~\ref{fig:falcon_correct_sequence} illustrates the probability of each object at various sequence positions;  Figure~\ref{fig:llama_incorrect_distributed}, Figure~\ref{fig:pyt_incorrect_distributed} and Figure~\ref{fig:falcon_incorrect_distributed} shows the impact on probabilities after substituting 40 objects dispersed within the sequence with incorrect ones. Figure~\ref{fig:llama_incorrect_simultaneous}, Figure~\ref{fig:pyt_incorrect_simultaneous} and Figure~\ref{fig:falcon_incorrect_simultaneous} visualizes the effect of replacing objects at simultaneous positions. Figure~\ref{fig:llama_unknown_distributed}, Figure~\ref{fig:pyt_unknown_distributed}, Figures~\ref{fig:falcon_unknown_distributed},  Figure~\ref{fig:llama_unknown_simultaneous}, Figure~\ref{fig:pyt_unknown_simultaneous} and Figure~\ref{fig:falcon_unknown_simultaneous} present the outcomes of using unknown subject-object pairs as replacements. We used a horizontal dashed line showing an average probability of the correct examples. The yellow star marks the example at position 114 in the sequence.

\section{Details about the human-generated prompts and machine-mined prompts}

\label{appendix:hgp_mmp}

We list the used human-generated and machine-mined prompts from \citep{jiang2020can} in Table \ref{tab:hgp_mmp} with subjects denoted as <`subject'>.

\begin{table}[ht]
\centering
\caption{Templates for Selected Relations}
\label{tab:hgp_mmp}
\small{
\begin{tabular}{|c|c|c|c|}
\hline
\textbf{Relation Name} & \textbf{Index} & \textbf{HGP Template} & \textbf{MMP Template} \\ \hline
  & 1 & \{subject\} means & \{subject\} is a small \\
Instance of & 2 & \{subject\} is one & \{subject\} and liberal \\
  & 3 & \{subject\} is a & \{subject\} artist \\
\hline
& 1 & \{subject\} is playing music & \{subject\} series of \\
Genre & 2 & \{subject\} play & \{subject\} favorite \\
 & 3 & \{subject\} performs & \{subject\} is an american \\
\hline
 & 1 & \{subject\} plays in position & \{subject\} substitutions : \\
Position played on team / speciality & 2 & \{subject\} plays at position & \{subject\} substitutes : \\
 & 3& \{subject\} is in the position & \\
\hline
 & 1 & The original language of \{subject\} is & \{subject\} a. r. rahman \\
Original language of film/TV show & 2 & The source language of \{subject\} is & \\
 & 3 & The default language of \{subject\} is & \\
\hline
& 1 & The capital of \{subject\} is & \{subject\} united states embassy in \\
Capital & 2 & The capital city of \{subject\} is & \{subject\} representative legislature \\
 & 3 & Its capital \{subject\} is & \{subject\} rock band from \\
\hline
 & 1 & \{subject\} is a native language of & \{subject\} descent \\
Native language & 2 & The mother tongue of \{subject\} is & \{subject\} speak the \\
 & 3 & \{subject\} means & \{subject\} population or a widely spoken \\
\hline
 & 1 & \{subject\} is named after & \{subject\} and produces \\
Named after & 2 & \{subject\} is named for & \{subject\} variety of standard ) \\
 & 3 & \{subject\} is called after & \{subject\} official \\
\hline
 & 1& The official language \{subject\} is & \{subject\} professor of \\
Official language & 2 & \{subject\} is & \{subject\} is the official language in \\
& 3 & \{subject\} is officially & \{subject\} is the official language spoken in \\
\hline
  & 1 & \{subject\} is developed by & \{subject\} was developed by \\
Developer & 2 & \{subject\} is created by & \{subject\} 2008 \\
 & 2 & \{subject\} is designed by & \{subject\} references external links \\
\hline
 & 1 & \{subject\} was originally aired on & \{subject\} premiered on \\
Original broadcaster & 2 & \{subject\} was originally broadcast on & \{subject\} aired on \\
 & 3 & \{subject\} was originally shown in & \{subject\} 2021 \\
\hline
 & 1 & in order to communicate in  & \  \\
Language spoken, written or signed & 2 & \{subject\} used to communicate in &  \\
 & 3 &  & \\
\hline
 & 1 & \{subject\} is represented by music label & \{subject\} attributed to the \\
Manufacturer & 2 & \{subject\} is represented by the record label & \{subject\} 113 \\
 & 3 & \{subject\} is represented by & \{subject\} cedar point \\ \hline
\end{tabular}}
\end{table}
\clearpage

\section{Additional results}
\label{appendix:all_eval_results}
\subsection{Model Name Simplification}
\label{appendix:model_name_simplification}
We list all the models and their simplified names we evaluated in the paper in Table~\ref{table:all_models_name}.

\begin{table}[ht]
\centering
\caption{Model Name Simplifications}
\begin{tabular}{|c|c|}
\hline
\textbf{Original Name} & \textbf{Simplified Name in Paper} \\ \hline
\href{https://huggingface.co/mistralai/Mixtral-8x7B-v0.1}{mistral-mixtral-8x7B-v0.1} & Mixtral-8x7B  \\ \hline
\href{https://huggingface.co/NousResearch/Nous-Hermes-2-Mixtral-8x7B-SFT}{Nous-Hermes-2-Mixtral-8x7B-SFT} & Mixtral-8x7B-FT1 \\ \hline
\href{https://huggingface.co/NousResearch/Nous-Hermes-2-Mixtral-8x7B-SFT}{Nous-Hermes-2-Mixtral-8x7B-DPO} & Mixtral-8x7B-FT2 \\ \hline
\href{https://huggingface.co/mistralai/Mistral-7B-v0.1}{mistral-7b} & Mistral-7B \\ \hline
\href{https://huggingface.co/mistralai/Mistral-7B-Instruct-v0.1}{mistral-instruct-7b} & Mistral-7B-FT1 \\ \hline
\href{https://huggingface.co/teknium/OpenHermes-2.5-Mistral-7B}{openhermes-2.5-mistral-7b} & Mistral-7B-FT2 \\ \hline
\href{https://huggingface.co/meta-llama/Llama-2-70b}{llama2-70b} & Llama2-70B \\ \hline
\href{https://huggingface.co/meta-llama/Llama-2-70b-chat}{llama2-70b-chat} & Llama2-70B-FT1 \\ \hline
\href{https://huggingface.co/meta-llama/Llama-2-13b}{llama2-13b} & Llama2-13B \\ \hline
\href{https://huggingface.co/meta-llama/Llama-2-13b-chat}{llama2-13b-chat} & Llama2-13B-FT1 \\ \hline
\href{https://huggingface.co/lmsys/vicuna-13b-v1.5}{vicuna-13b-v1.5} & Llama2-13B-FT2 \\ \hline
\href{https://huggingface.co/meta-llama/Llama-2-7b}{llama2-7b} & Llama2-7B \\ \hline
\href{https://huggingface.co/meta-llama/Llama-2-7b-chat}{llama2-7b-chat} & Llama2-7B-FT1 \\ \hline
\href{https://huggingface.co/lmsys/vicuna-7b-v1.5}{vicuna-7b-v1.5} & Llama2-7B-FT2 \\ \hline
\href{https://huggingface.co/google/gemma-7b}{gemma-7b} & Gemma-7B \\ \hline
\href{https://huggingface.co/google/gemma-7b-it}{gemma-7b-it} & Gemma-7B-FT1 \\ \hline
\href{https://huggingface.co/google/gemma-2b}{gemma-2b} & Gemma-2B \\ \hline
\href{https://huggingface.co/google/gemma-2b-it}{gemma-2b-it} & Gemma-2B-FT1 \\ \hline
\href{https://huggingface.co/huggyllama/llama-65b}{llama-65b} & Llama-65B \\ \hline
\href{https://huggingface.co/huggyllama/llama-33b}{llama-33b} & Llama-33B \\ \hline
\href{https://huggingface.co/huggyllama/llama-13b}{llama-13b} & Llama-13B \\ \hline
\href{https://huggingface.co/lmsys/vicuna-13b-v1.3}{vicuna-13b-1.3} & Llama-13B-FT1 \\ \hline
\href{https://huggingface.co/huggyllama/llama-7b}{llama-7b} & Llama-7B \\ \hline
\href{https://huggingface.co/lmsys/vicuna-7b-v1.3}{vicuna-7b-1.3} & Llama-7B-FT1 \\ \hline
\href{https://huggingface.co/tiiuae/falcon-7b}{falcon-7b} & Falcon-7B \\ \hline
\href{https://huggingface.co/tiiuae/falcon-instruct-7b}{falcon-instruct-7b} & Falcon-7B-FT1 \\ \hline
\href{https://huggingface.co/mosaicml/mpt-7b}{mpt-7b} & MPT-7B \\ \hline
\href{https://huggingface.co/EleutherAI/gpt-neox-20b}{gpt-neox-20b} & GPT-NEOX-20B \\ \hline
\href{https://huggingface.co/facebook/opt-30b}{opt-30b} & OPT-30B \\ \hline
\href{https://huggingface.co/facebook/opt-13b}{opt-13b} & OPT-13B \\ \hline
\href{https://huggingface.co/facebook/opt-6.7b}{opt-6.7b} & OPT-6.7B \\ \hline
\href{https://huggingface.co/facebook/opt-2.7b}{opt-2.7b} & OPT-2.7B \\ \hline
\href{https://huggingface.co/facebook/opt-1.3b}{opt-1.3b} & OPT-1.3B \\ \hline
\href{https://huggingface.co/facebook/opt-350m}{opt-350m} & OPT-350M \\ \hline
\href{https://huggingface.co/facebook/opt-125m}{opt-125m} & OPT-125M \\ \hline
\href{https://huggingface.co/EleutherAI/gpt-j-6b}{gpt-j-6b} & GPT-J-6B \\ \hline
\href{https://huggingface.co/EleutherAI/pythia-12b}{pythia-12b} & Pythia-12B \\ \hline
\href{https://huggingface.co/EleutherAI/pythia-6.9b}{pythia-6.9b} & Pythia-6.9B \\ \hline
\href{https://huggingface.co/EleutherAI/pythia-2.8b}{pythia-2.8b} & Pythia-2.8B \\ \hline
\href{https://huggingface.co/EleutherAI/pythia-1.4b}{pythia-1.4b} & Pythia-1.4B \\ \hline
\href{https://huggingface.co/EleutherAI/pythia-1b}{pythia-1b} & Pythia-1B \\ \hline
\href{https://huggingface.co/EleutherAI/pythia-410m}{pythia-410m}& Pythia-410M \\ \hline
\href{https://huggingface.co/EleutherAI/pythia-160m}{pythia-160m} & Pythia-160M \\ \hline
\href{https://huggingface.co/EleutherAI/pythia-70m}{pythia-70m}& Pythia-70M \\ \hline
\href{https://huggingface.co/bigscience/bloomz-7b1}{bloom-7.1b} & Bloom-7.1B \\ \hline
\href{https://huggingface.co/bigscience/bloomz-3b}{bloom-3b}& Bloom-3B \\ \hline
\href{https://huggingface.co/bigscience/bloomz-1b7}{bloom-1.7b} & Bloom-1.7B \\ \hline
\href{https://huggingface.co/bigscience/bloomz-1b1}{bloom-1.1b} & Bloom-1.1B \\ \hline
\href{https://huggingface.co/bigscience/bloom-560m}{bloom-560m}& Bloom-560M \\ \hline
\end{tabular}
\label{table:all_models_name}
\end{table}

\subsection{Additional results on baseline comparison}
\label{appendix:baseline_comp}

\begin{figure}[h!]
\centering
\hfill
\subfloat[Relation: instance of]{%
  \includegraphics[width=0.33\textwidth]{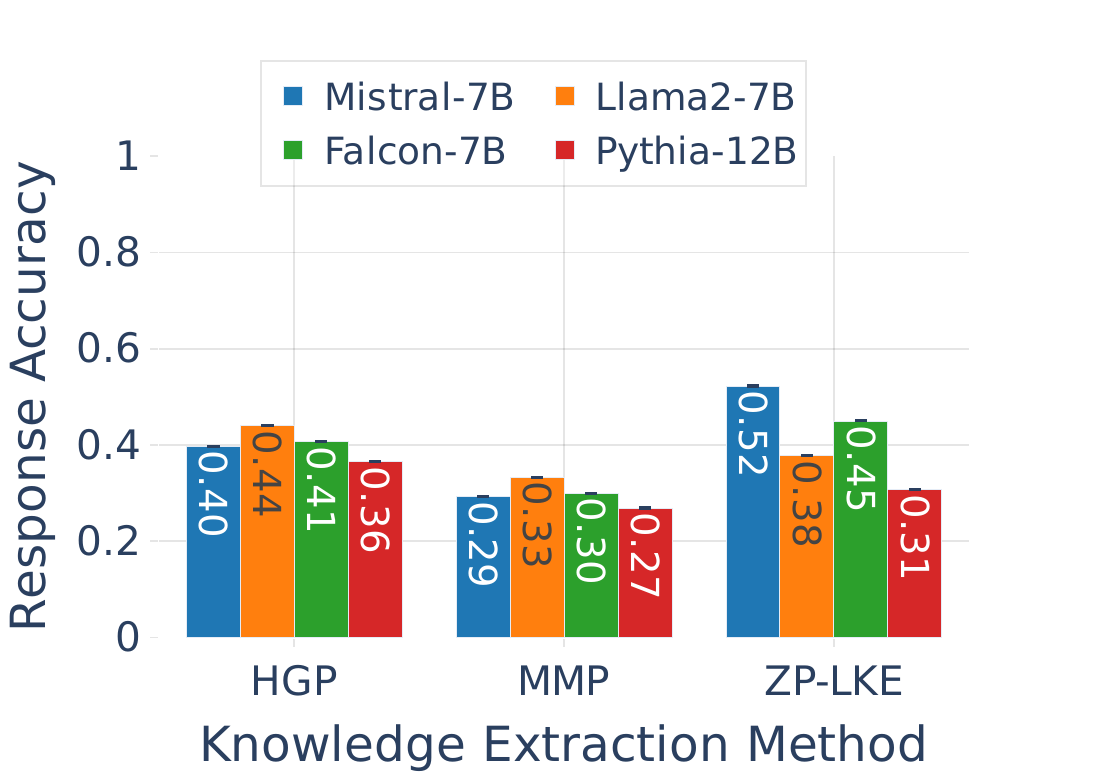}%
  \label{fig:gen_instance_of}%
}
\hfill
\subfloat[Relation: genre]{%
  \includegraphics[width=0.33\textwidth]{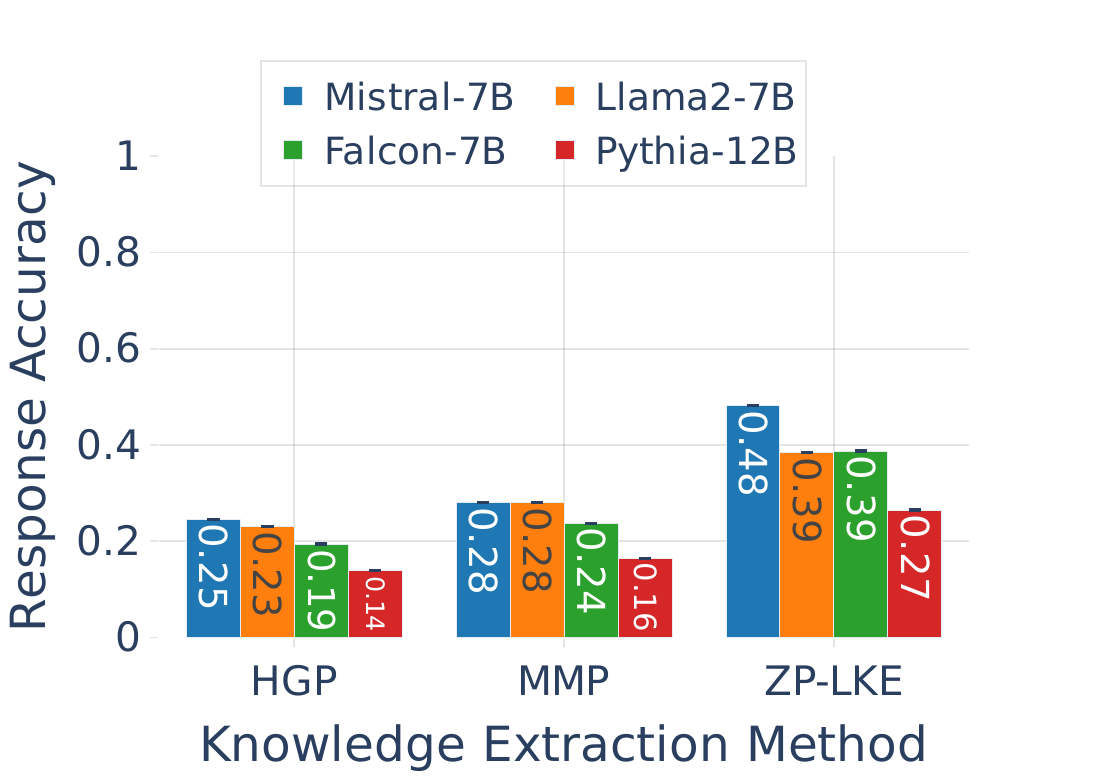}%
  \label{fig:gen_genre}%
}
\subfloat[Relation: language spoken, written or signed]{%
  \includegraphics[width=0.33\textwidth]{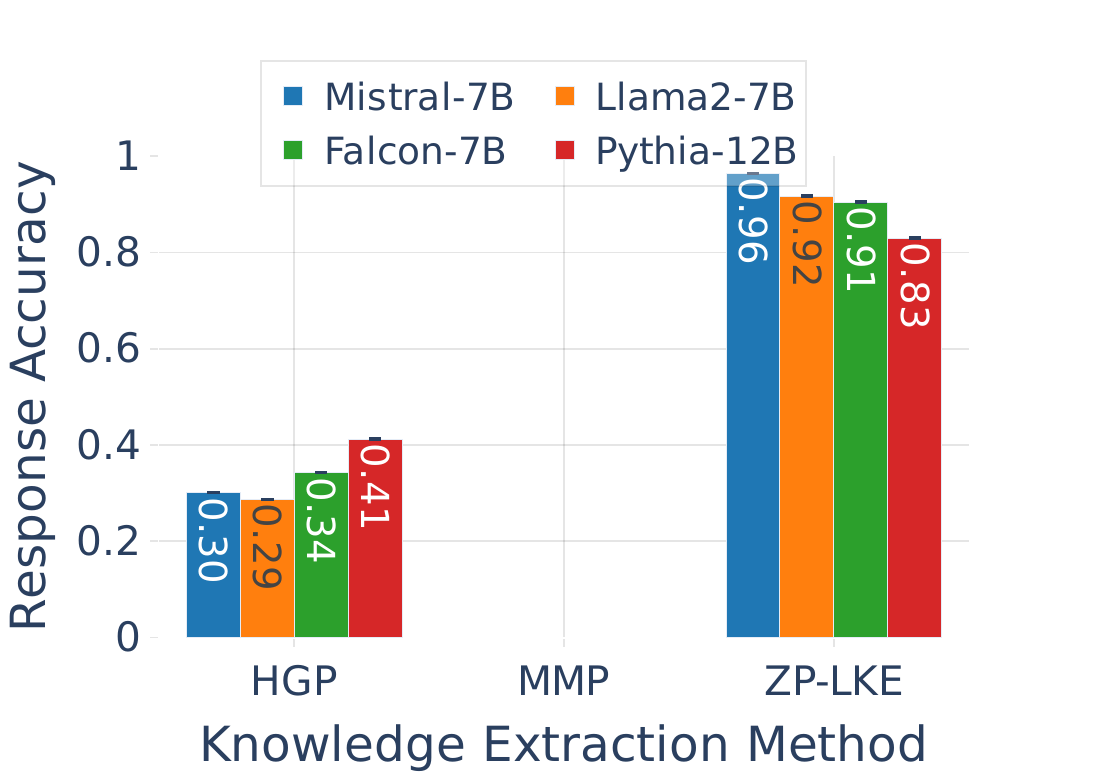}%
  \label{fig:language spoken, written or signed}%
}
\hfill
\subfloat[Relation: position played on team / speciality]{%
  \includegraphics[width=0.33\textwidth]{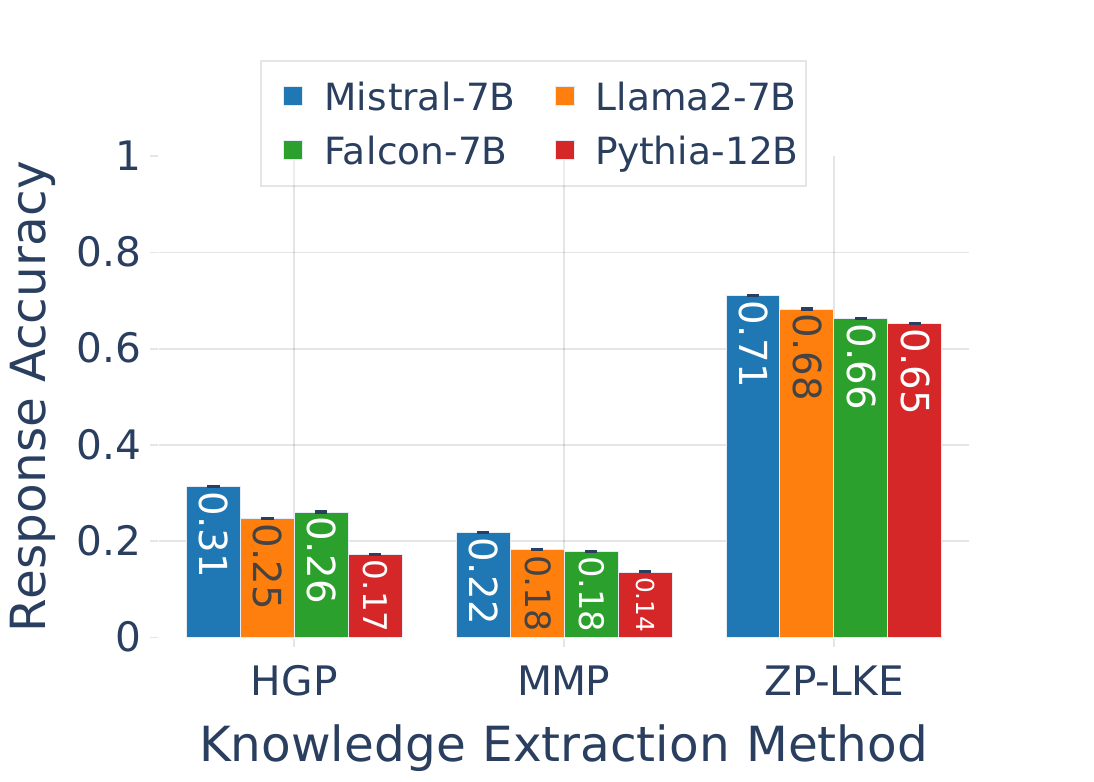}%
  \label{fig:position played on team / speciality}%
}
\subfloat[Relation: original language of film/TV show]{%
  \includegraphics[width=0.33 \textwidth]{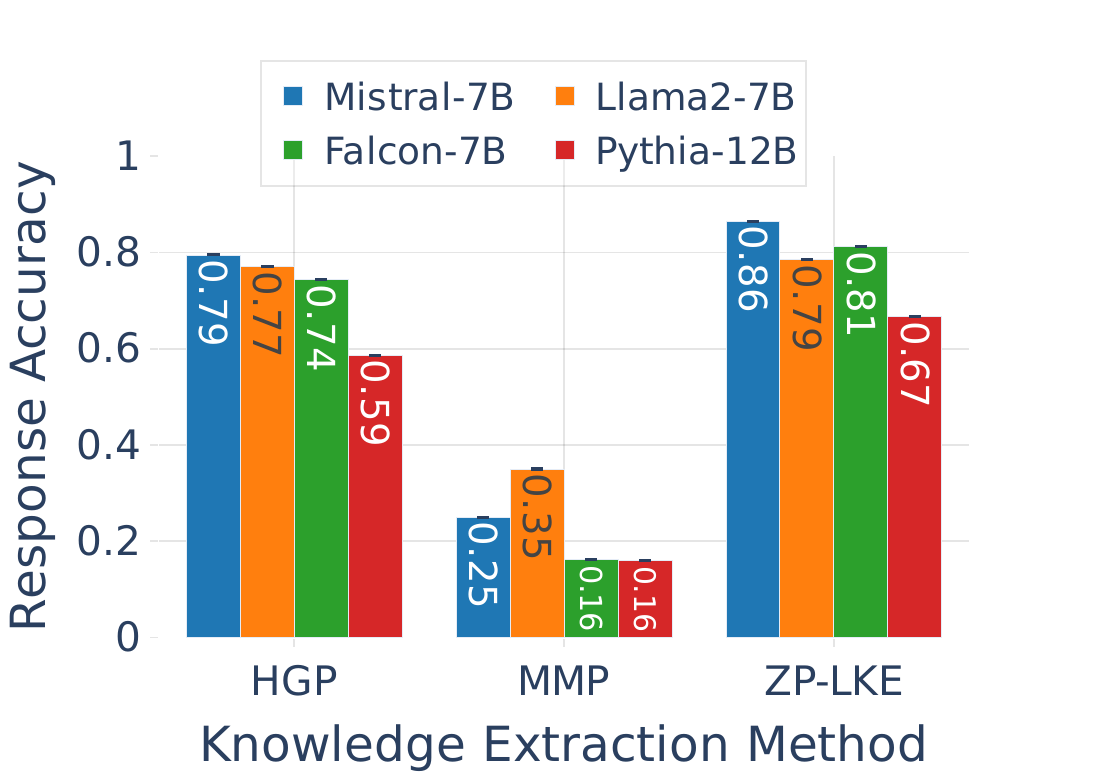}%
  \label{fig:original language of film/TV show}%
}
\subfloat[Relation: capital]{%
  \includegraphics[width=0.33 \textwidth]{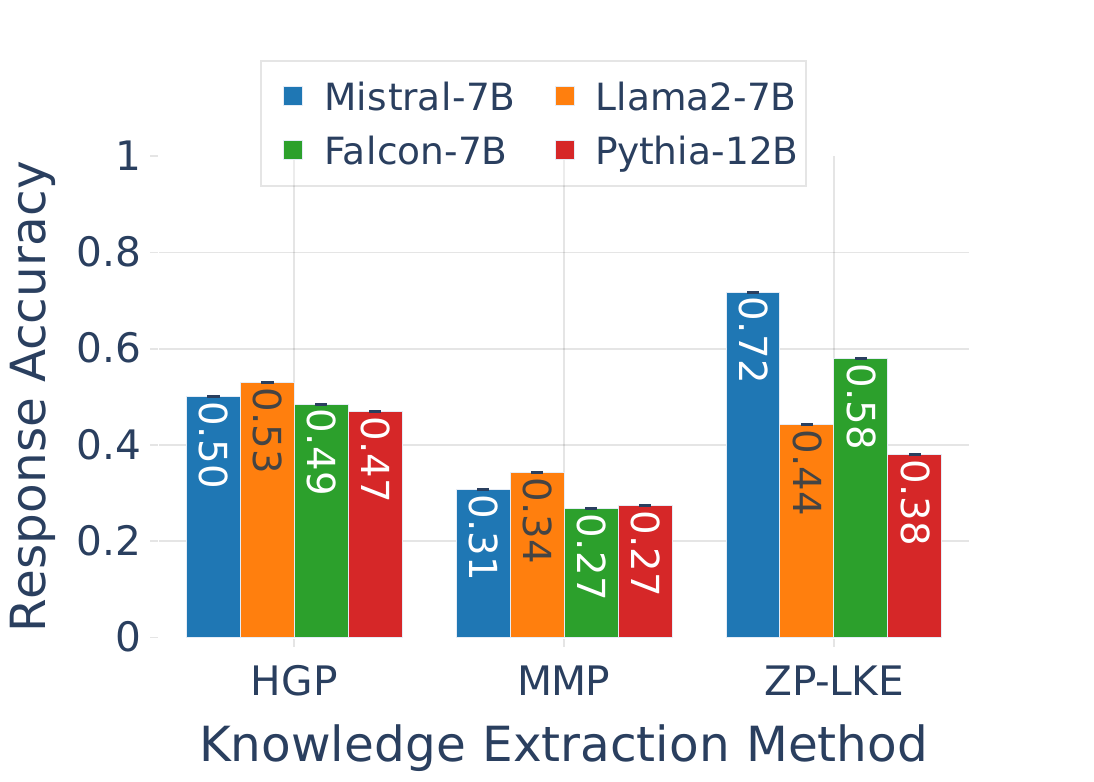}%
  \label{fig:capital}%
}
\hfill
\subfloat[Relation: native language]{%
  \includegraphics[width=0.33 \textwidth]{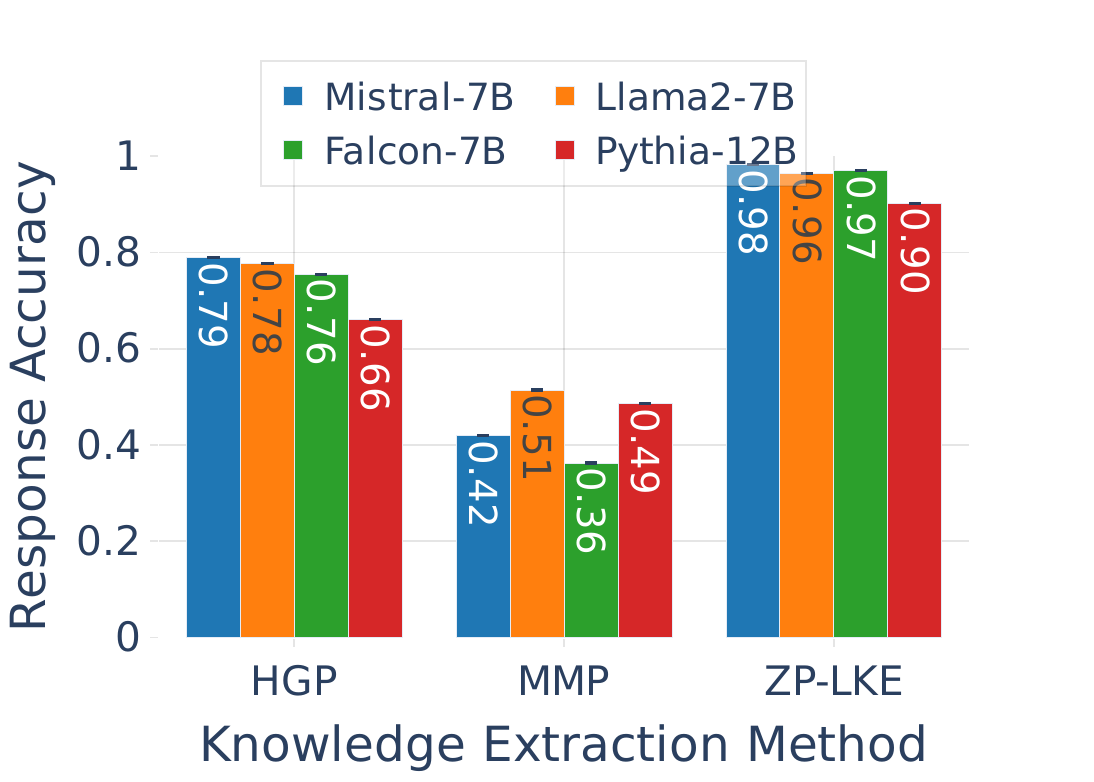}%
  \label{fig:native language}%
}
\subfloat[Relation: named after]{%
  \includegraphics[width=0.33 \textwidth]{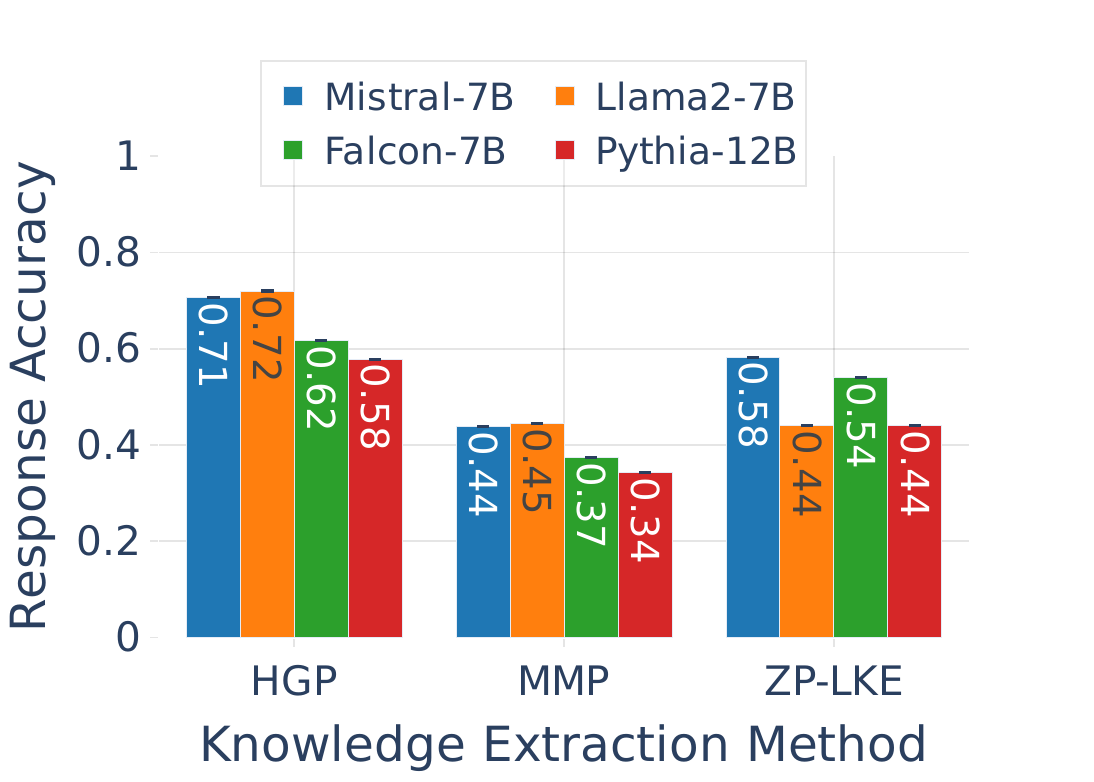}%
  \label{fig:native language}%
}
\subfloat[Relation: official language]{%
  \includegraphics[width=0.33 \textwidth]{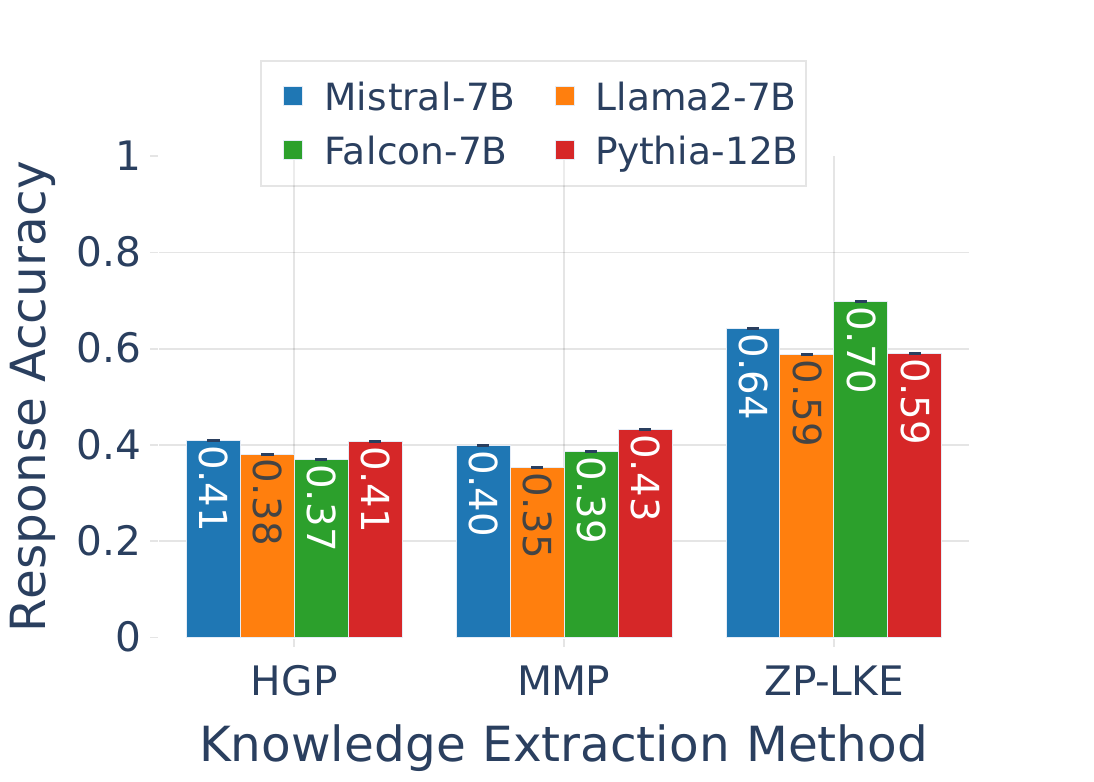}%
  \label{fig:official language}%
}
\hfill
\subfloat[Relation: developer]{%
  \includegraphics[width=0.33 \textwidth]{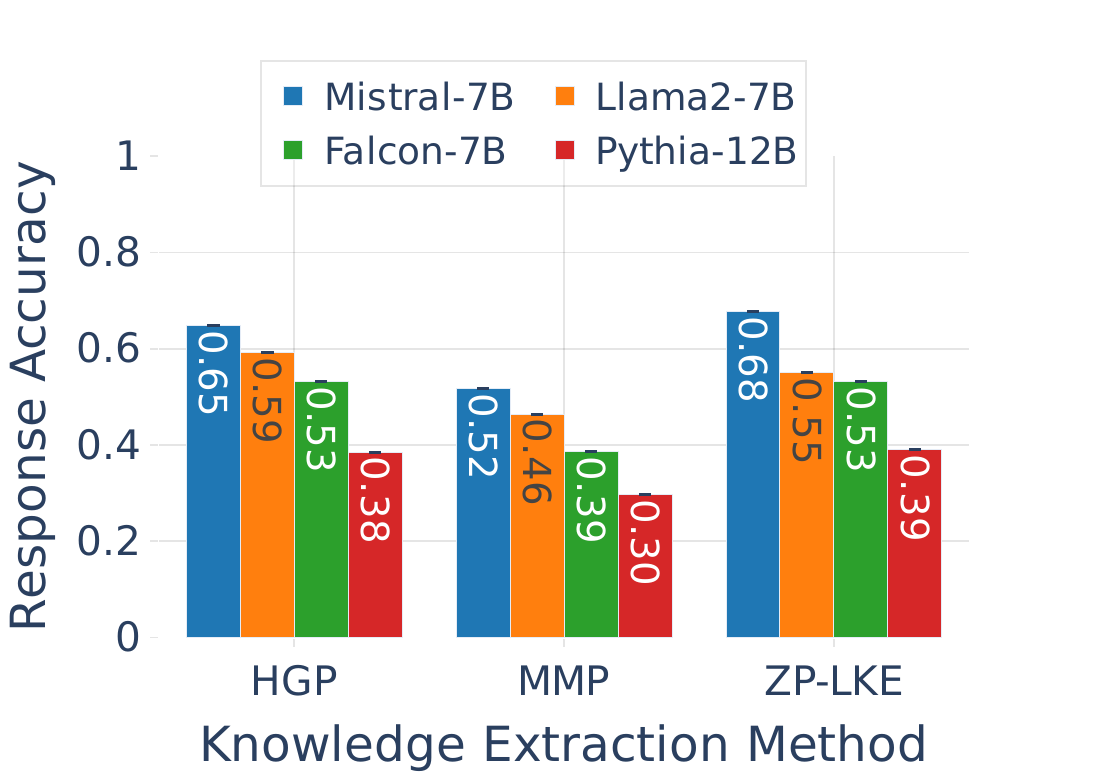}%
  \label{fig:developer}%
}
\subfloat[Relation: original broadcaster]{%
  \includegraphics[width=0.33 \textwidth]{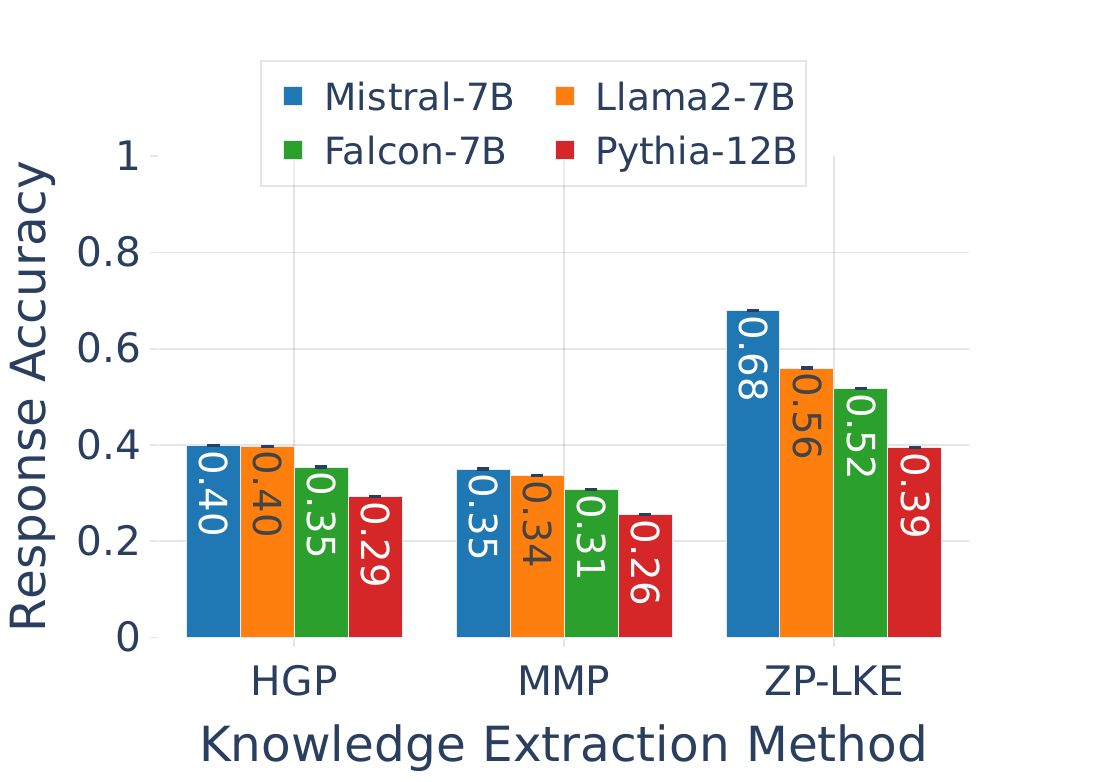}%
  \label{fig:original broadcaster}%
}
\subfloat[Relation: manufacturer]{%
  \includegraphics[width=0.33 \textwidth]{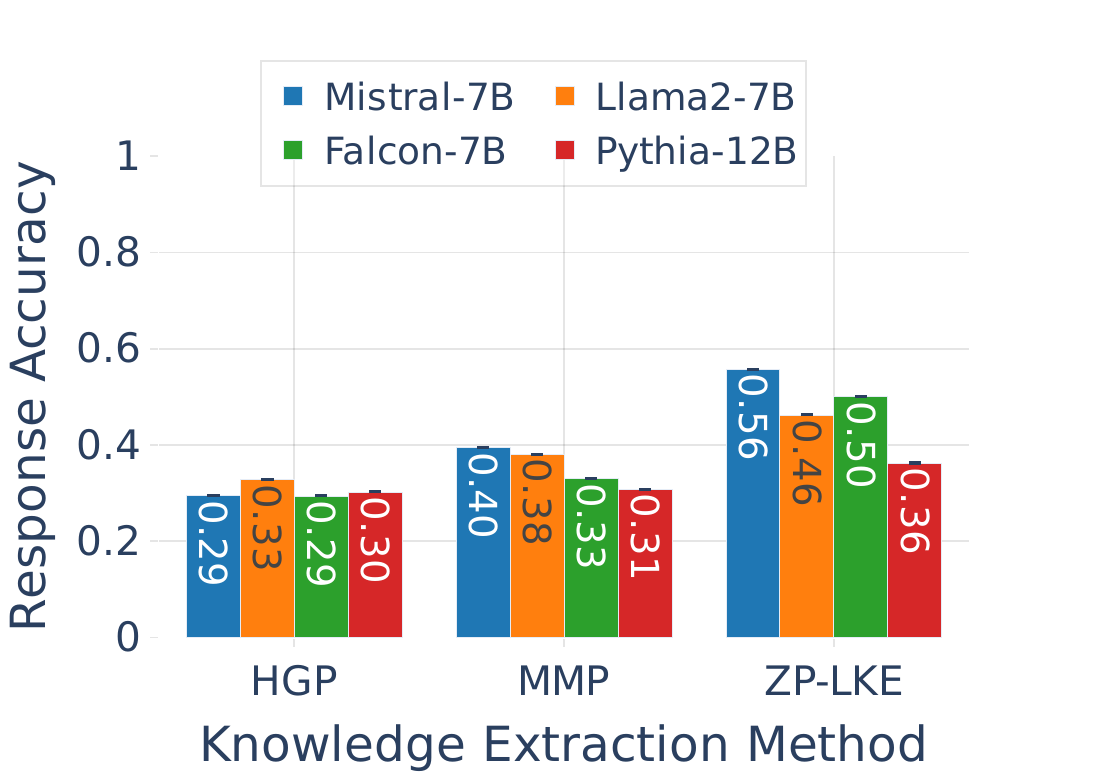}%
  \label{fig:manufacturer}%
}
\caption{Response accuracy across all the 12 relations} 
\label{fig:respons_acc_all}
\end{figure}

\begin{figure}[h!]
\centering
\hfill
\subfloat[Relation: instance of]{%
  \includegraphics[width=0.33\textwidth]{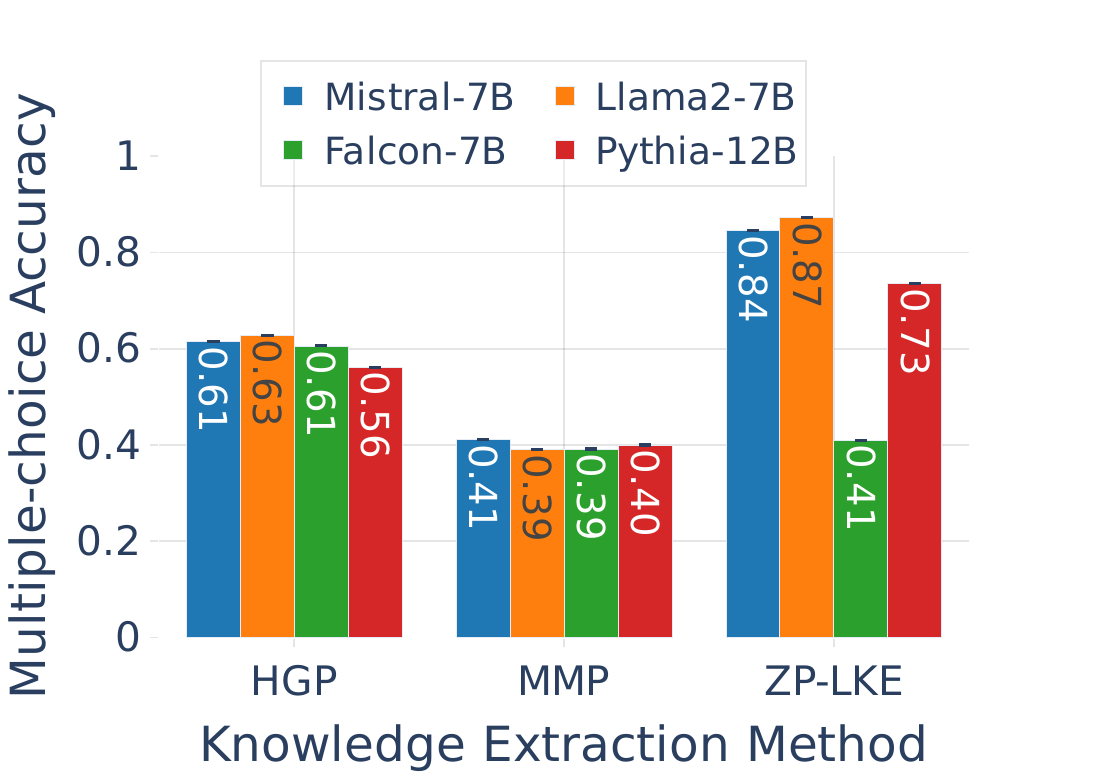}%
  \label{fig:gen_instance_of}%
}
\hfill
\subfloat[Relation: genre]{%
  \includegraphics[width=0.33\textwidth]{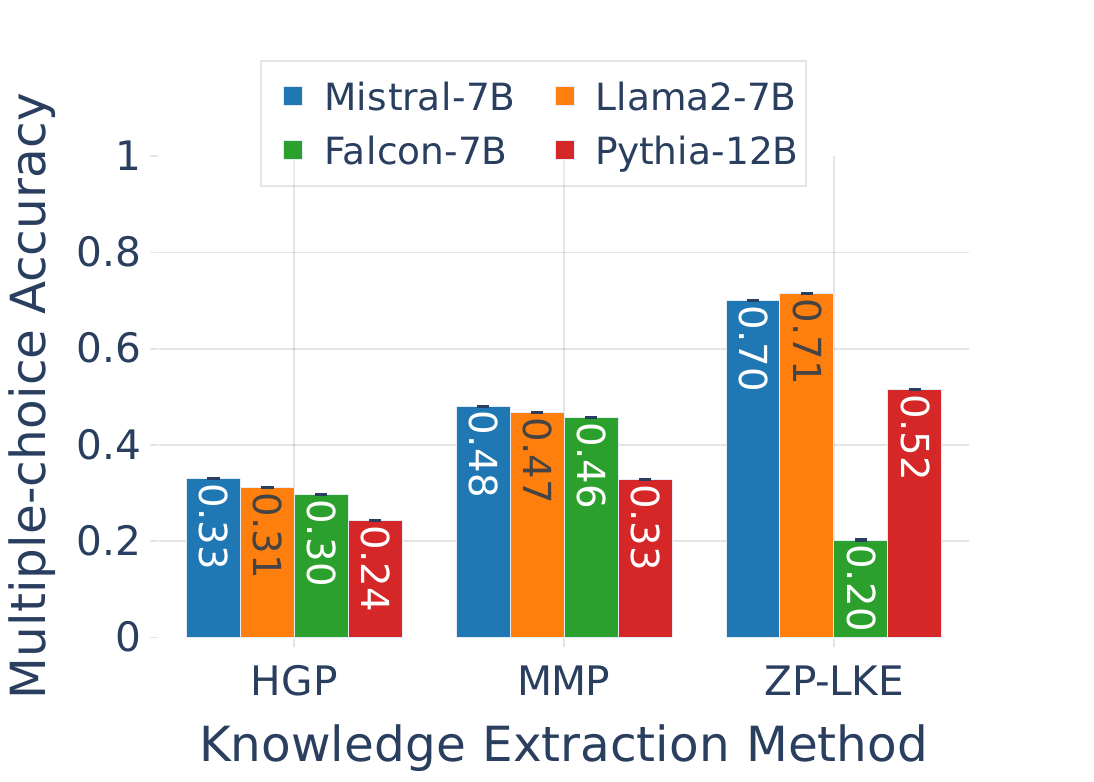}%
  \label{fig:gen_genre}%
}
\subfloat[Relation: language spoken, written or signed]{%
  \includegraphics[width=0.33\textwidth]{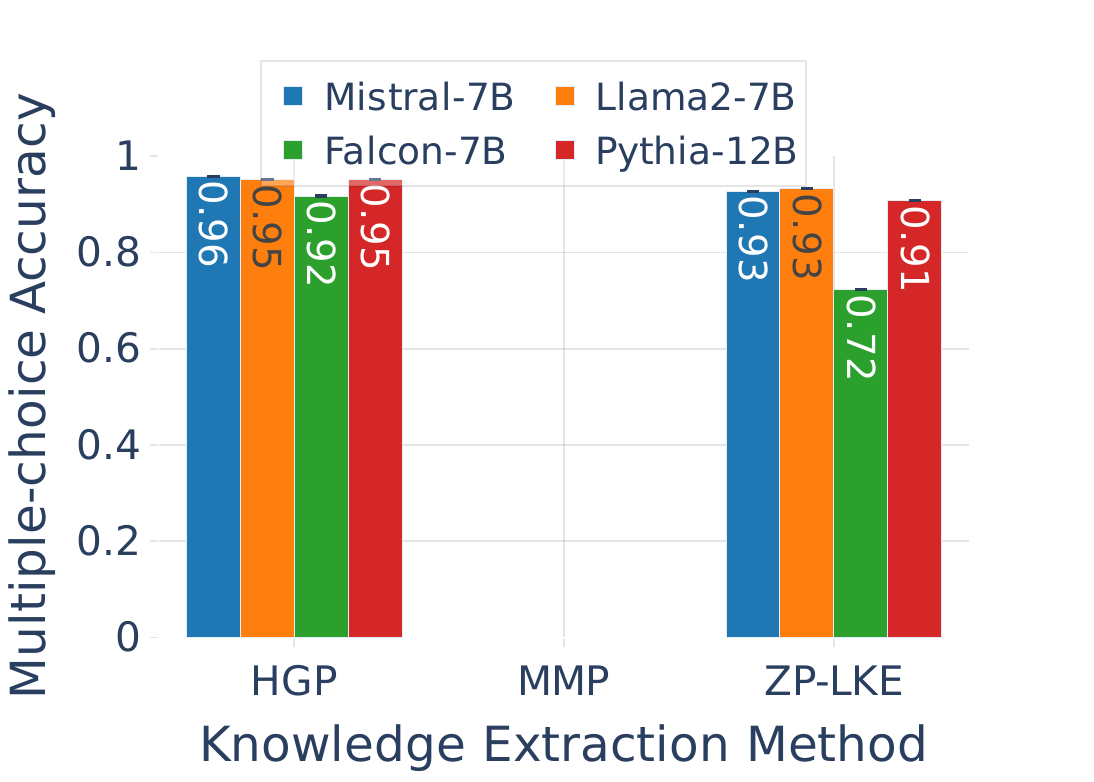}%
  \label{fig:language spoken, written or signed}%
}
\hfill
\subfloat[Relation: position played on team / speciality]{%
  \includegraphics[width=0.33\textwidth]{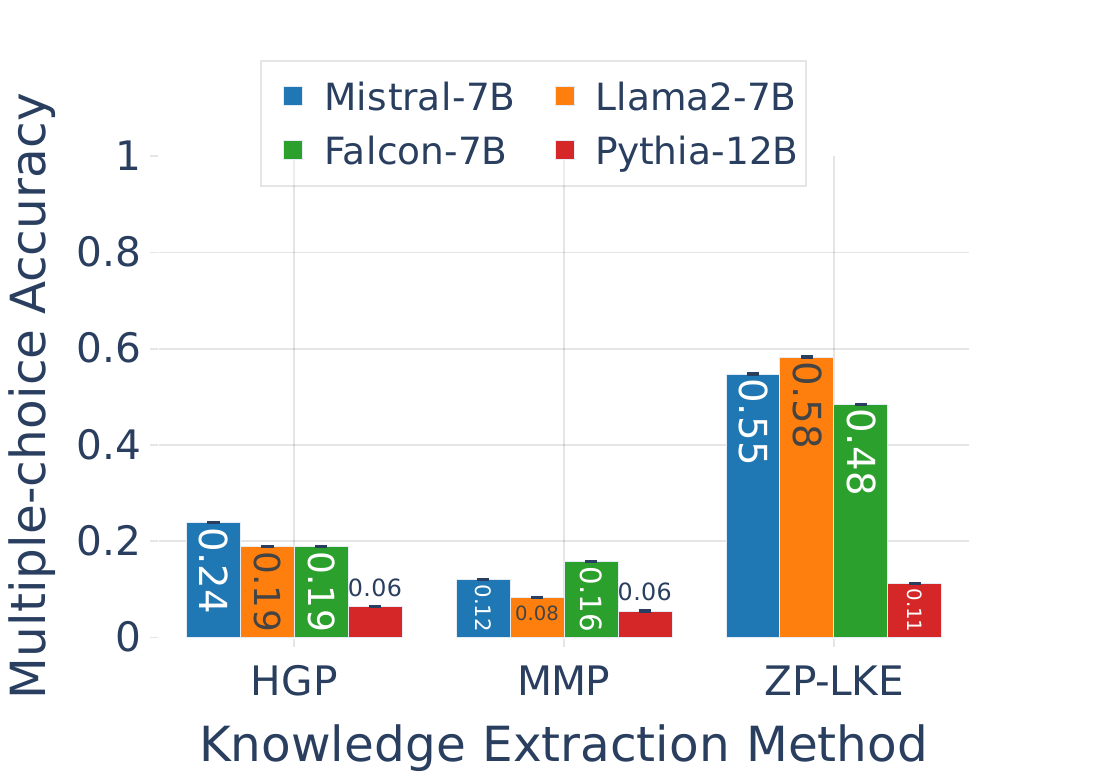}%
  \label{fig:position played on team / speciality}%
}
\subfloat[Relation: original language of film/TV show]{%
  \includegraphics[width=0.33 \textwidth]{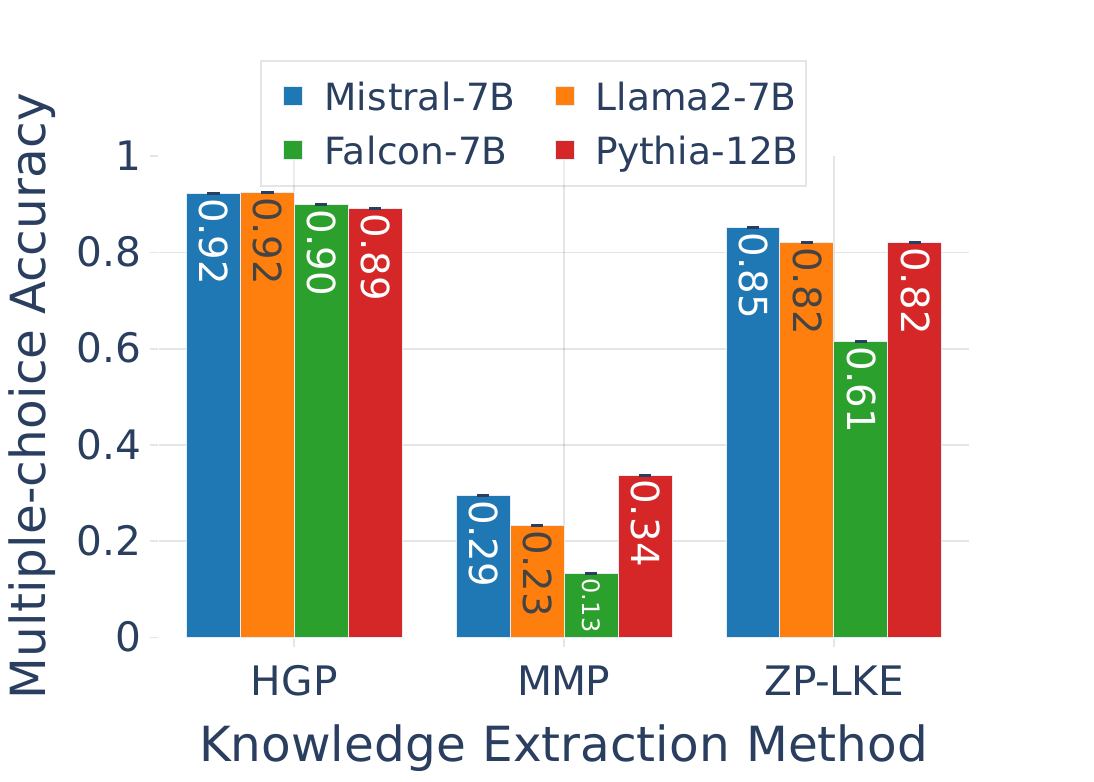}%
  \label{fig:original language of film/TV show}%
}
\subfloat[Relation: capital]{%
  \includegraphics[width=0.33 \textwidth]{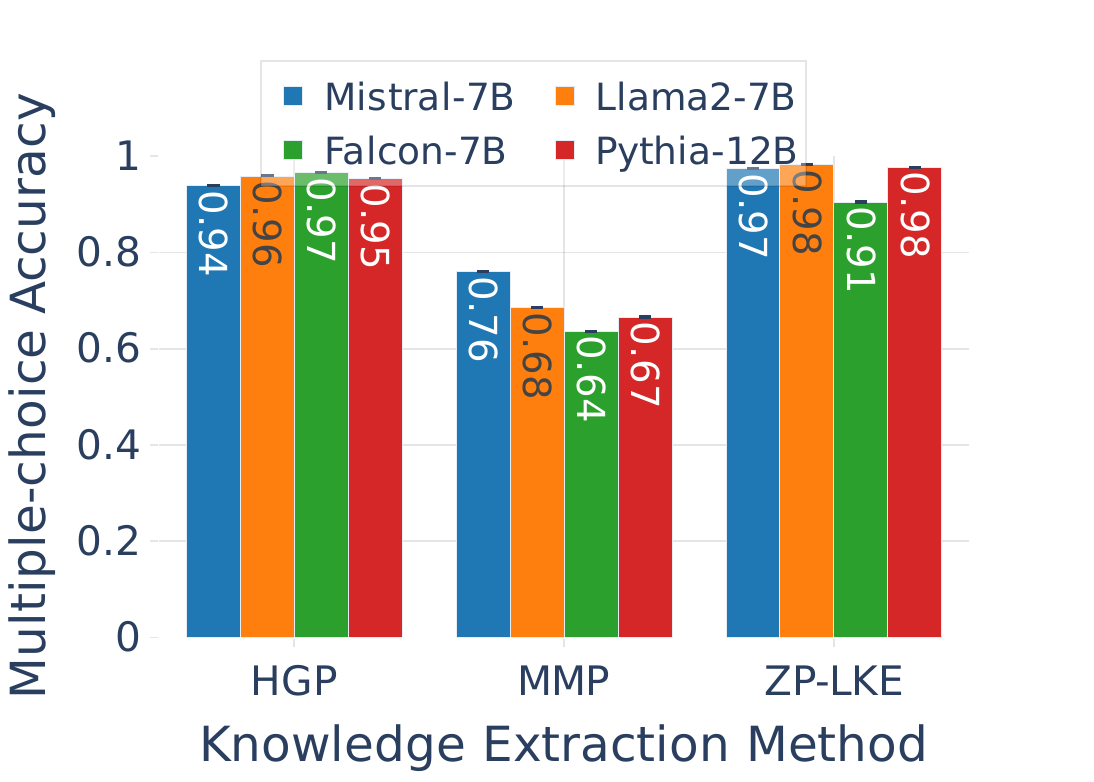}%
  \label{fig:capital}%
}
\hfill
\subfloat[Relation: native language]{%
  \includegraphics[width=0.33 \textwidth]{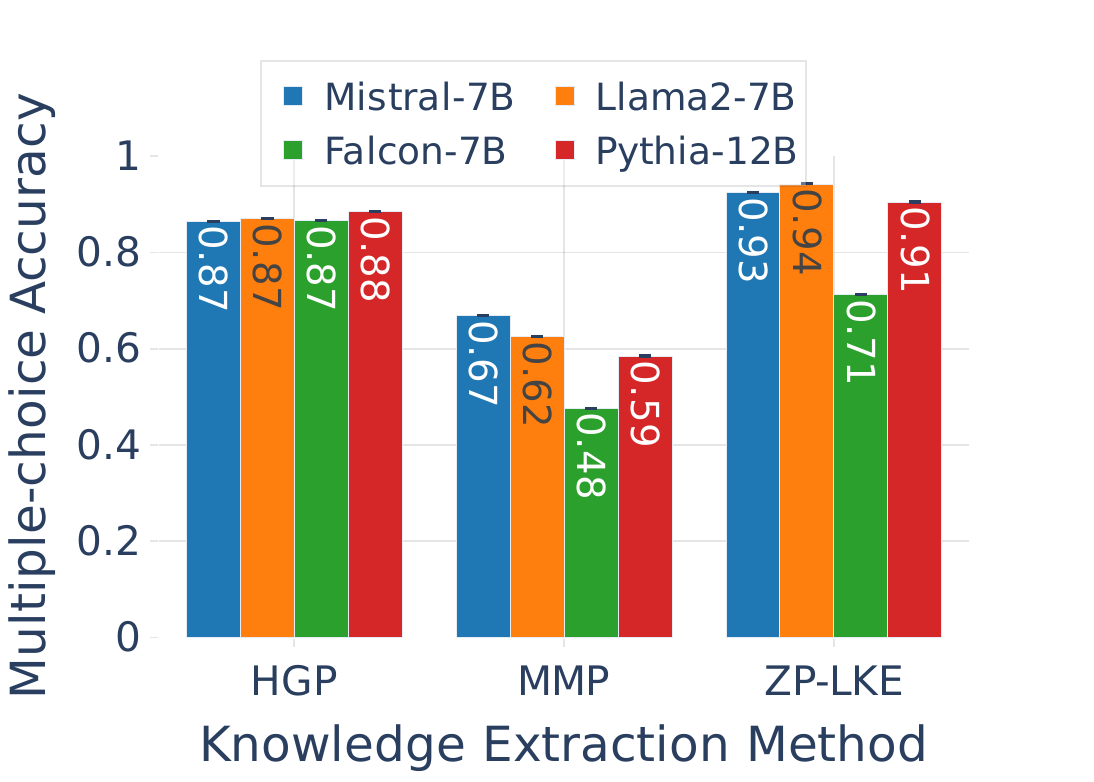}%
  \label{fig:native language}%
}
\subfloat[Relation: named after]{%
  \includegraphics[width=0.33 \textwidth]{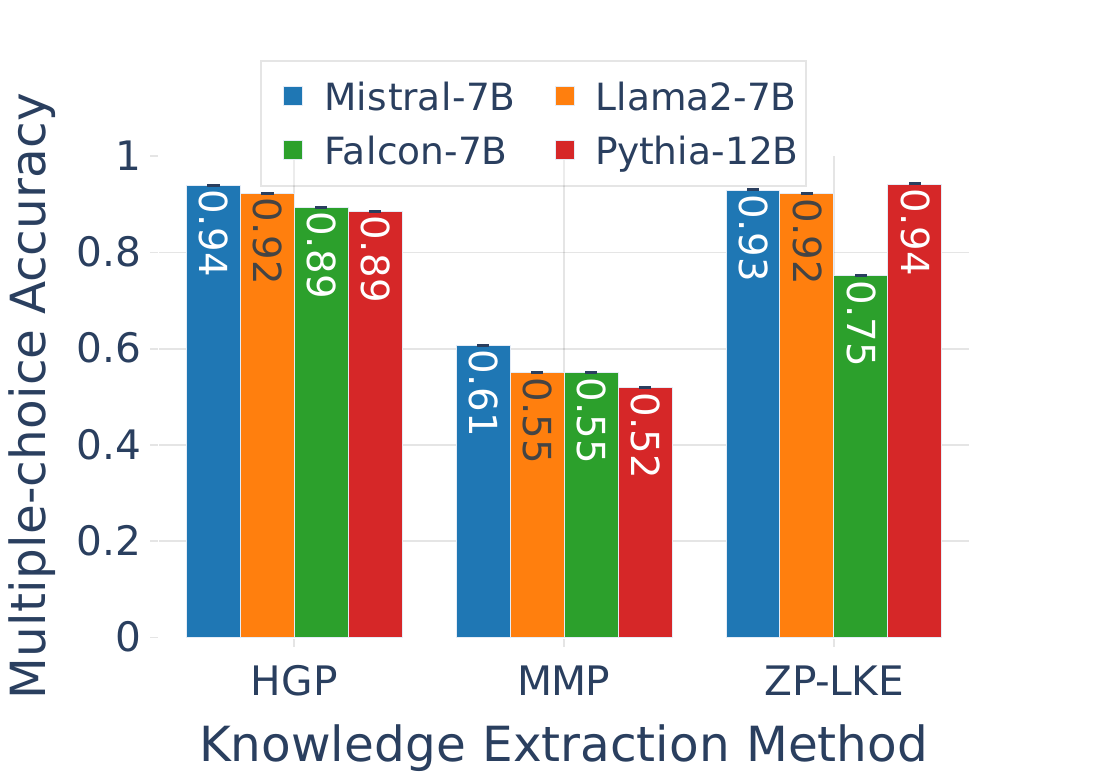}%
  \label{fig:native language}%
}
\subfloat[Relation: official language]{%
  \includegraphics[width=0.33 \textwidth]{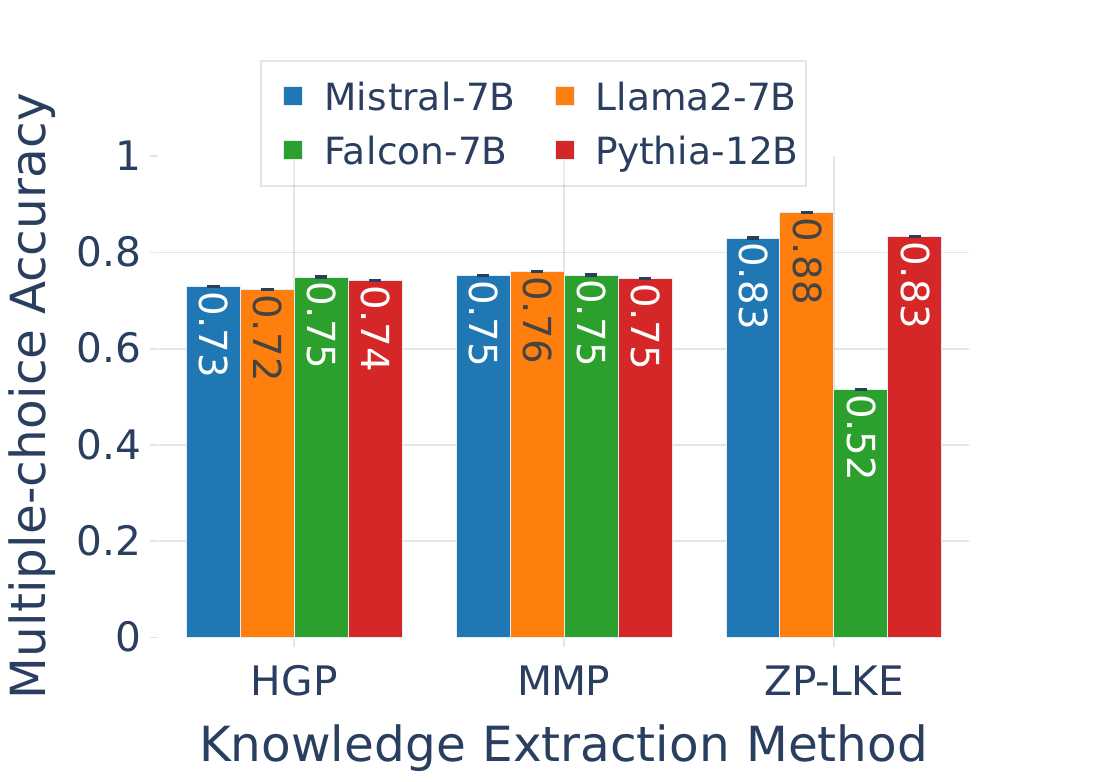}%
  \label{fig:official language}%
}
\hfill
\subfloat[Relation: developer]{%
  \includegraphics[width=0.33 \textwidth]{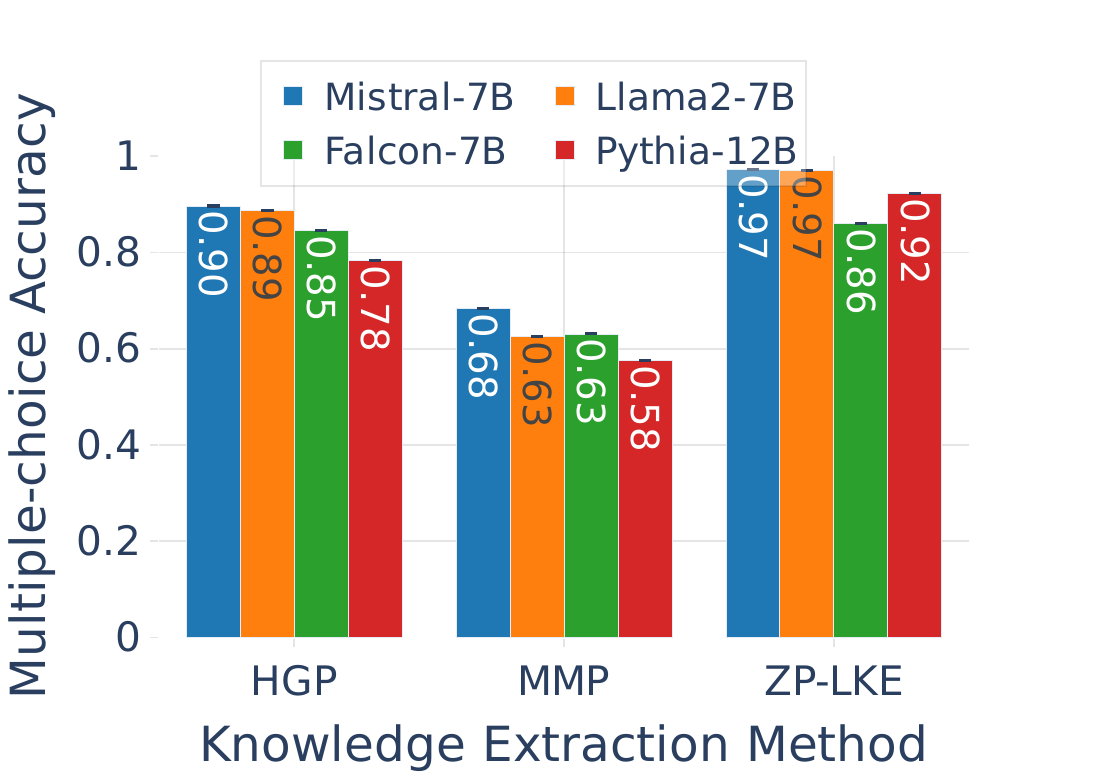}%
  \label{fig:developer}%
}
\subfloat[Relation: original broadcaster]{%
  \includegraphics[width=0.33 \textwidth]{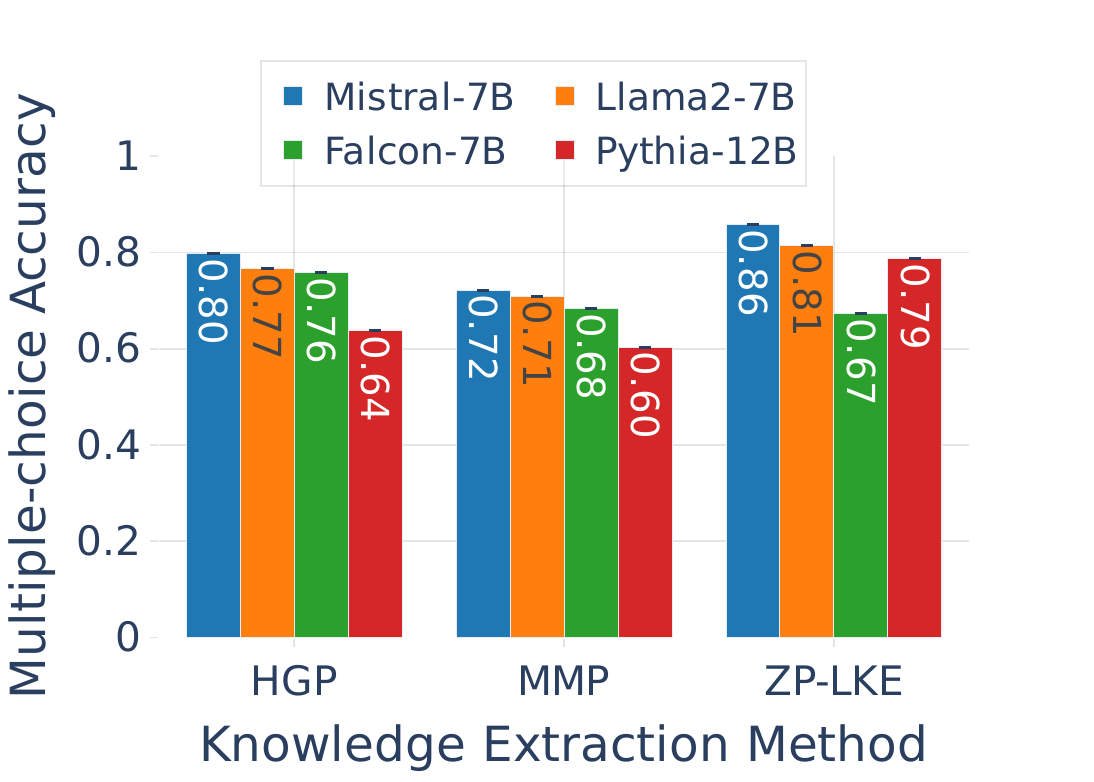}%
  \label{fig:original broadcaster}%
}
\subfloat[Relation: manufacturer]{%
  \includegraphics[width=0.33 \textwidth]{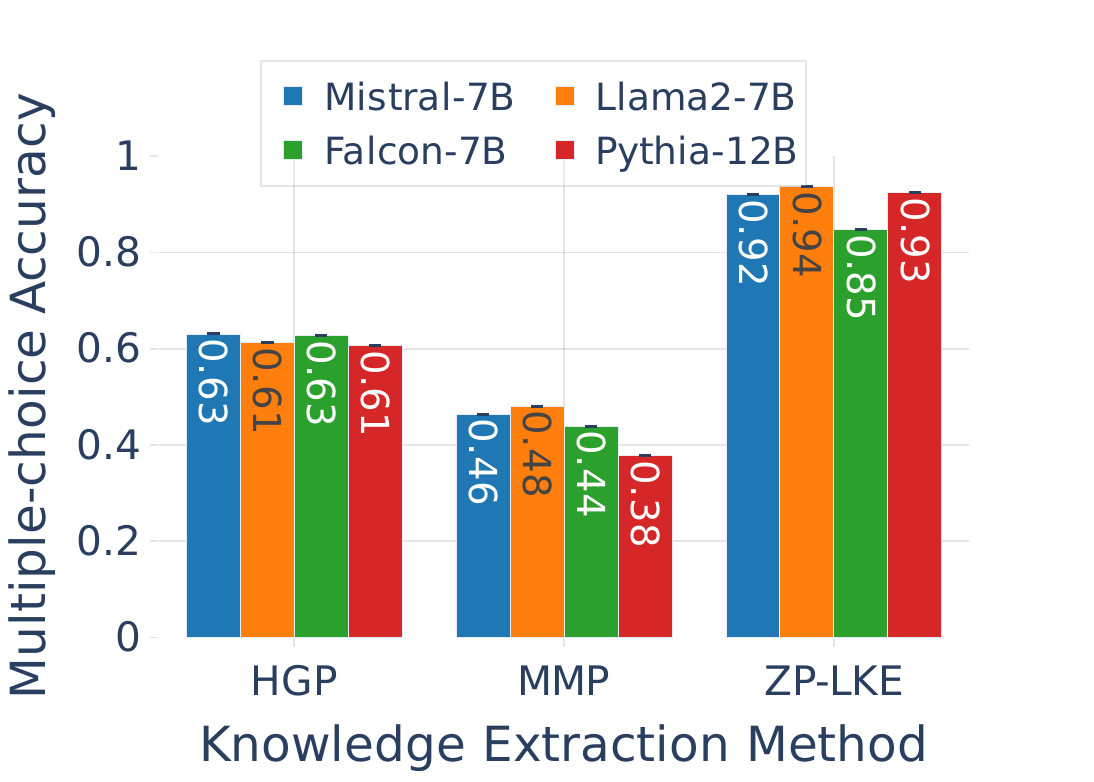}%
  \label{fig:manufacturer}%
}
\caption{Multiple-choice accuracy across all the 12 relations}
\label{fig:mul_acc_all}

\end{figure}

We compare \iclke~on 12 relations from \trexmc: \textit{capital, named after, developer, manufacturer, genre, instance of, native language, original broadcaster, language spoken written or signed, original language of film / TV show, official language, position played on team/speciality.}
We chose those 12 relations from \trexmc~ that are found to be in common with ~\cite{jiang2020can} where they define the templates for HGP and MMP. 
We evaluated 4 models (Mistral-7B, Llama-7B, Falcon-7B, and Pythia-12B) and showed all the results in  Figure~\ref{fig:respons_acc_all} and Figure~\ref{fig:mul_acc_all}.

\subsection{Full order of models and relations}
\label{appendix:all_relation_results}

We evaluated 49 models on 50 relations by our \iclke~. Table~\ref{table:model_order} shows the ordered models by the average accuracy of all the 50 relations. Table~\ref{table:ordered_relatoion} shows the ordered relations by the average accuracy of all the 49 models.

\begin{table}[htbp]
\centering
\caption{Model Performance Comparision}
\begin{tabular}{lcc}
\toprule
Model               & Average Accuracy & Standard Deviation \\
\midrule
Llama2-70B          & 0.8511           & 0.17591 \\
Mixtral-8x7B-SFT    & 0.84765          & 0.16919 \\
Mixtral-8x7B        & 0.84605          & 0.16653 \\
Llama-65B           & 0.84185          & 0.17528 \\
Mixtral-8x7B-DPO    & 0.81535          & 0.17580 \\
Llama-33B           & 0.81255          & 0.19088 \\
Mistral-7B          & 0.79310          & 0.20000 \\
Llama2-13B          & 0.78692          & 0.21892 \\
Llama-13B           & 0.76845          & 0.21796 \\
Llama2-70B-chat     & 0.75815          & 0.21272 \\
Llama2-7B           & 0.74945          & 0.24069 \\
Vicuna-13B          & 0.74940          & 0.21427 \\
Gemma-7B            & 0.74717          & 0.25668 \\
Openhermes-2.5      & 0.74365          & 0.21241 \\
Vicuna-13B-2        & 0.74080          & 0.22807 \\
Vicuna-7B-2         & 0.71695          & 0.24016 \\
Falcon-7B           & 0.70190          & 0.27052 \\
Vicuna-7B           & 0.70155          & 0.24724 \\
Llama2-13B-chat     & 0.69387          & 0.22966 \\
Llama-7B            & 0.69260          & 0.27912 \\
Gemma-2B            & 0.66600          & 0.28627 \\
GPT-NEOX-20B        & 0.66145          & 0.30972 \\
Llama2-7B-chat      & 0.66130          & 0.24996 \\
Mistral-instruct-7B & 0.66120          & 0.26173 \\
MPT-7B              & 0.64545          & 0.30638 \\
Pythia-12B          & 0.63325          & 0.32412 \\
OPT-6.7B            & 0.62110          & 0.31313 \\
GPT-J-6B            & 0.60965          & 0.32319 \\
OPT-13B             & 0.60845          & 0.31017 \\
Pythia-6.9B         & 0.59185          & 0.32359 \\
Bloom-7.1B          & 0.58270          & 0.31404 \\
OPT-30B             & 0.57925          & 0.31813 \\
Pythia-2.8B         & 0.57580          & 0.32773 \\
Pythia-1.4B         & 0.56330          & 0.33600 \\
Gemma-7B-instruct   & 0.55327          & 0.30689 \\
OPT-2.7B            & 0.55109          & 0.33260 \\
Bloom-3B            & 0.54375          & 0.29199 \\
Pythia-1B           & 0.54220          & 0.31560 \\
OPT-1.3B            & 0.53610          & 0.33335 \\
Bloom-1.1B          & 0.51115          & 0.29346 \\
OPT-350M            & 0.50735          & 0.30716 \\
Gemma-2B-instruct   & 0.49474          & 0.29628 \\
Pythia-410M         & 0.47995          & 0.29598 \\
Bloom-1.7B          & 0.47660          & 0.29658 \\
OPT-125M            & 0.45195          & 0.29330 \\
Bloom-560M          & 0.38465          & 0.28747 \\
Pythia-160M         & 0.37145          & 0.28505 \\
Pythia-70M          & 0.31260          & 0.27404 \\
Falcon-instruct-7B  & 0.00605          & 0.01459 \\
\bottomrule
\label{table:model_order}
\end{tabular}
\end{table}

\begin{table}[h!]
\centering
\caption{Relations and their average accuracies}
\begin{tabular}{cll}
\toprule
Order & Relation & Average Accuracy \\
\midrule
1 & publication date & 0.992071428571429 \\
2 & inception & 0.983214285714286 \\
3 & point in time & 0.975714285714286 \\
4 & drafted by & 0.922214285714286 \\
5 & native language & 0.8825 \\
6 & production company & 0.873428571428571 \\
7 & languages spoken, written or signed & 0.865071428571429 \\
8 & performer & 0.831142857142857 \\
9 & has played at & 0.826642857142857 \\
10 & capital & 0.815857142857143 \\
11 & is made by & 0.815357142857143 \\
12 & producer & 0.794714285714286 \\
13 & record label & 0.794571428571429 \\
14 & named after & 0.791071428571429 \\
15 & developer & 0.786928571428571 \\
16 & publisher & 0.7835 \\
17 & original broadcaster & 0.781214285714286 \\
18 & cast member & 0.777 \\
19 & home venue & 0.771714285714286 \\
20 & has subsidiary & 0.754142857142857 \\
21 & manufacturer & 0.749928571428571 \\
22 & screenwriter & 0.732285714285714 \\
23 & contains the administrative territorial entity & 0.7255 \\
24 & creates & 0.721214285714286 \\
25 & official language & 0.709857142857143 \\
26 & mother & 0.697857142857143 \\
27 & part of the series & 0.692214285714286 \\
28 & founded by & 0.684714285714286 \\
29 & original language of film or TV show & 0.6825 \\
30 & date of birth & 0.668857142857143 \\
31 & date of death & 0.641594184576485 \\
32 & instance of & 0.588990518331226 \\
33 & position played on team / speciality & 0.537642857142857 \\
34 & genre & 0.536 \\
35 & distributed by & 0.522785714285714 \\
36 & parent taxon & 0.488428571428571 \\
37 & director & 0.432928571428571 \\
38 & author & 0.331285714285714 \\
39 & father & 0.309214285714286 \\
40 & educated at & 0.306285714285714 \\
41 & characters & 0.282857142857143 \\
42 & composer & 0.276785714285714 \\
43 & child & 0.259142857142857 \\
44 & lyrics by & 0.258428571428571 \\
45 & sibling & 0.250285714285714 \\
46 & spouse & 0.238785714285714 \\
47 & is a tributary of & 0.212142857142857 \\
48 & cause of death & 0.206 \\
49 & discoverer or inventor & 0.173142857142857 \\
50 & student of & 0.123357142857143 \\
\bottomrule
\end{tabular}

\label{table:ordered_relatoion}
\end{table}

\subsection{Relation accuracy correlation of all the pre-trained models}

\begin{figure}[t]
    \centering
    \includegraphics[width=1\textwidth]{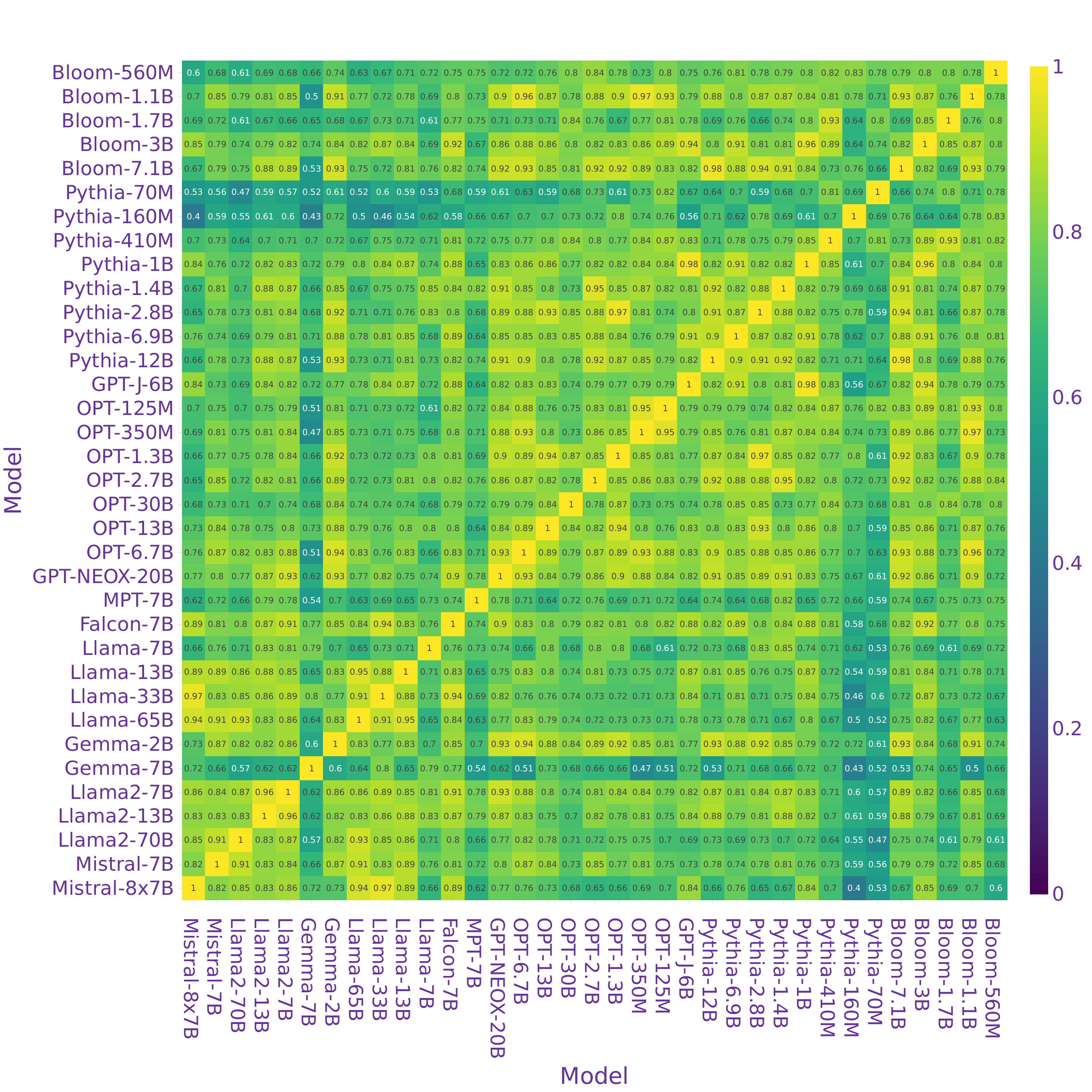}
    \caption{
    \textbf{[Pearson Correlation Coefficients Between All Pre-trained Models]}
    We calculated the Pearson correlation coefficients for each model pair among 49 models across 50 relations.
    }
    \label{fig:all_models_correlation}
\end{figure}

In Table \ref{fig:all_models_correlation}, we show the Pearson correlation coefficients between each model pair's performance across the 50 relations.

\begin{table}[h]
    \small
    \centering
     \scalebox{0.70}{
    \begin{tabular}{|l|l|l|l|l|l|l|l|}
    \hline
        \textbf{Order/Model} & \textbf{Mistral-8x7B} & \textbf{Mistral-7B} & \textbf{Llama2-70B} & \textbf{Llama2-13B} & \textbf{Llama2-7B} & \textbf{Gemma-7B} & \textbf{Gemma-2B} \\ \hline
        1 & publication date & point in time & point in time & publication date & publication date & point in time & point in time \\ \hline
        2 & point in time & date of death & inception & point in time & inception & inception & inception \\ \hline
        3 & inception & publication date & publication date & inception & point in time & publication date & publication date \\ \hline
        … & …… & …… & …… & ; & ; & ; & ~ \\ \hline
        48 & discoverer or inventor & discoverer or inventor & student of & discoverer or inventor & educated at & date of death & position played on team / speciality \\ \hline
        49 & cause of death & cause of death & cause of death & cause of death & cause of death & instance of & discoverer or inventor \\ \hline
        50 & student of & student of & is a tributary of & student of & student of & date of birth & student of \\ \hline
        \textbf{Order/Model} & \textbf{Llama-65B} & \textbf{Llama-33B} & \textbf{Llama-13B} & \textbf{Llama-7B} & \textbf{Falcon-7B} & \textbf{MPT-7B} & \textbf{GPT-NEOX-20B} \\ \hline
        1 & publication date & publication date & publication date & publication date & point in time & publication date & publication date \\ \hline
        2 & point in time & point in time & point in time & point in time & inception & inception & inception \\ \hline
        3 & inception & inception & inception & inception & publication date & point in time & date of death \\ \hline
        … & …… & …… & …… & ; & ; & ; & ~ \\ \hline
        48 & discoverer or inventor & discoverer or inventor & discoverer or inventor & student of & discoverer or inventor & student of & discoverer or inventor \\ \hline
        49 & cause of death & cause of death & cause of death & instance of & is a tributary of & is a tributary of & lyrics by \\ \hline
        50 & student of & student of & student of & date of birth & student of & educated at & student of \\ \hline
        \textbf{Order/Model} & \textbf{OPT-30B} & \textbf{OPT-13B} & \textbf{OPT-6.7B} & \textbf{OPT-2.7B} & \textbf{OPT-1.3B} & \textbf{OPT-350M} & \textbf{OPT-125M} \\ \hline
        1 & inception & publication date & publication date & inception & publication date & inception & inception \\ \hline
        2 & publication date & inception & inception & publication date & inception & publication date & publication date \\ \hline
        3 & point in time & point in time & date of death & point in time & drafted by & point in time & point in time \\ \hline
        … & …… & …… & …… & ; & ; & ; & ~ \\ \hline
        48 & position played on team / speciality & composer & lyrics by & discoverer or inventor & director & is a tributary of & student of \\ \hline
        49 & discoverer or inventor & student of & student of & student of & student of & spouse & is a tributary of \\ \hline
        50 & director & date of birth & discoverer or inventor & instance of & date of birth & student of & educated at \\ \hline
        \textbf{Order/Model} & \textbf{GPT-J-6B} & \textbf{Pythia-12B} & \textbf{Pythia-6.9B} & \textbf{Pythia-2.8B} & \textbf{Pythia-1.4B} & \textbf{Pythia-1B} & \textbf{Pythia-410M} \\ \hline
        1 & inception & point in time & publication date & publication date & publication date & publication date & publication date \\ \hline
        2 & publication date & publication date & inception & inception & inception & inception & inception \\ \hline
        3 & point in time & inception & point in time & point in time & date of death & point in time & drafted by \\ \hline
        … & …… & …… & …… & …… & …… & …… & …… \\ \hline
        48 & lyrics by & lyrics by & lyrics by & student of & discoverer or inventor & lyrics by & student of \\ \hline
        49 & student of & genre & director & date of birth & lyrics by & student of & discoverer or inventor \\ \hline
        50 & date of death & director & date of death & lyrics by & director & date of death & is a tributary of \\ \hline
        \textbf{Order/Model} & \textbf{Pythia-160M} & \textbf{Pythia-70M} & \textbf{Bloom-7.1B} & \textbf{Bloom-3B} & \textbf{Bloom-1.7B} & \textbf{Bloom-1.1B} & \textbf{Bloom-560M} \\ \hline
        1 & publication date & publication date & publication date & publication date & publication date & publication date & publication date \\ \hline
        2 & point in time & point in time & inception & inception & inception & inception & inception \\ \hline
        3 & date of death & native language & date of death & point in time & point in time & date of death & point in time \\ \hline
        … & …… & …… & …… & …… & …… & …… & …… \\ \hline
        48 & student of & official language & lyrics by & is a tributary of & screenwriter & is a tributary of & is a tributary of \\ \hline
        49 & capital & instance of & student of & spouse & student of & spouse & director \\ \hline
        50 & director & date of death & spouse & student of & spouse & student of & student of \\ \hline
    \end{tabular}
    }
    \caption{Top 3 and Bottom 3 relations for each pre-trained model}
    \label{table:top3_relation_order}
\end{table}

\clearpage
\subsection{Impact of finetuning}
\label{appendix:impact_finetuning}
We also show the results for the average subsumption rate ($\asubr$) in Table \ref{tab:avg_subsumption_ft} for base models and fine-tuned models over the relations in \trexmc.

\begin{table}
    \centering
    \begin{tabular}{ccccc|c}

        Family  & Model Type & Accuracy & Model Type & Accuracy  & $\asubr$ \\
        \midrule
        Llama-7B & Base & 0.699 & FT-1 & 0.693  & 0.779  \\
        Llama-13B & Base & 0.770 & FT-1 & 0.735  & 0.854  \\
        Llama2-7B & Base & 0.741 & FT-1 & 0.712  & 0.808  \\
        Llama2-7B & Base & 0.741 & FT-2 & 0.664  & 0.790  \\
        Llama2-13B & Base & 0.771 & FT-1 & 0.748  & 0.831  \\
        Llama2-13B & Base & 0.771 & FT-2 & 0.692  & 0.801  \\
        Llama2-70B & Base & 0.846 & FT-1 & 0.739  & 0.811  \\
        Mistral-7B & Base & 0.793 & FT-1 & 0.639  & 0.793 \\
        Mistral-7B & Base & 0.793 & FT-2 & 0.750  & 0.869 \\
        Mixtral-7Bx8 & Base & 0.832 & FT-1 & 0.835  & 0.928 \\        
        Mixtral-7Bx8 & Base & 0.832 & FT-2 &  0.817  & 0.911 \\     
        Gemma-2B & Base & 0.666 & FT-1 &   0.488  & 0.577\\ 
        Gemma-7B & Base &  0.749 & FT-1 &    0.511  & 0.557\\ 
        \bottomrule
    \end{tabular}
    \caption{
    Average subsumption rate ($\asubr$) for base models and fine-tuned models over the relations in \trexmc.
    Despite being fine-tuned on smaller datasets, fine-tuned models  (low $\asubr$). The results are based on \iclke. Accuracy in this table is multiple-choice accuracy.
    }
    \label{tab:avg_subsumption_ft}
\end{table}

\clearpage

\end{document}